\documentclass[10pt,twocolumn,letterpaper]{article}

\usepackage{cvpr}
\usepackage{times}
\usepackage{epsfig}
\usepackage{graphicx}
\usepackage{amsmath}
\usepackage{amssymb}
\usepackage{epstopdf}
\usepackage{textcomp}
\usepackage{algorithmic} 
\usepackage{algorithm}

\newcommand{\calN}{{\cal N}}

\cvprfinalcopy 

\def\cvprPaperID{1914} 

\ifcvprfinal\fi
\begin{document}

\title{Efficient Hierarchical Graph-Based Segmentation of RGBD Videos}
\author{
Steven Hickson$^1$\\
{\tt\small shickson@gatech.edu}
\and
Stan Birchfield$^2$\\
{\tt\small stanleyb@microsoft.com}
\and
Irfan Essa$^1$\\
{\tt\small irfan@cc.gatech.edu}
\and
Henrik Christensen$^1$\\
{\tt\small hic@cc.gatech.edu}
\\
$^1$Georgia Institute of Technology, Atlanta, GA, USA
 , $^2$Microsoft Robotics, Seattle, WA, USA
\\
\and
\url{http://www.cc.gatech.edu/cpl/projects/4dseg}
}

\maketitle

\begin{abstract}
We present an efficient and scalable algorithm for segmenting 3D RGBD point clouds by combining depth, color, and temporal information using a multistage, hierarchical graph-based approach.  Our algorithm processes a moving window over several point clouds to group similar regions over a graph, resulting in an initial over-segmentation.  These regions are then merged to yield a dendrogram using agglomerative clustering via a minimum spanning tree algorithm.  Bipartite graph matching at a given level of the hierarchical tree yields the final segmentation of the point clouds by maintaining region identities over arbitrarily long periods of time. We show that a multistage segmentation with depth then color yields better results than a linear combination of depth and color. Due to its incremental processing, our algorithm can process videos of any length and in a streaming pipeline. The algorithm's ability to produce robust, efficient segmentation is demonstrated with numerous experimental results on challenging sequences from our own as well as public RGBD data sets.
\end{abstract}

\section{Introduction}

\noindent
Segmentation is used as a building block for solving problems such as object detection, activity recognition, and scene understanding. Traditionally, segmentation involves finding pixels with similar perceptual appearance in a \emph{single image} and grouping them.  In recent years, there has been increasing emphasis upon \emph{video segmentation}, in which pixels with similar appearance and spatiotemporal continuity are grouped together over a video volume\cite{Paris2007,Grundmann2010,Xu2012}.  Given the widespread availability of depth data from low-cost RGBD sensors we are interested in robust unsupervised segmentation of \emph{streaming 3D point clouds} with temporal coherence.

\noindent
Our algorithm is aimed at \emph{super-toxel} extraction; since voxels are limited to 3D data, we define a new term \emph{toxels}. Toxels are  \emph{temporal voxels}, which can be understood as a generalization of voxels for 4D hyperrectangles. Each toxel is a hypercube discretization of a continuous 4D spatiotemporal hyperrectangle. Voxels are generally used to refer to the discretization of 3D volumes or the discretization of 2D frames over time. When these are combined, we get XYZT data, which can be discretized using toxels. Super-toxels are just large groupings of toxels.\\

\begin{figure}[t]
\begin{center}
   \includegraphics[width=0.9\linewidth]{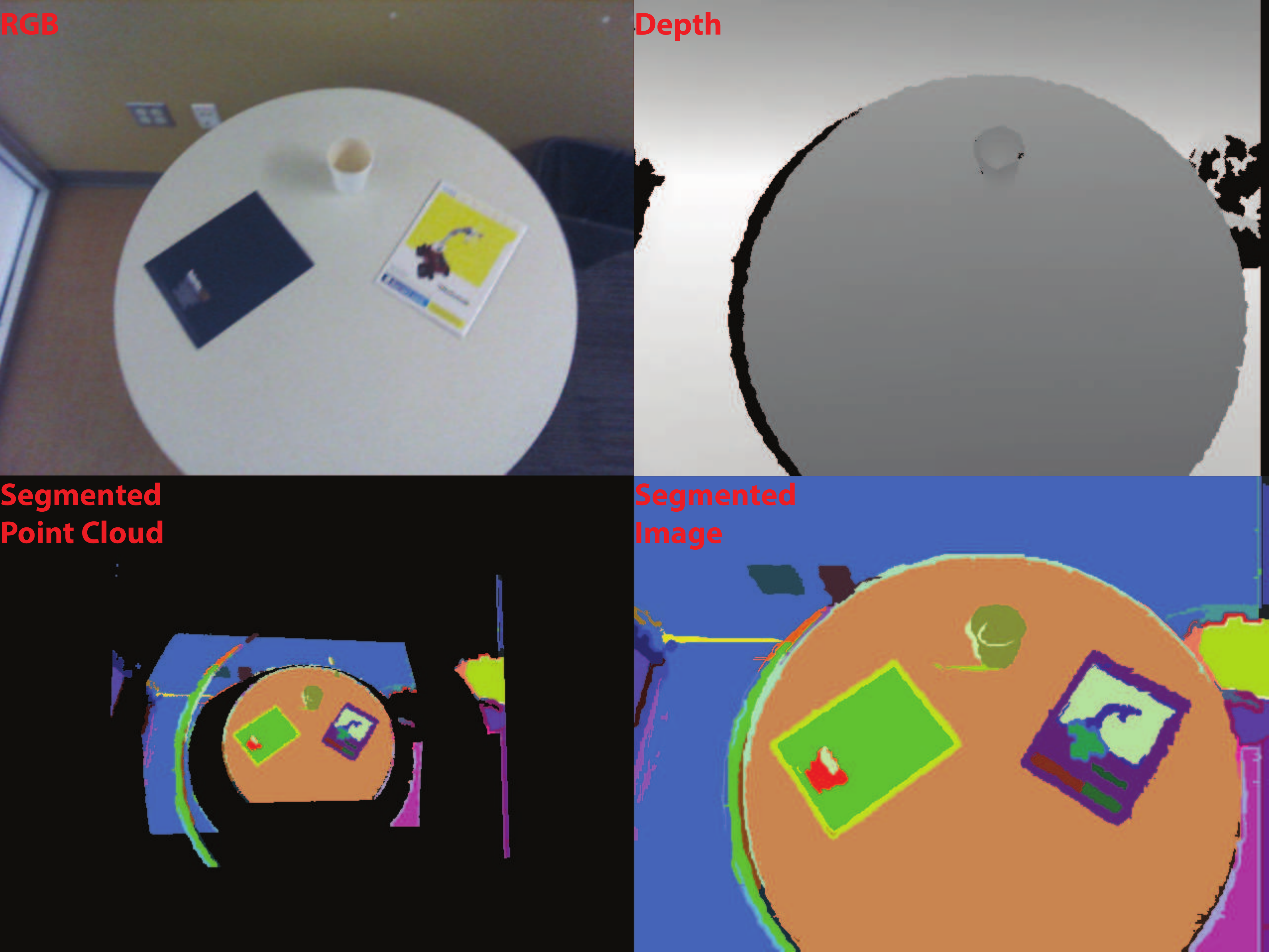}
\end{center}
   \caption{Segmentation of streaming 3D Point clouds. Top: RGB and depth maps used as input in our algorithm. Bottom: Rendered version of the point cloud and its frame segmentation output. }
\label{fig_intro}
\end{figure}

\noindent
Our method uses measurements such as color, spatial coordinates, and RGBD optical flow to build a hierarchical region tree for each sequence of $n$ frames (we use $n=8$), which are then sequentially matched to produce long-term continuity of region identities throughout the sequence. This bottom-up design avoids limitations on the length of the video or on the amount of memory needed due to the high volume of data. In contrast to segmentation-tracking methods, we do not assume \emph{a priori} knowledge of the content of the scene and/or objects, and we do not perform initialization of seeds or any other type of human intervention. Our motivation is that regions should not merge across depth boundaries despite color similarity.  
Our method was tested on several challenging sequences from the TUM Dataset \cite{sturm2012benchmark}, the NYU Dataset \cite{Silberman2012}, and our own examples. It produces meaningful segmentation, even under difficult conditions such as significant motion of scene objects, camera motion, illumination changes, and partial occlusion.

\noindent\textbf{Our primary contributions are:} (1) A robust and scalable, RBGD video segmentation framework for streaming data. (2) A streaming method that maintains temporal consistency and can be run on a robot or on videos of indefinite length. (3) An efficient framework that runs at 0.8 fps but can be downsized to run pairwise in near real-time at 15 fps. (4) A nonlinear multistage method to segment color and depth that enforces that regions never merge over depth boundaries despite color similarities. And finally (5) The ability to segment and maintain temporal coherence with camera movement and object movement. The code and data are publicly available at the paper web page.

\begin{figure*}[t]
\begin{center}
   \includegraphics[width=1\linewidth]{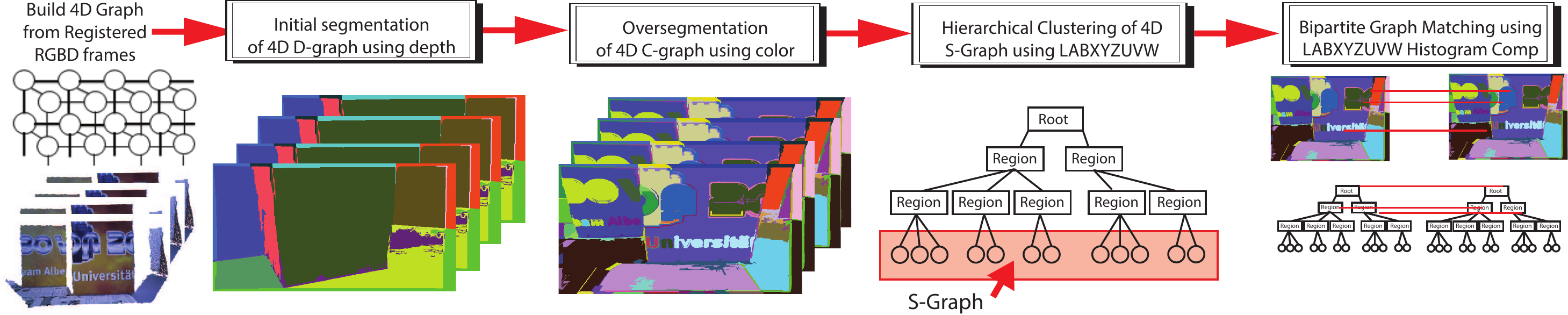}          
\end{center}
\caption{Schematic of our method to segment streaming 3D point clouds.}   
\label{fig_method}
\end{figure*}

\section{Related Work}

\noindent
One way to organize related work is use the five types of algorithms for super-voxel video segmentation analyzed by Xu and Corso~\cite{Xu2012}.
(A)~Paris and Durand ~\cite{Paris2007} propose a \emph{mean shift} method that achieves hierarchical segmentation in videos using topological persistence using the classic mode-seeking meanshift algorithm interpreted under Morse theory as a topological decomposition of the feature space.
(B)~\cite{Shi2000} use \emph{Nystrom normalized cuts}, in which the Nystrom approximation is applied  to solve the normalized cut problem, for  spatiotemporal grouping.
(C)~\emph{Segmentation by Weighted Aggregation (SWA)}~\cite{Sharon2006} is a variant of optimizing the normalized cut that computes a hierarchy of sequentially coarser segments by an algebraic multigrid solver.
(D)~\emph{Graph-Based (GB)} is an adaptation of the Felzenszwalb and Huttenlocher image segmentation algorithm~\cite{Felzenszwalb2004}  to video segmentation by building the graph in the spatiotemporal volume where voxels (volumetric pixels) are nodes connected to 26 neighbors.
(E)~\emph{Hierarchical graph based (GBH)} is an algorithm for video segmentation proposed in~\cite{Grundmann2010} that iteratively builds a tree structure of region graphs, starting from  over-segmented spatiotemporal volumes  obtained using the method illustrated above. The regions are described by LAB histograms of the voxel members, the edge weights are defined by the $\chi^2$ distance, and the regions are merged using the same technique as in~\cite{Felzenszwalb2004}.

\noindent
Surveying the literature, there has been little work on unsupervised segmentation of streaming 3D point clouds besides \cite{abramov2012depth}, which uses a parallel Metropolis algorithm and the Potts model and focuses only on a specific environment and \cite{weikersdorfer}, which combines color and depth and uses spectral graph clustering without a hierarchical method. We are unable to compare against either of these since no code or data was released. Additionally, most of the research conducted in 3D segmentation of point clouds has been focused on static scenes, without considering time. For example, planar and non-planar surfaces are segmented from 3D point clouds using either NURBS~\cite{Holz2012, AndreasRichtsfeld2012}, surface normal segmentation \cite{strom2010graph}, or 3D shape matching~\cite{Munoz2009} to identify objects with a particular shape. Scene modeling is performed by analyzing support relationships of the regions~\cite{Silberman2012} or contextual modeling of both super-pixel MRFs and paths in segmentation trees~\cite{Ren2012}. Temporal evolution of the 3D point cloud has been considered in cases where a learned 3D model of the segmented object is available, such as in the simultaneous segmentation and pose tracking approach of~\cite{Prisacariu2012}, the rigid object tracking approach of~\cite{Rusu2011}, or the segmentation-tracking method of an arbitrary untrained object in~\cite{Teichman2012}.

\section{Proposed Method}

\noindent
Our approach involves four main steps, as illustrated in Figure~\ref{fig_method}. First, a segmentation of 8 consecutive frames using the depth and motion information is performed. In the second step, an over-segmentation of the frames is done using color and motion information while respecting depth boundaries. Next, histograms of the resulting regions are used to build a hierarchical segmentation of the spatiotemporal volume represented as a dendrogram, which can then yield a particular segmentation depending on the desired segmentation level output. The final step performs a bipartite graph matching with the 8 previous frames with 4 frames overlapping to enforce the consistency of region identities over time.

\subsection{Spatiotemporal segmentation}

\paragraph{Graph-based segmentation:} A natural approach to segmenting RGBD images would be to simply use \cite{Felzenszwalb2004} after setting the edge weights to a weighted combination of differences in depth and color:  $(1-\alpha)d_C(p_1,p_2) + \alpha d_D(p_1,p_2)$, where $p_1$ and $p_2$ are two pixels, $d_C$ is the difference in color, $d_D$ is the difference in depth, and $\alpha$ is a scalar weighting the relative importance between the two.  We shall show in the experimental results in Figure~\ref{fig_fig6} that this approach does not yield good results in practice, because there is no value of $\alpha$ that will work consistently well for a variety of images.  In fact, in many cases no value of $\alpha$ works well even for the various regions of a single image. Instead we use the multistage approach described in Section \ref{sec_supertoxel}.

\paragraph{RGBD optical flow:} We compute optical flow between each consecutive pair of RGBD images.  First, the depth values are converted to XYZ coordinates in 3D space by using the internal and external calibration parameters of the visible and IR cameras on the sensor to register the RGB and depth images.  Then we compute dense optical flow between the two RGB images.  Previous extensions of optical flow to 3D~\cite{Barron2005} assume a continuous  spatiotemporal volume of intensity values on which differentiation is then performed. In our case, however, the depth values are already known, thus enabling a much simpler calculation.  We compute 2D optical flow  $(\Delta{i},\Delta{j})$ using a consecutive pair of RGB images based on Farneback's method~\cite{farneback2003two}.  Therefore, since the RGB and depth images are registered, for the value $D_{n-1}(i,j)$ in the depth map at location $(i,j)$ at time $n$, there exists a corresponding depth value $D_{n}(i+\Delta{i},j+\Delta{j})$ in the next frame.  The optical flow, w, along the $z$ axis, is then simply the difference between the two depths:
\begin{equation}  
w = D_{n}(i+\Delta{i},j+\Delta{j}) - D_{n-1}(i,j).
\label{eq_w_comp}
\end{equation}
We also project the motion of the $(i,j)$ pixels into 3D space using the camera calibration parameters. 

Hence, the RGBD scene flow can be solved by extending Equation~\ref{eq_w_comp} and the calibration of the data and is defined as:
\begin{equation}
\begin{split}
(u,v,w) = & ((i_{n} - i_{n-1}) \frac{z_{n}}{F_x} , \\
& (j_{n} - j_{n-1}) \frac{z_{n}}{F_y} , \\
& D_{n}(i+\Delta{i},j+\Delta{j}) - D_{n-1}(i,j)).
\end{split}
\label{eq_flow}
\end{equation}

\subsubsection{Segmentation using depth and color}\label{sec_supertoxel}

Our approach relies on the following observation:
If the scene consists only of convex objects, then every depth discontinuity corresponds to an object boundary.
This observation was motivated by a study on primates \cite{livingstone1988segregation} which concluded that depth and color perception are handled by separate channels in our nervous system that function quite distinctly.
Obviously the world does not consist only of convex objects, but nevertheless we have found this observation to have great practical utility in everyday scenes.  We exploit this observation by first segmenting based on depth alone, then based on color while preventing any merging to occur across depth discontinuities.  We show in the experimental results that this approach yields more accurate segmentation than combining depth and color into a single graph.

We build a graph (called the D-graph) shown in Figure~\ref{fig_method}, in which each node corresponds to a toxel.  Within each graph, nodes are connected to their 26-neighbors with edges whose weights are the absolute difference in the depth of the two toxels:  $|D(x,y,z,t)-D(x',y',z',t')|$, where $D(x,y,z,t)$ is a function that yields the depth values from the spatiotemporal volume, and $(x',y',z',t') \in \calN(x,y,z,t)$, where $\calN(x,y,z,t)$ is the neighborhood of toxel $(x,y,z,t)$.  Edges are then constructed between each frame $D(x,y,z,t-1)$ and $D(x,y,z,t)$ by connecting toxels based on the computed optical flow. Toxel $(x,y,z,t-1)$ in $D$ is connected with toxel $(x+u,y+v,z+w,t)$ in $D$ with an edge whose weight is given by $|D(x,y,z,t-1)-D(x+u,y+v,z+w,t)|$.  The internal difference of a region $Int(R)$ is set as the highest edge weight in the minimum spanning tree of region $R$, and regions are merged according to \cite{Felzenszwalb2004}. The constant that determines the size of depth segments is called $k_{depth}$.

After the graph-based algorithm produces a segmentation according to the depth, another graph (called the C-graph) is created with the same nodes as before shown in Figure~\ref{fig_method}.  As before, edges are created for each of the 26-neighbors, and edges between frames are set according to the RGBD optical flow vectors to be $|C(x,y,z,t-1)-C(x+u,y+v,z+w,t)|$ where $C(x,y,z,t)$ is a function that returns the euclidean difference in color. We use CIE LAB color space to ensure perceptual uniformity of color differences.  Invoking the observation above, the edge weights connecting toxels that belong to different regions after processing the D-graph are set to infinity in the C-graph to forbid merging across depth discontinuities.  The constant that determines the size of color segments is called $k_{color}$.  After running the graph-based algorithm on the C-graph, we have an over-segmentation consisting of super-toxel regions across the spatiotemporal hyperrectangle.

\subsection{Hierarchical Processing}

\noindent
Once the over-segmentation has been performed, feature vectors are computed for each super-toxel region.  For feature vectors, we use histograms of color, 3D position, and 3D optical flow, called LABXYZUVW histograms, using all the toxels in a region.  For computational efficiency and storage requirements, rather than computing a single 9-dimensional histogram, we compute nine 1-dimensional histograms.  Experimentally we find that 20 bins in each of the LAB and UVW features and 30 bins in each of the XYZ features is adequate to balance generalization with discriminatory power.  

\noindent
Using these feature vectors, a third graph (called the S-graph) is created to represent the regions as shown in Figure~\ref{fig_method}. Each node of the graph is a region computed by the super-toxel segmentation in Section \ref{sec_supertoxel} and edges are formed to join regions that are adjacent to each other (meaning that at least one toxel in one region is a 26-neighbor of at least one toxel in the other region).  The weight of each edge is the difference between the two feature vectors, for which we use Equation~\ref{eq_h}.  On this graph, instead of running the graph-based algorithm of~\cite{Felzenszwalb2004}, we use Kruskal's algorithm to compute the minimum spanning tree of the entire set of regions.  The result of this algorithm is a dendrogram that captures a hierarchical representation of the merging in the form of a tree, in which the root node corresponds to the merging of the entire set.  

\noindent
By selecting a different threshold, a different cut in this minimum spanning tree can be found.  To make the threshold parameter intuitive, we choose the percentage of regions to merge, which we denote by $\zeta$.  The value determines the coarseness of the produced results, with $\zeta=0$ indicating that all the regions are retained from the previous step, whereas $\zeta=100\%$ indicates that all the regions are merged into one.  For values between these two extremes, the dendrogram is recursively traversed, accumulating all nodes until the desired percentage is obtained.  Although the value $\zeta$ can be changed on-the-fly for any given image without incurring much computation to recompute the segments, we generally set $\zeta=65\%$ for the entire sequence to avoid having to store all dendrograms for all frames.

\subsection{Bipartite Graph Matching}	

\noindent
After the dendrogram of the current pair of frames has been thresholded according to $\zeta$, correspondence must be established between this thresholded dendrogram and that of the previous pair of frames.  To achieve this correspondence, we perform bipartite graph matching between the two thresholded dendrograms using the stable marriage solution \cite{gale1962college}.  Figure~\ref{fig_method} depicts the bipartite graph matching process, which considers the difference in region sizes, the change in location of the region centroid (after applying the 3D optical flow), and the difference between the LABXYZUVW histograms.

\noindent
Data is treated as two lists from the tree cut. The histogram match from each region's perspective is found with the histogram SAD difference equation between Region $R$ and Region $S$ being:
\small
\begin{equation}  
\begin{split}
\Delta H = \sum_{i=1}^{\mbox{NUMBINS}}  \lvert \frac{R_l[i]}{R_N} - \frac{S_l[i]}{S_N} \rvert + \lvert \frac{R_a[i]}{R_N} - \frac{S_a[i]}{S_N} \rvert \\
+ \lvert \frac{R_b[i]}{R_N} - \frac{S_b[i]}{S_N} \rvert + \lvert \frac{R_x[i]}{R_N} - \frac{S_x[i]}{S_N} \rvert + \lvert \frac{R_y[i]}{R_N} - \frac{S_y[i]}{S_N} \rvert \\
+ \lvert \frac{R_z[i]}{R_N} - \frac{S_z[i]}{S_N} \rvert + \lvert \frac{R_u[i]}{R_N} - \frac{S_u[i]}{S_N} \rvert + \lvert \frac{R_v[i]}{R_N} - \frac{S_v[i]}{S_N} \rvert \\
+ \lvert \frac{R_w[i]}{R_N} - \frac{S_w[i]}{S_N} \rvert 
\end{split}
\end{equation}
\normalsize
where $R_N$ is the number of toxels in Region $R$ and $S_N$ is the number of toxels in Region $S$.

\noindent
For each histogram, we compute the distance the centroid has traveled over time and the size difference.\\
The difference the centroid has traveled is computed as the SAD of the center points normalized by the size:
\begin{equation}
\Delta d = \frac{\lvert R_x - S_x \rvert + \lvert R_y - S_y \rvert + \lvert R_z - S_z \rvert}{R_N}
\end{equation}
and the size difference is computed as the absolute value of the difference:
\begin{equation}
\Delta N = \lvert R_N - S_N \rvert
\end{equation}

\noindent
Starting with the closest histogram, we merge the regions $R$ and $S$ $iff$ they each consider each other the best choice using the weight as defined by Equation~\ref{eq_h}:
\begin{equation}
h = \beta \Delta H + \gamma \Delta d + \epsilon \Delta N
\label{eq_h}
\end{equation}
where $\beta$, $\gamma$, and $\epsilon$ are fixed constraints to prevent over-merging, found by computing statistics on multiple sequences and determining $3\sigma$ for each constraint.

\section{Experimental Results}

\noindent
Our algorithm was evaluated with several 3D RGBD sequences, some from the NYU Dataset \cite{Silberman2012}, some from the TUM Dataset \cite{sturm2012benchmark}, and some that we obtained.  To measure the accuracy of the algorithm, we annotated some images by hand to provide ground truth as 2D object boundaries.  To account for slight inaccuracies in the position error of a boundary point, we use the chamfer distance~\cite{Barrow1977}.  The boundary edges are extracted from the output segmentation images as well as of the ground truth images. For each pixel in the contour of the query image, we find the distance to the closest contour pixel in the ground-truth image (which is efficiently computed using the chamfer distance), and then we sum these distances and normalize by dividing by the number of pixels in the image. The error metric is shown in Equation~\ref{eq_err}.

\begin{equation}  
E_{bound}=  \frac{\sum_i \| {NN}^1( \mbox{Out}_i - \mbox{GroundTruth})\| }{\mbox{Width} * \mbox{Height}} ,
\label{eq_err}
\end{equation}
Where ${NN}^1$ represents the closest point of the boundary output $\mbox{Out}_i$ to the ground truth.

\begin{figure}[t]
	\centering
	\begin{tabular}{cc}
		\includegraphics[width=0.42\linewidth]{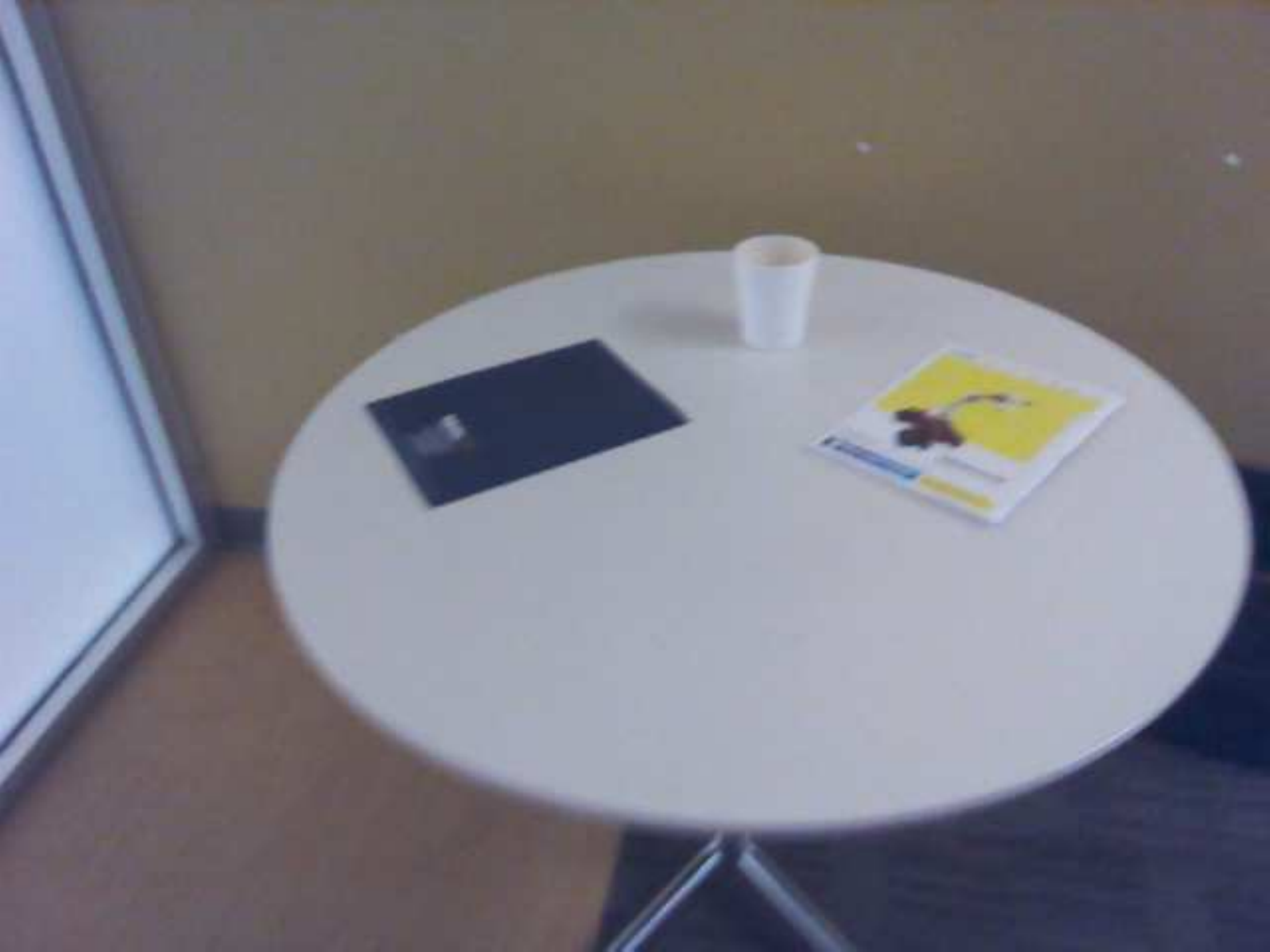} &
		\includegraphics[width=0.42\linewidth]{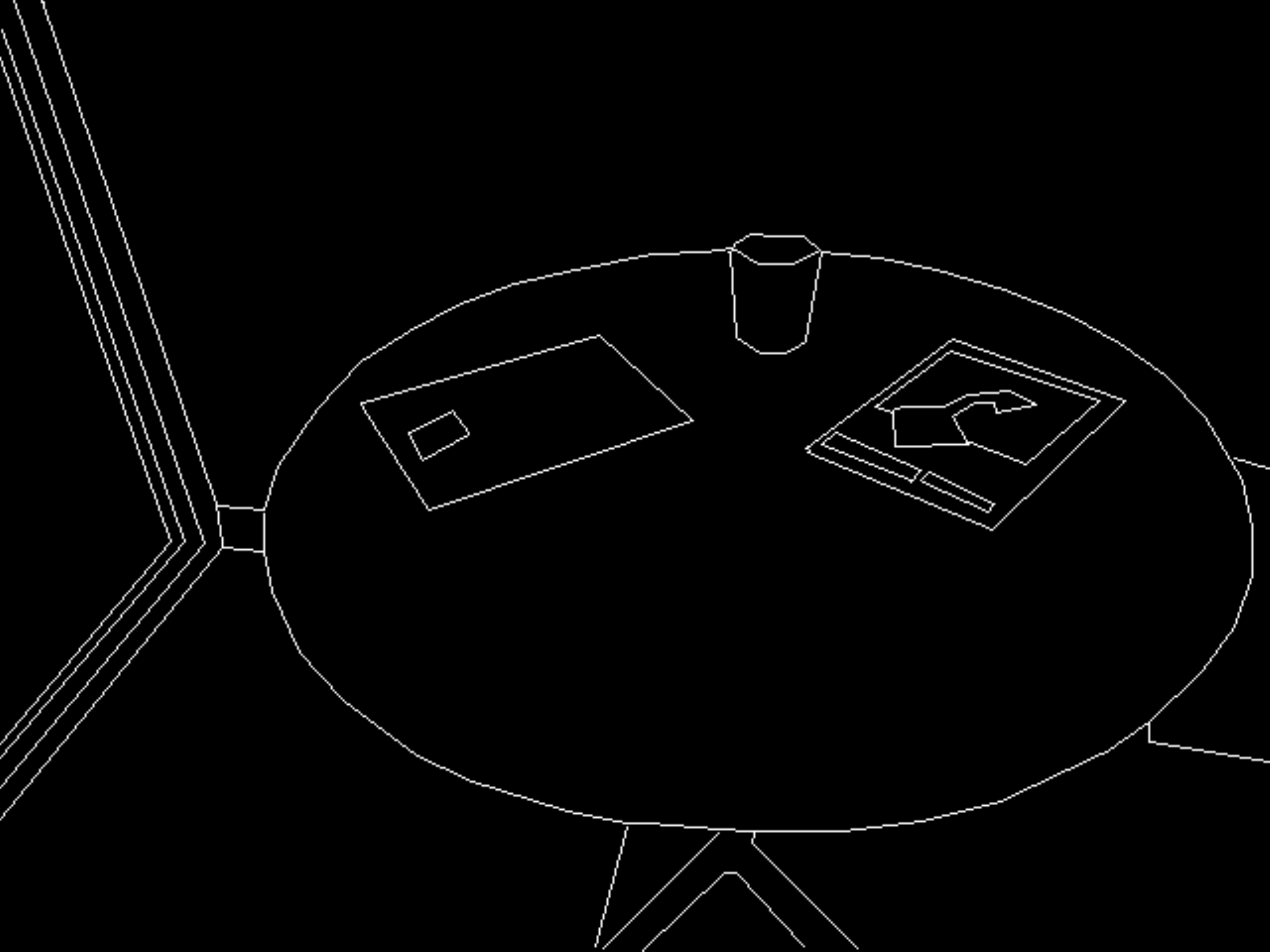} \\
		\includegraphics[width=0.42\linewidth]{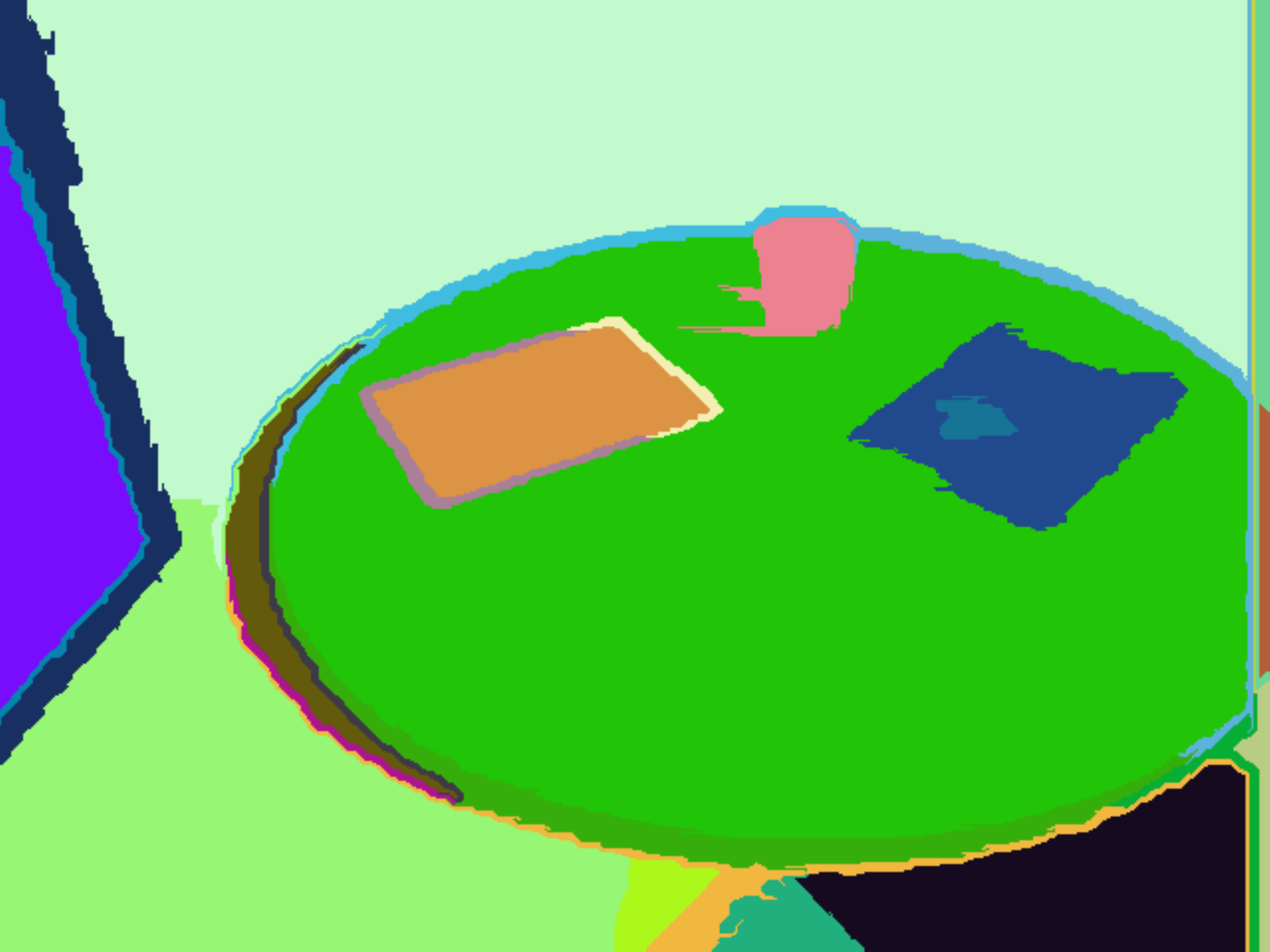} &		
		\includegraphics[width=0.42\linewidth]{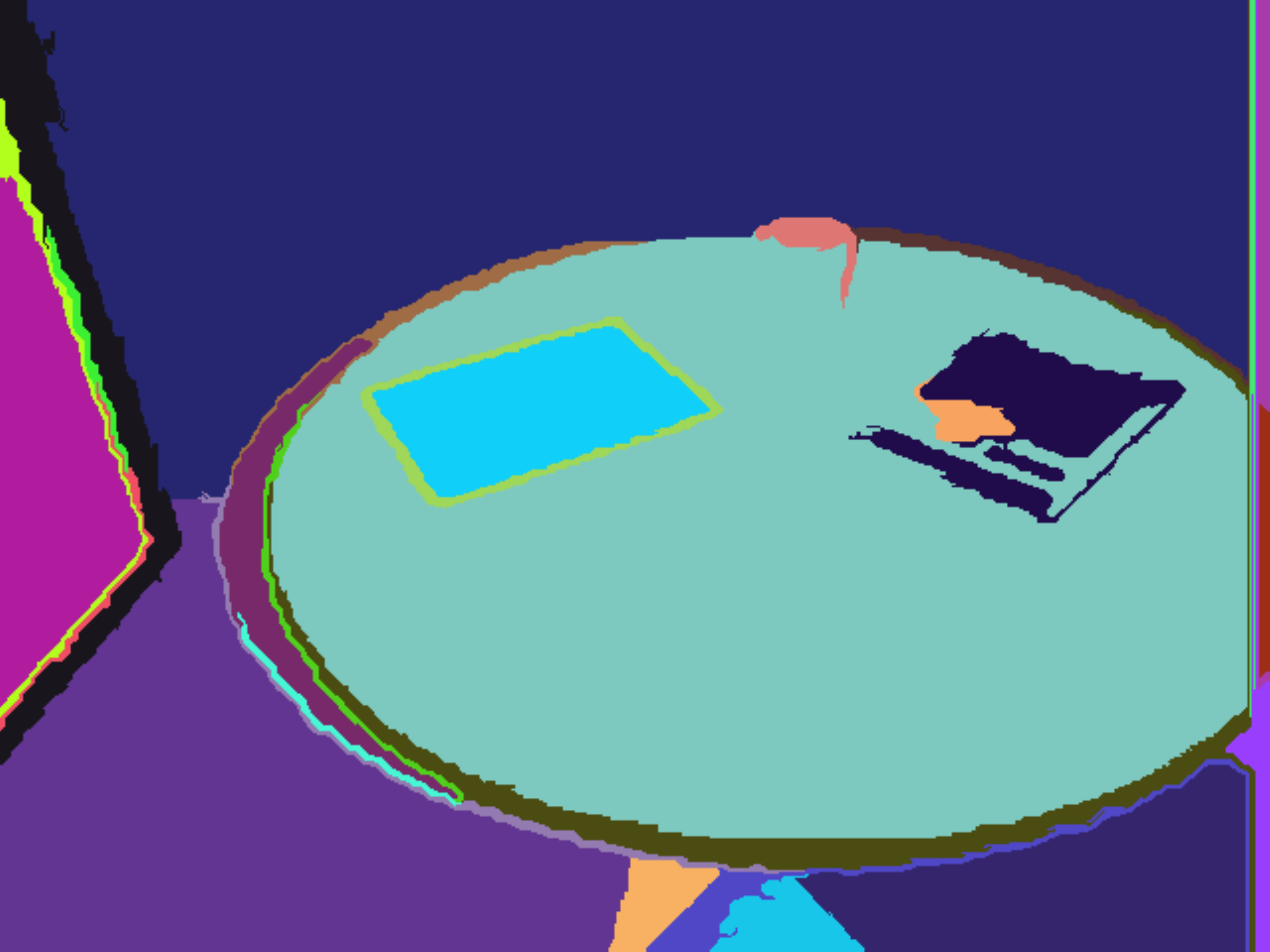} \\
		\includegraphics[width=0.42\linewidth]{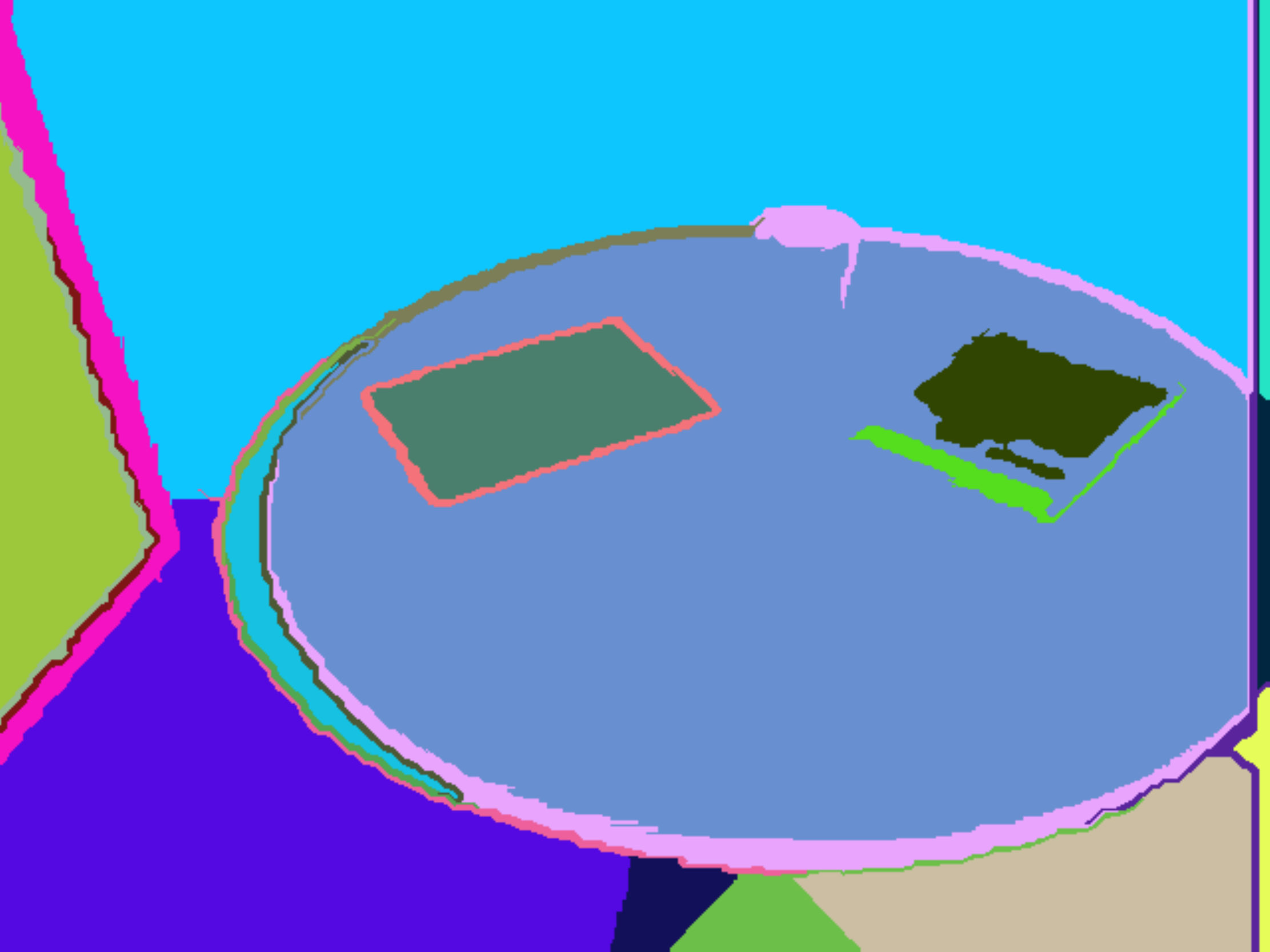} &		
		\includegraphics[width=0.42\linewidth]{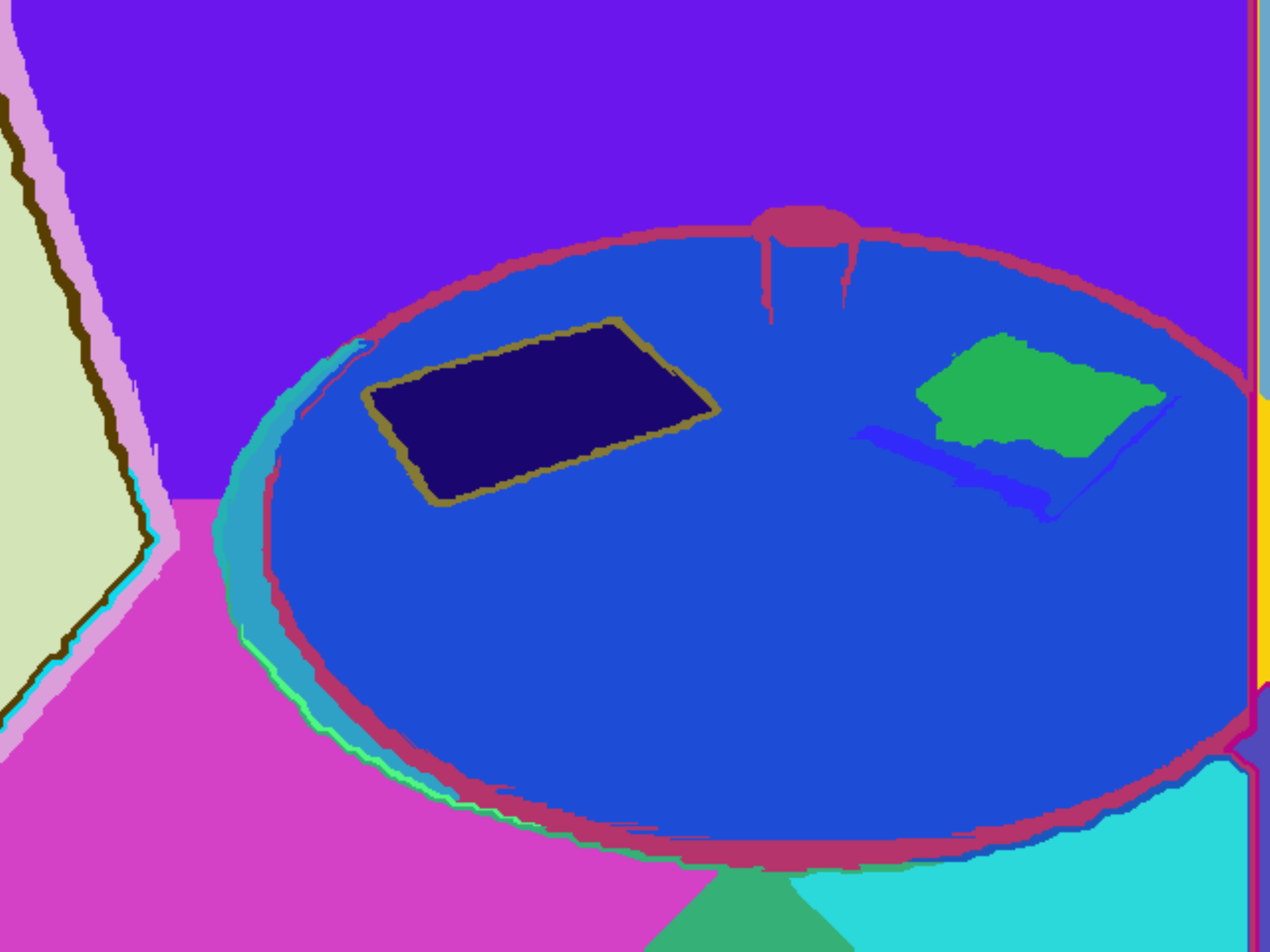} \\
		\includegraphics[width=0.42\linewidth]{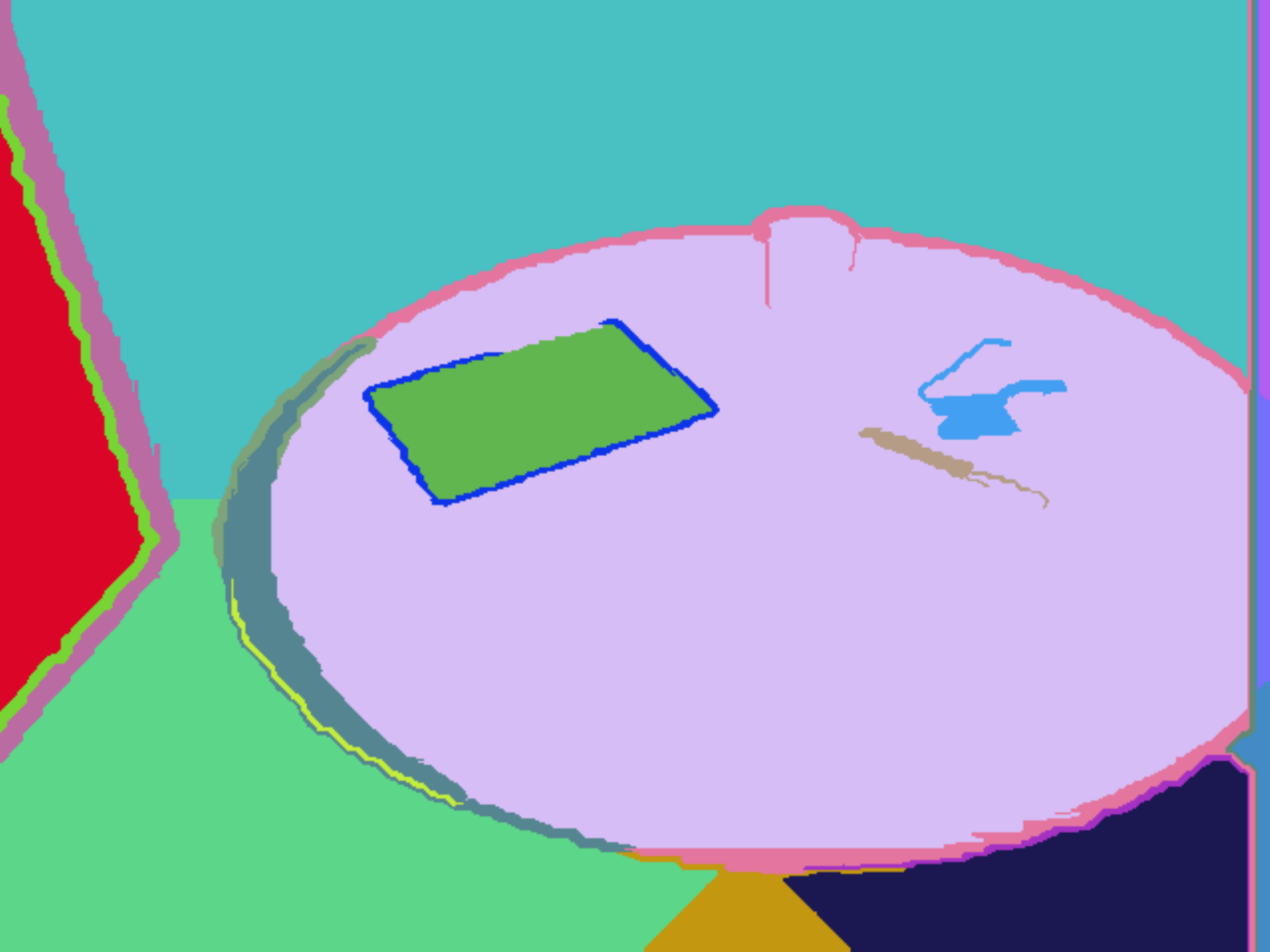} &		
		\includegraphics[width=0.42\linewidth]{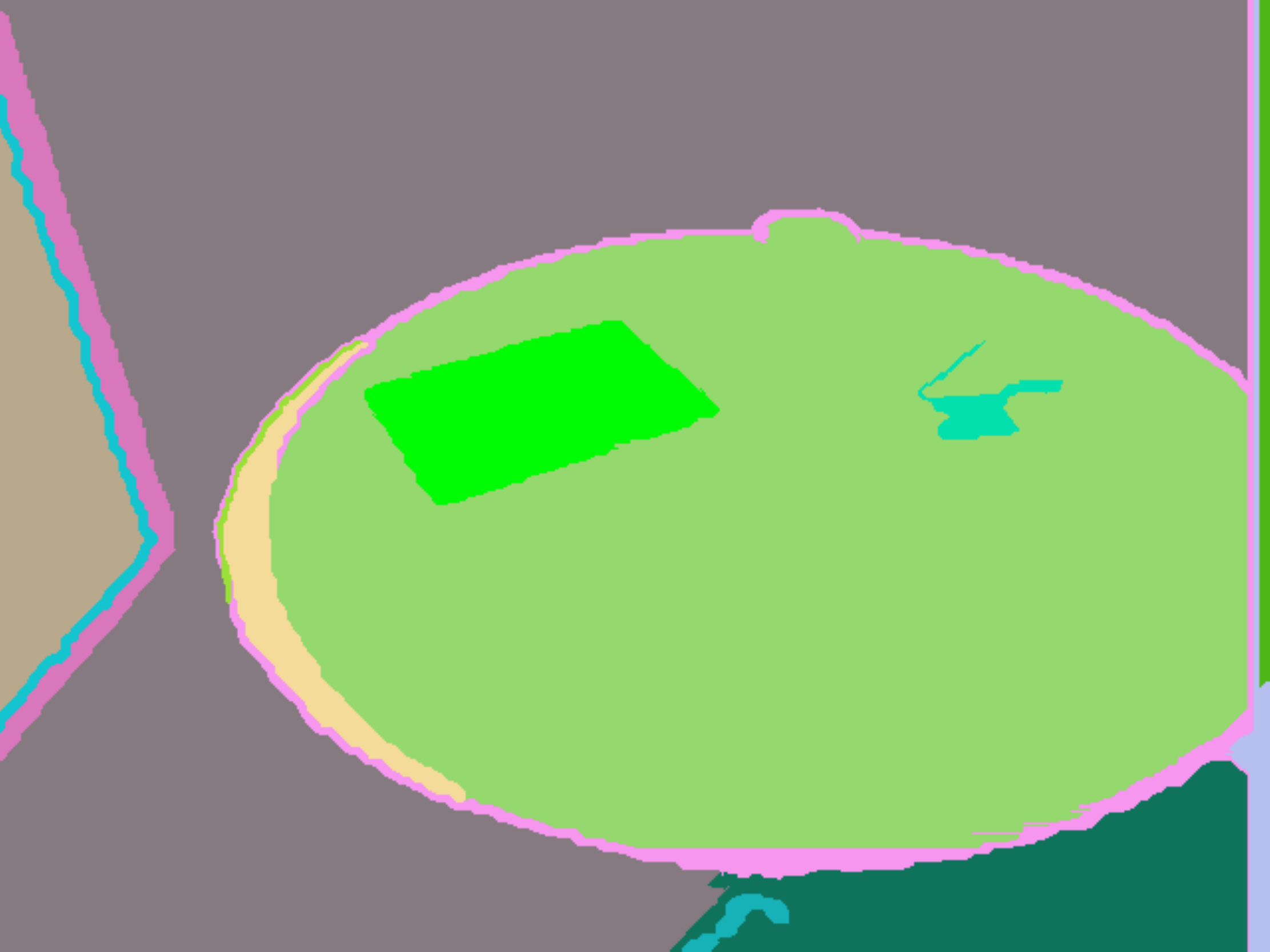} \\
		\includegraphics[width=0.42\linewidth]{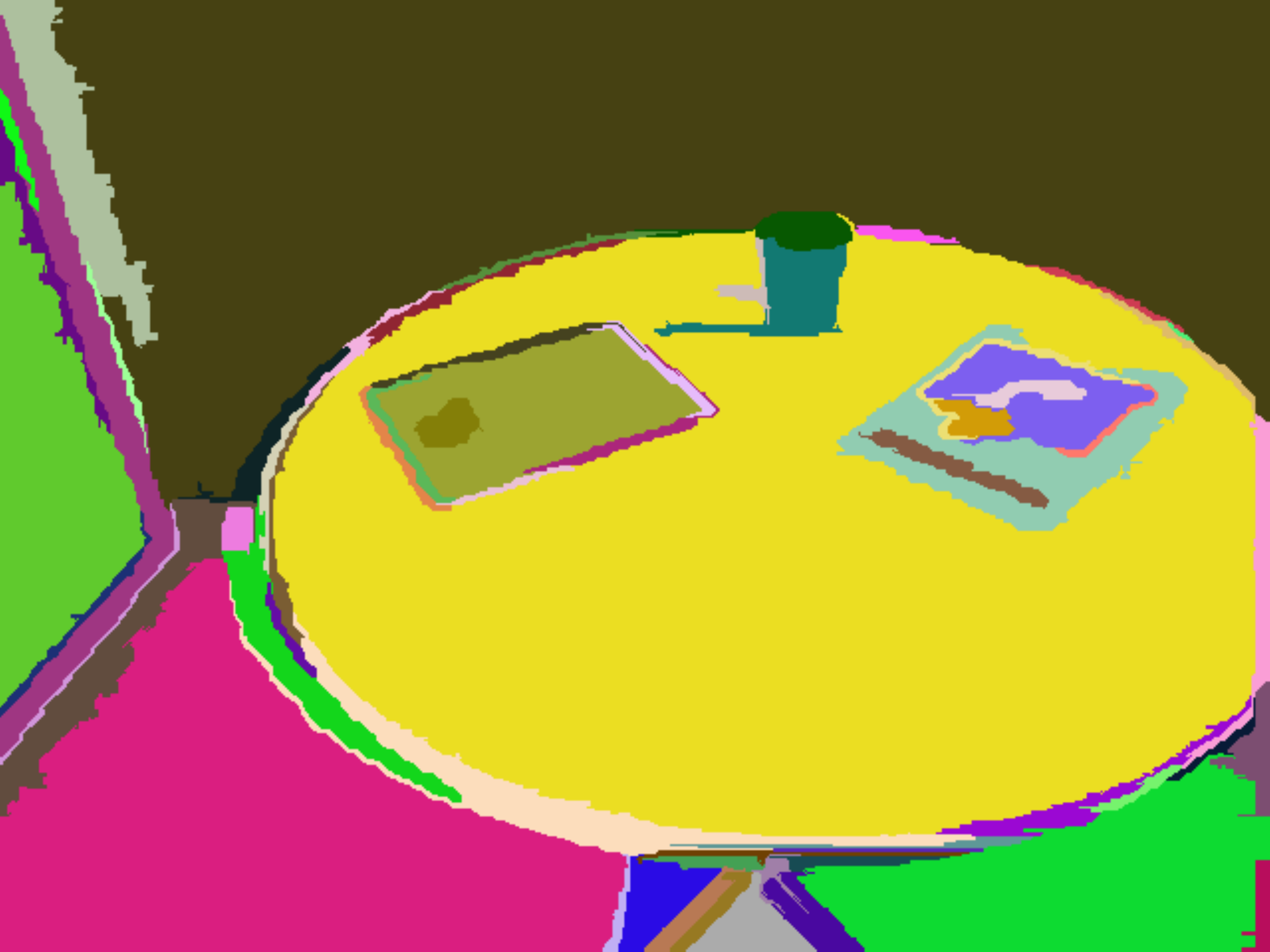} &
		\includegraphics[width=0.42\linewidth]{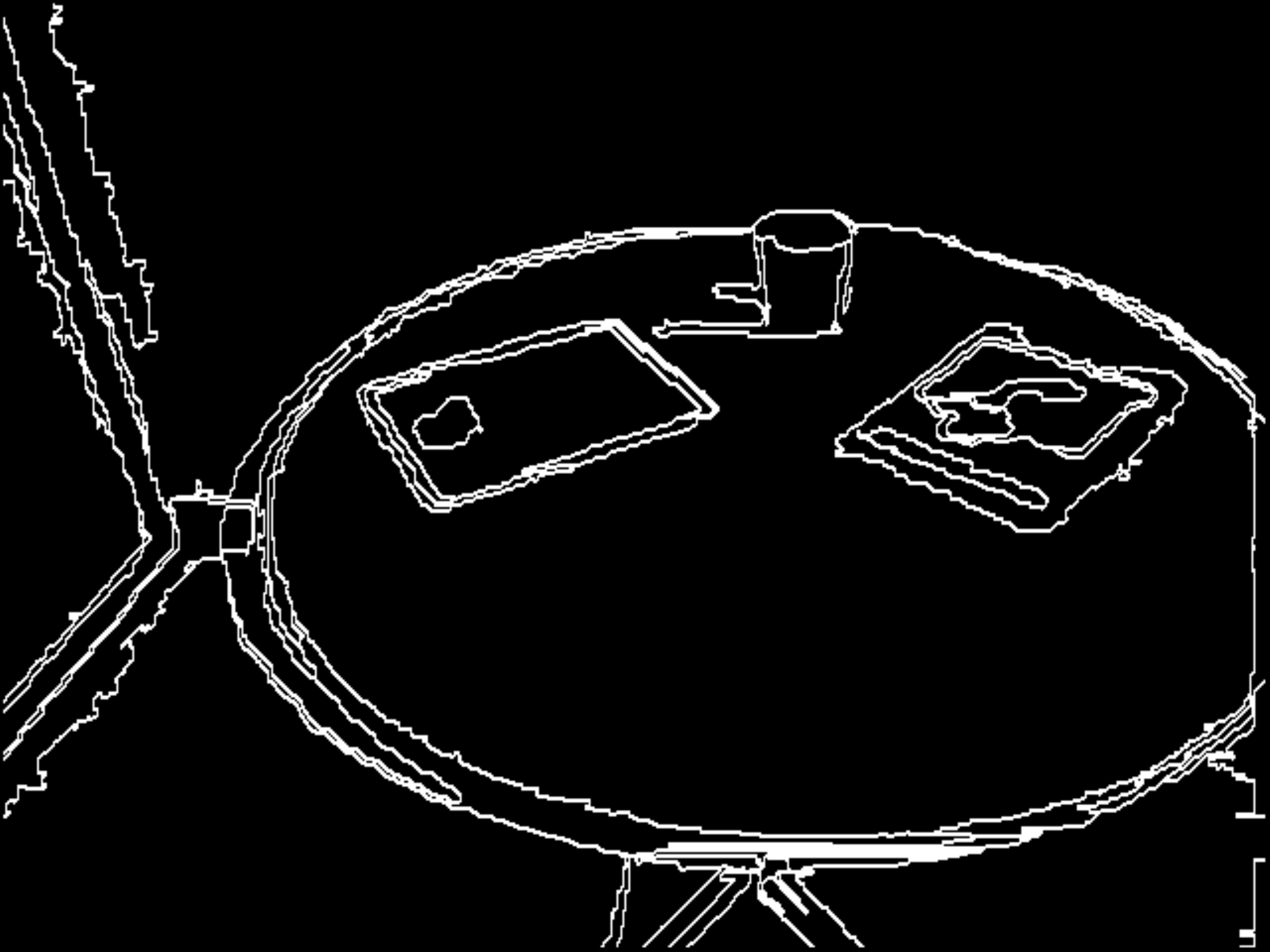}
	\end{tabular}		
	\caption{Top row:  original RGB image (depth image not shown), and ground truth object boundaries.  Middle three rows:  The output of the graph-based algorithm using a linear combination of color and depth, with $\alpha$ values of 0.1, 0.2, 0.4, 0.6, 0.8, and 0.9; none produces a correct result.  Bottom row:  The output of the initial over-segmentation step of our algorithm (left), as well as the boundaries of this output (right).}
	\label{fig_fig6}
\end{figure}

\noindent
As mentioned earlier, we were unable to find a value for $\alpha$ that enables the traditional graph-based algorithm in \cite{Felzenszwalb2004} to produce robust, accurate results across all sequences.  This is demonstrated in Figure~\ref{fig_fig6}, where the output for various values of $\alpha$ are shown, none of which clearly delineates the objects properly. The linear combination never properly segments the table, table legs, cup, magazine, and magazine colored features in Figure~\ref{fig_fig6}, whereas our over-segmentation does. In other words, the multistage segmentation combination of first depth and then color segmentation provides better results than any linear combination of color and depth features as the primate studies from \cite{livingstone1988segregation} suggested.

\begin{figure}
	\centering
	\begin{tabular}{cc}
		\includegraphics[width=0.5\linewidth]{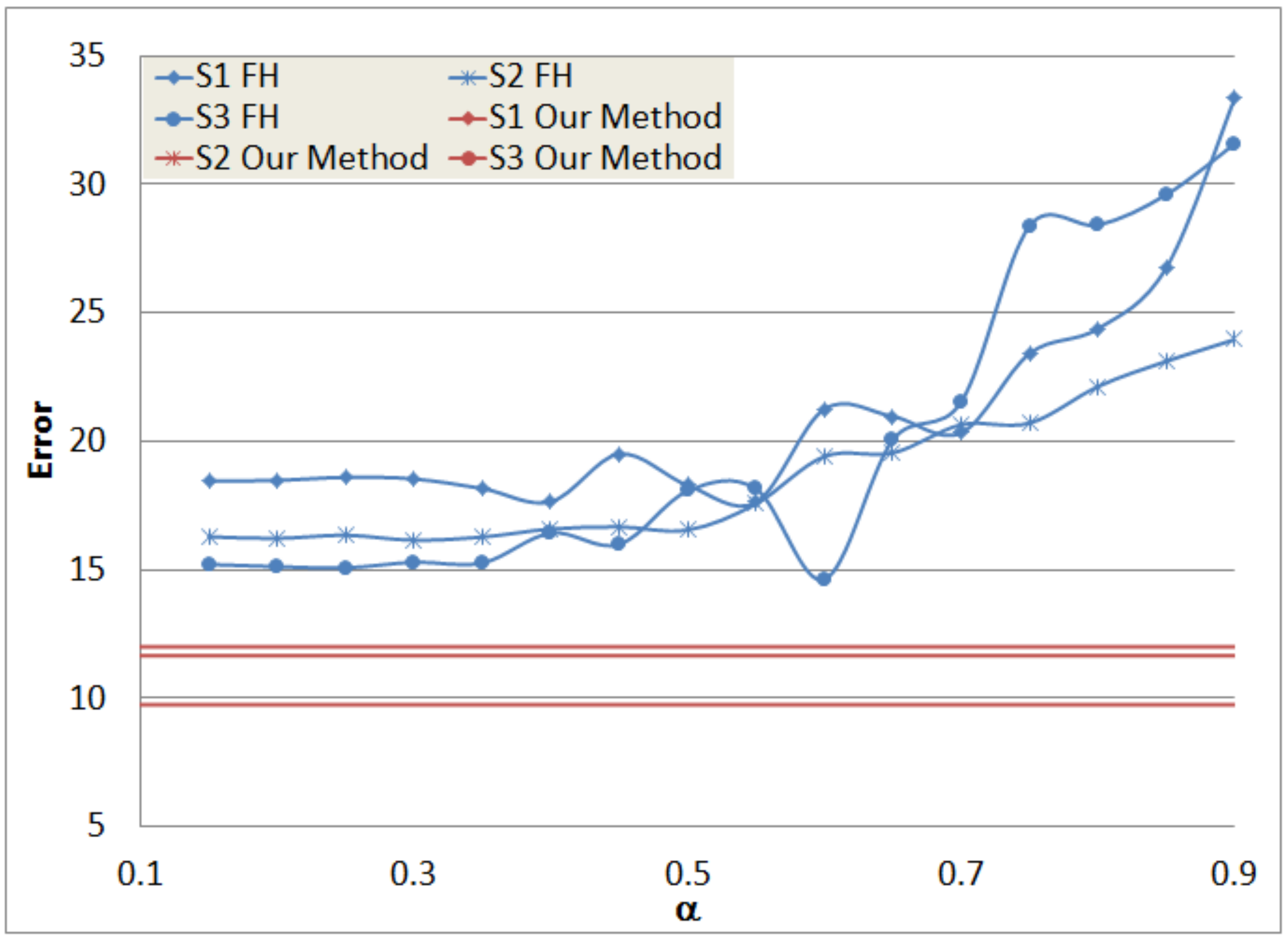} &
		\includegraphics[width=0.5\linewidth]{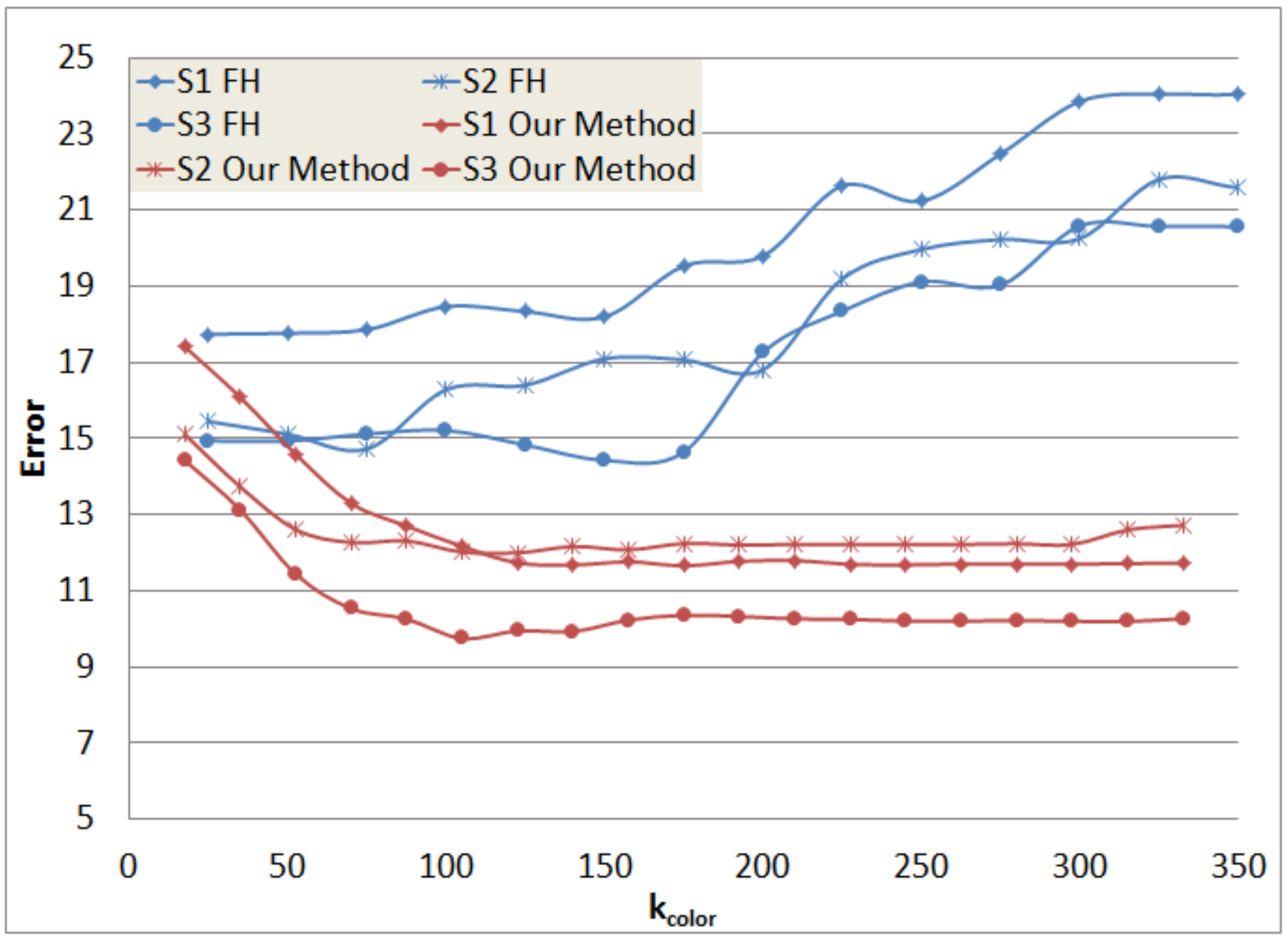}
	\end{tabular}
	\caption{Left:  The error based on the choice of $\alpha$ on the output of the linear combination algorithm on three sequences (blue lines), as well as the output of our hierarchical algorithm (red lines), which is not dependent on $\alpha$  
	Right:  Effect of varying $k_{color}$ on the output of our initial over-segmentation step, as well as the value of $k$ on the output of~\cite{Felzenszwalb2004} (with the best $\alpha$ possible.}
	\label{fig_fig7}
\end{figure}

\begin{figure*} [t]
    \centering   
    \begin{tabular}{rccccc}
      \textbf{(A)} & \includegraphics[width=0.16\linewidth]{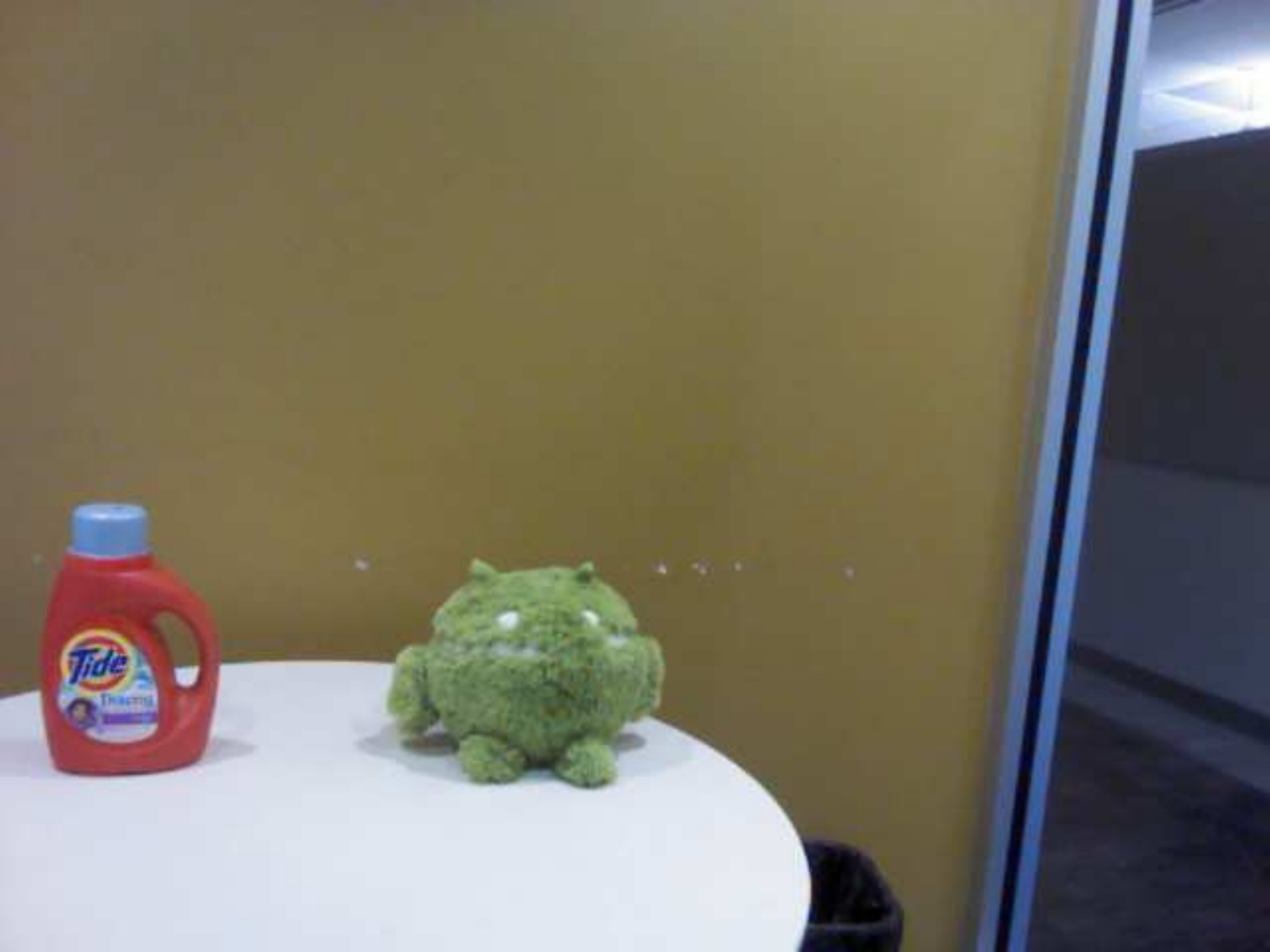} &
      \includegraphics[width=0.16\linewidth]{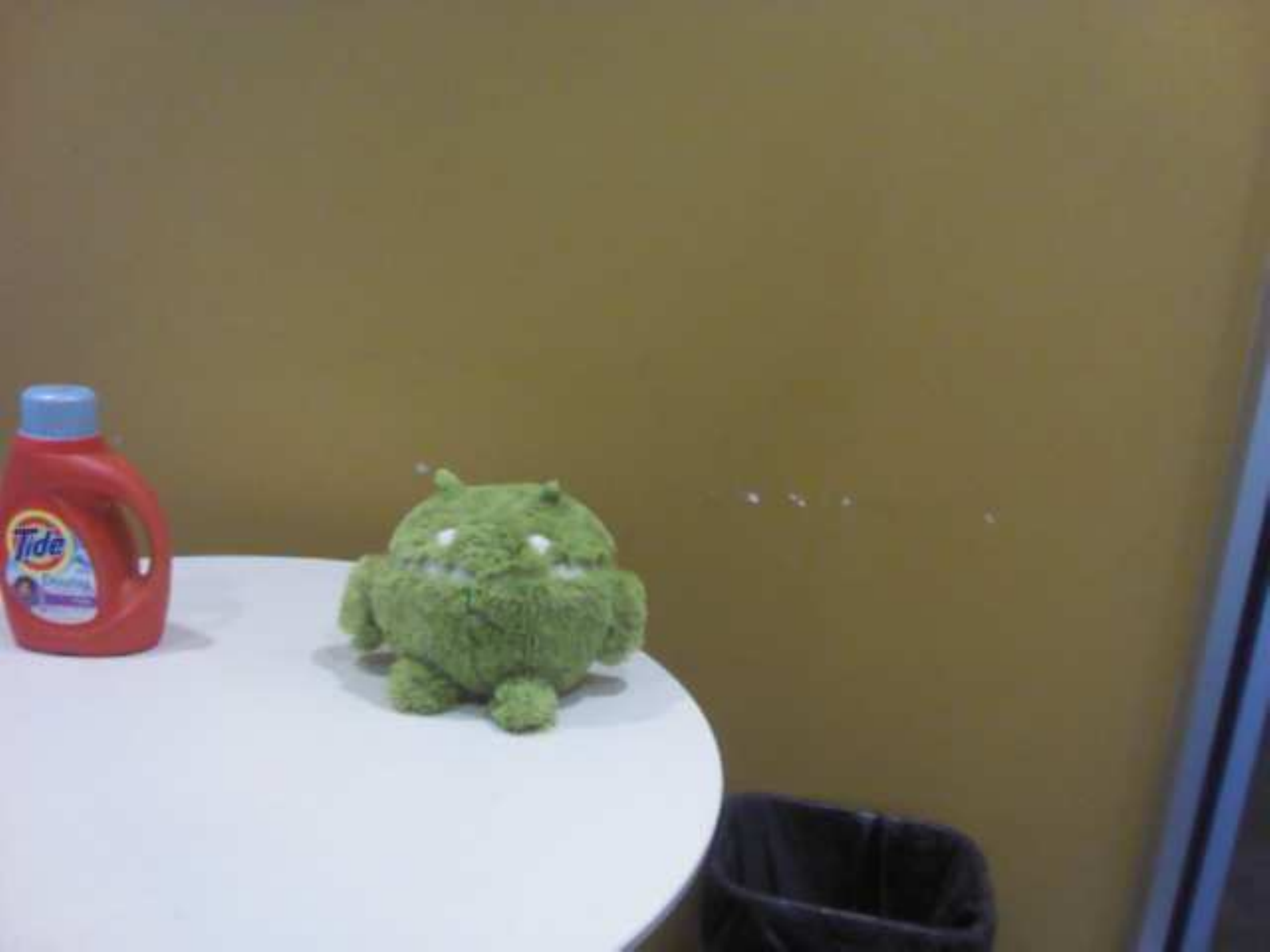} &
      \includegraphics[width=0.16\linewidth]{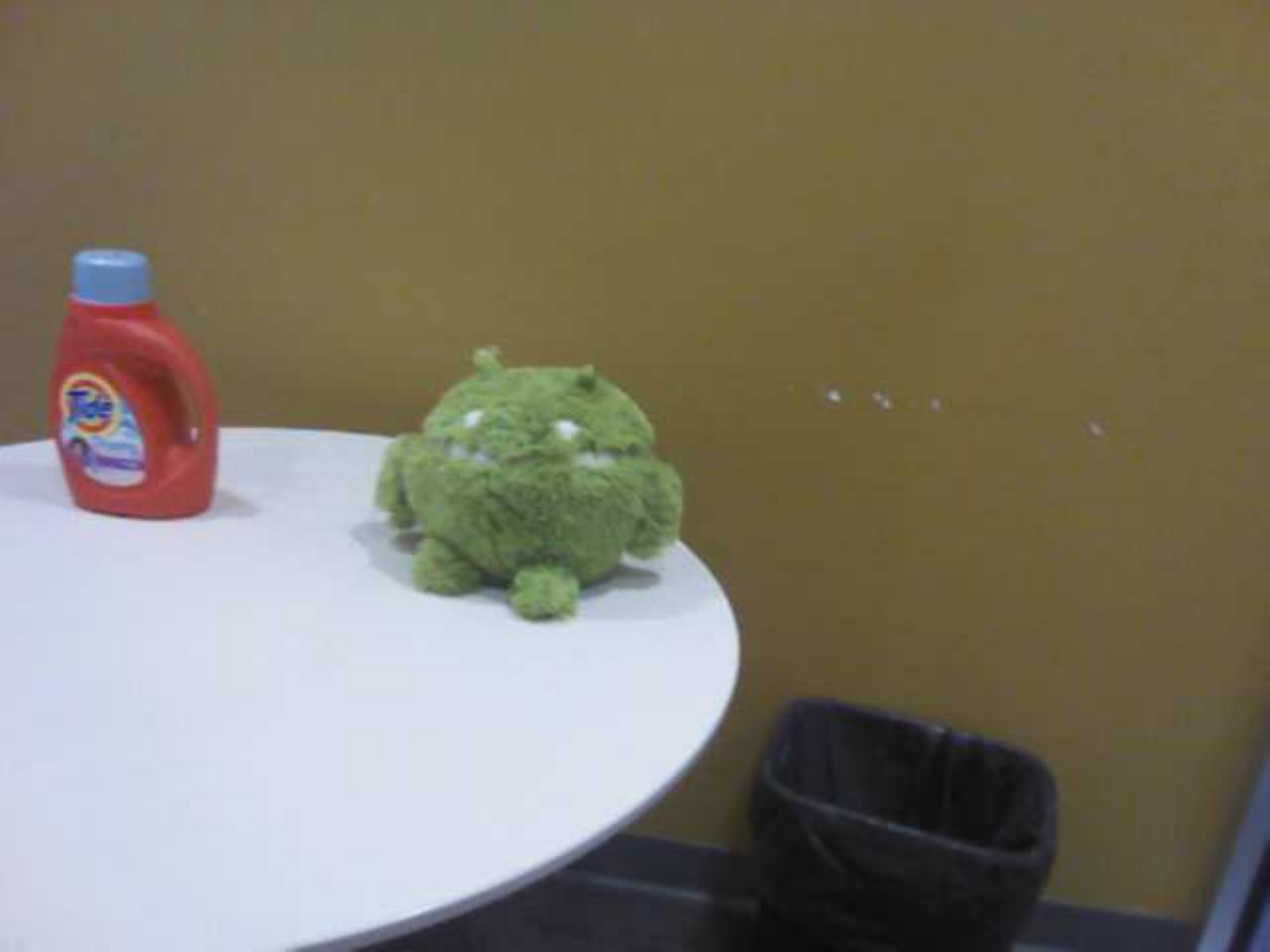} &
      \includegraphics[width=0.16\linewidth]{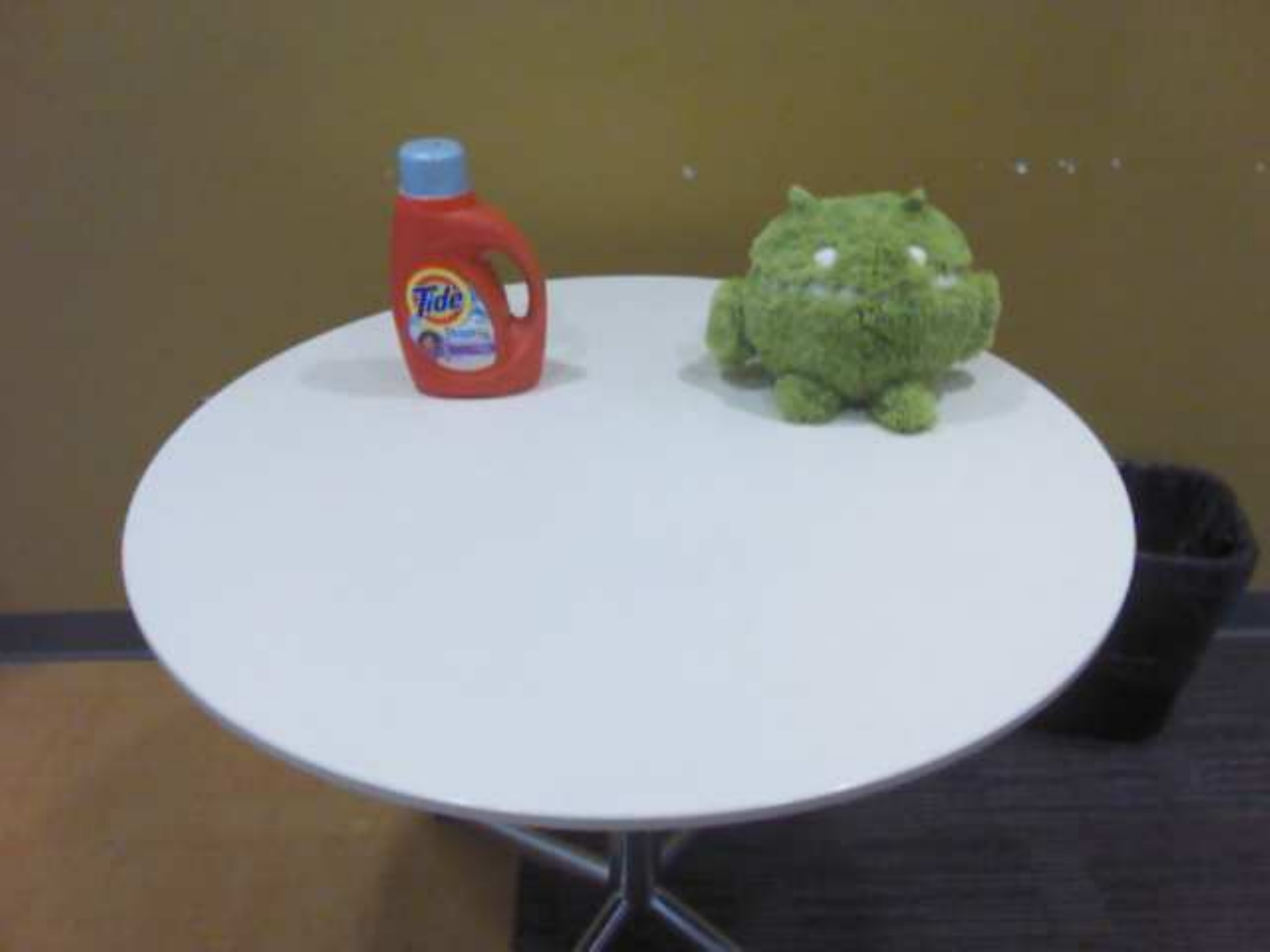} &
			\includegraphics[width=0.16\linewidth]{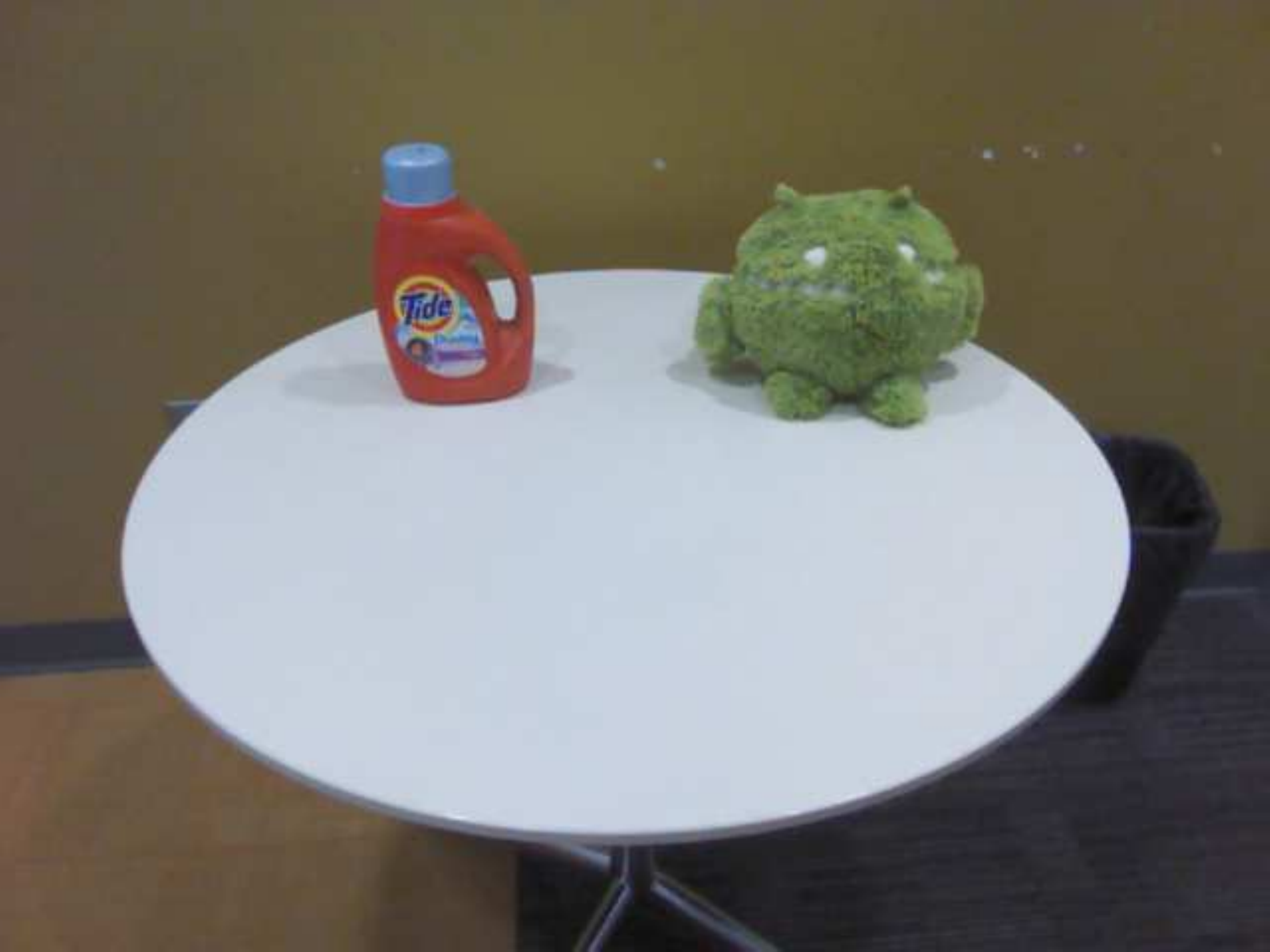}
    \end{tabular}
		\begin{tabular}{rccccc}
      \textbf{(B)} & \includegraphics[width=0.16\linewidth]{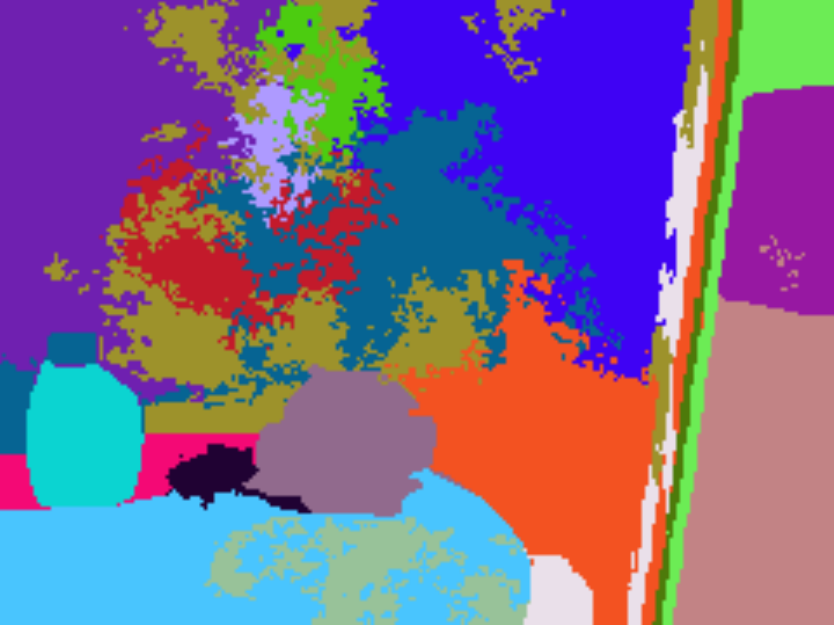} &
      \includegraphics[width=0.16\linewidth]{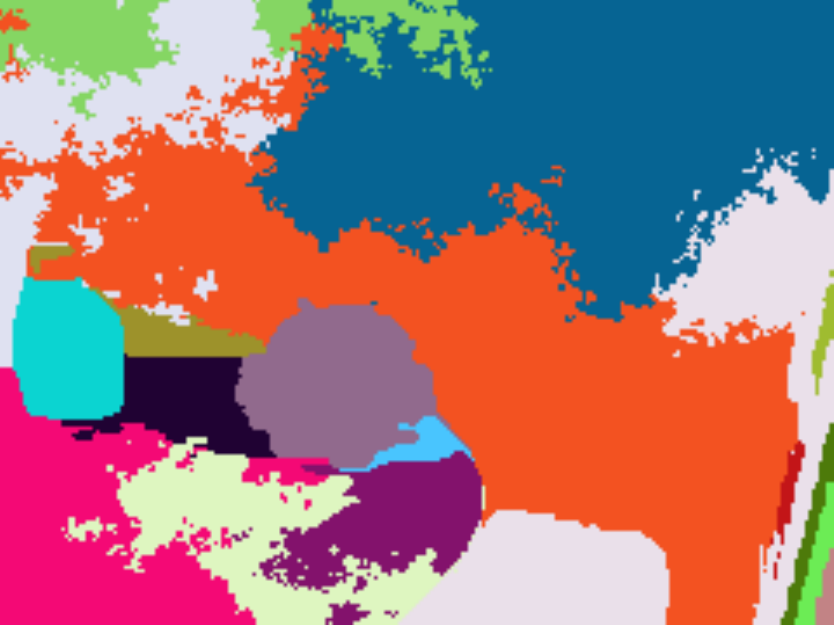} &
      \includegraphics[width=0.16\linewidth]{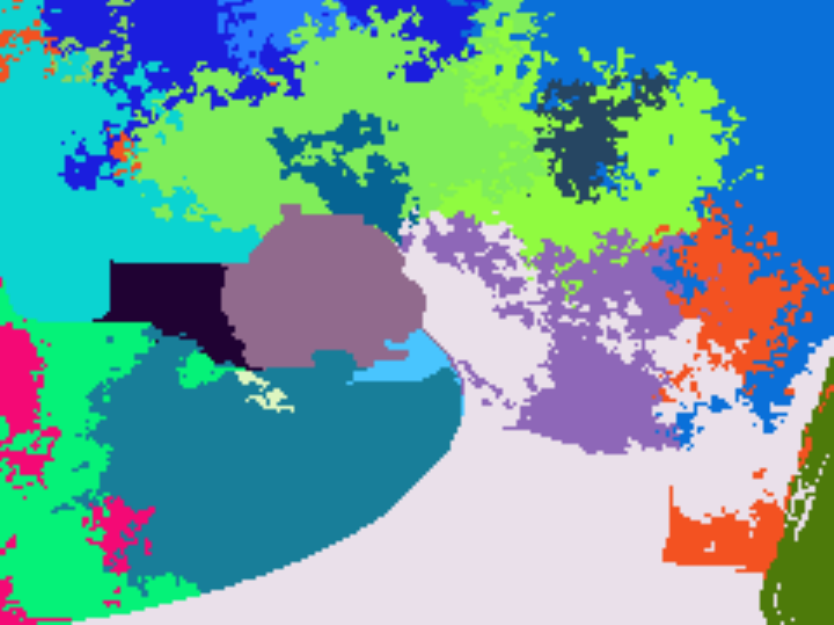} &
      \includegraphics[width=0.16\linewidth]{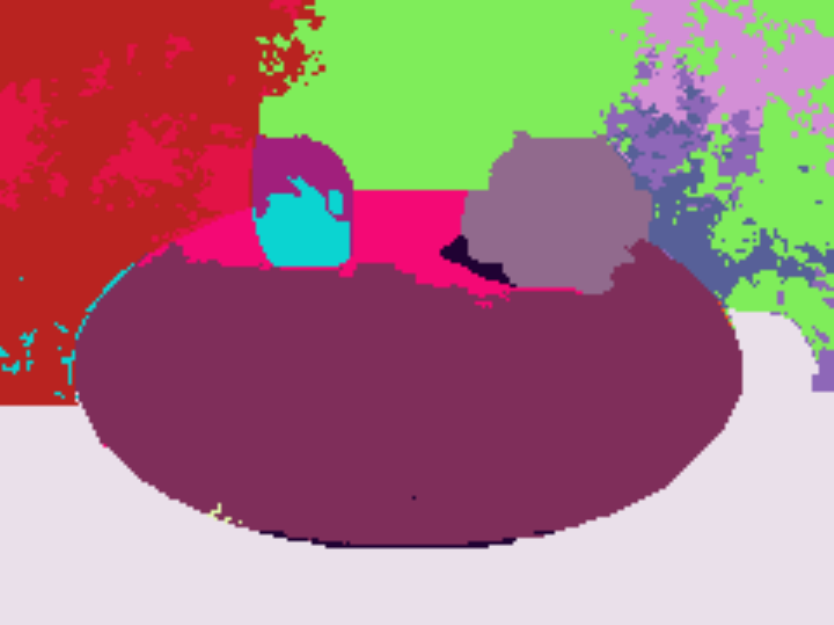} &
			\includegraphics[width=0.16\linewidth]{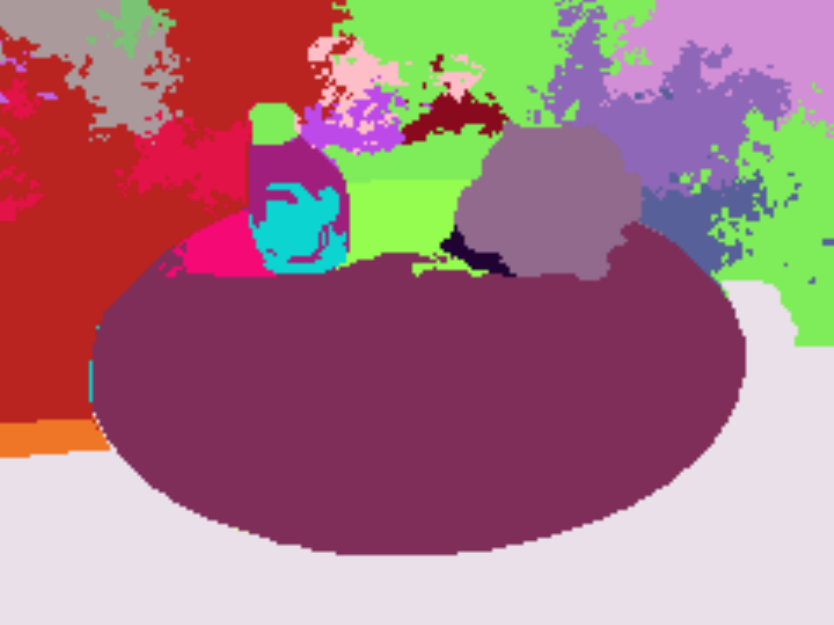}
    \end{tabular}
		\begin{tabular}{rccccc}
      \textbf{(C)} & \includegraphics[width=0.16\linewidth]{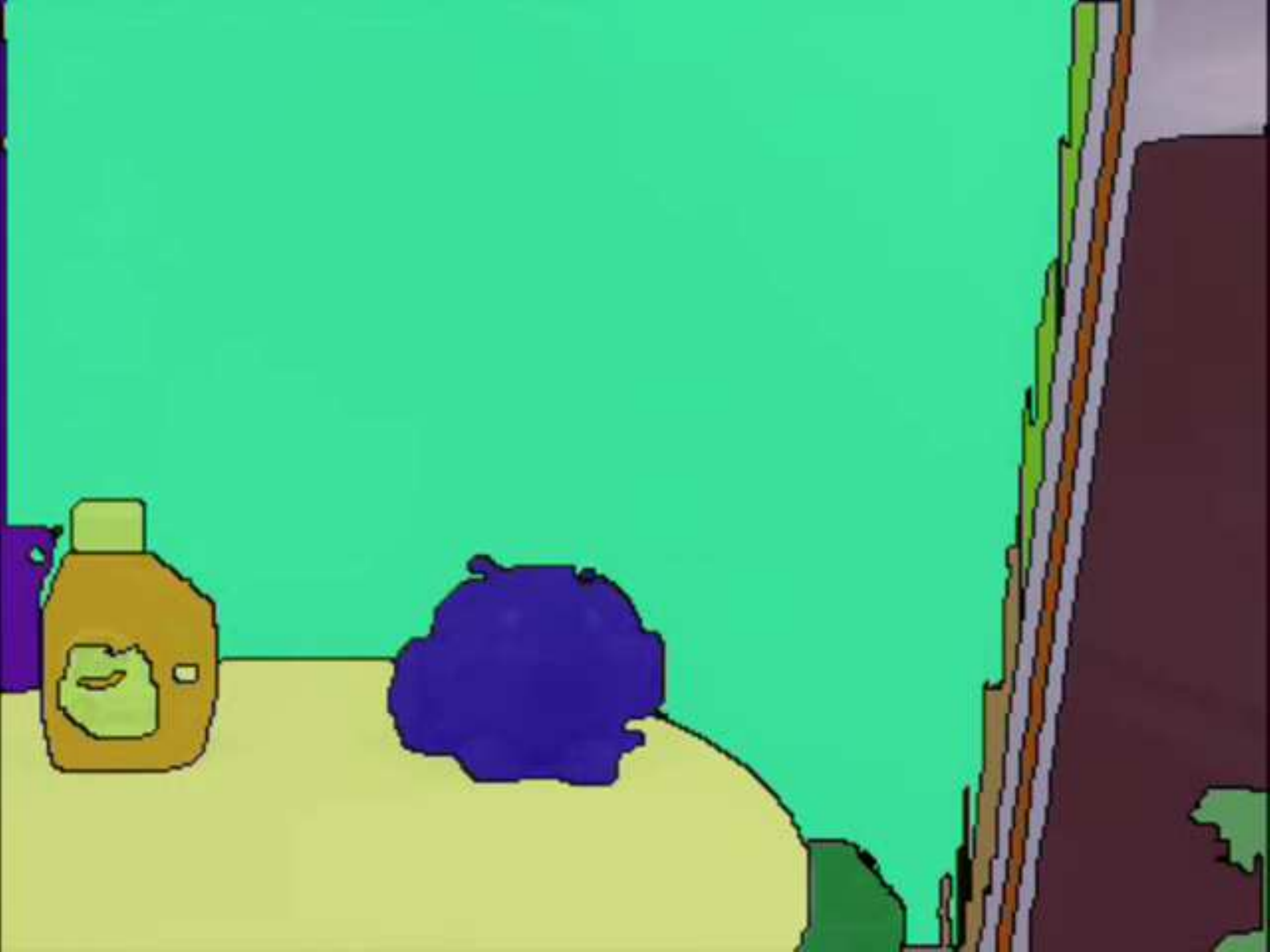} &
      \includegraphics[width=0.16\linewidth]{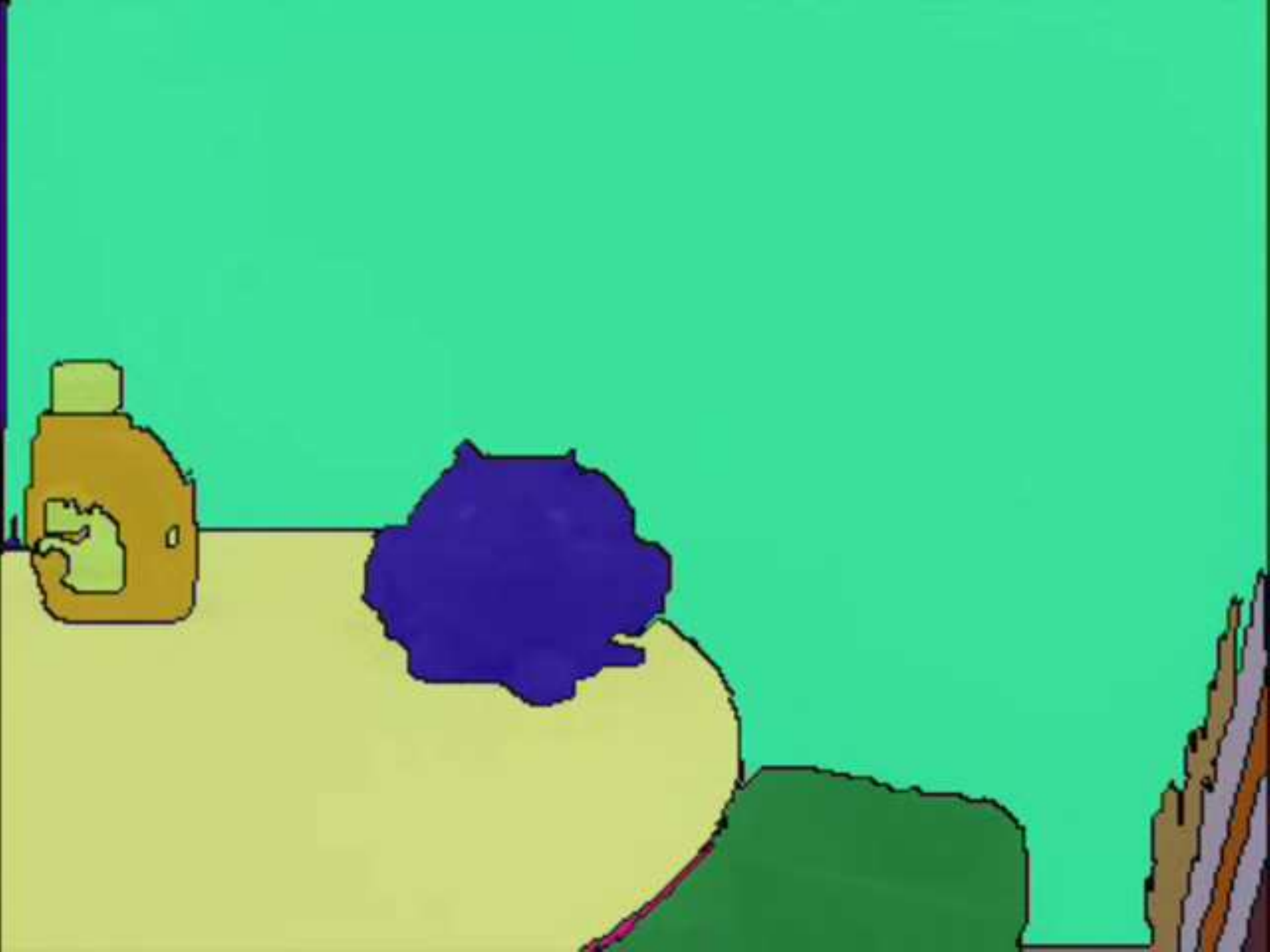} &
      \includegraphics[width=0.16\linewidth]{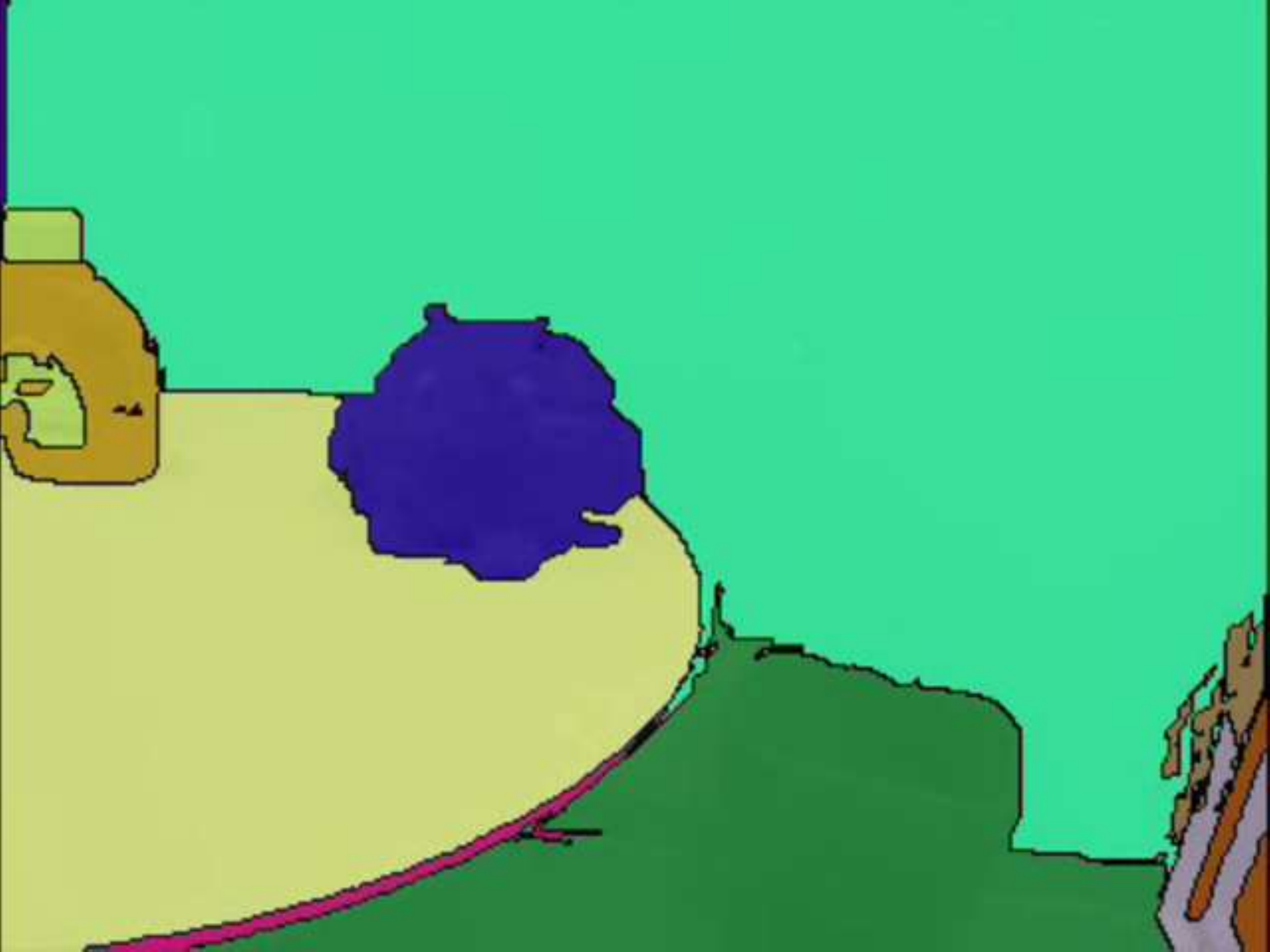} &
      \includegraphics[width=0.16\linewidth]{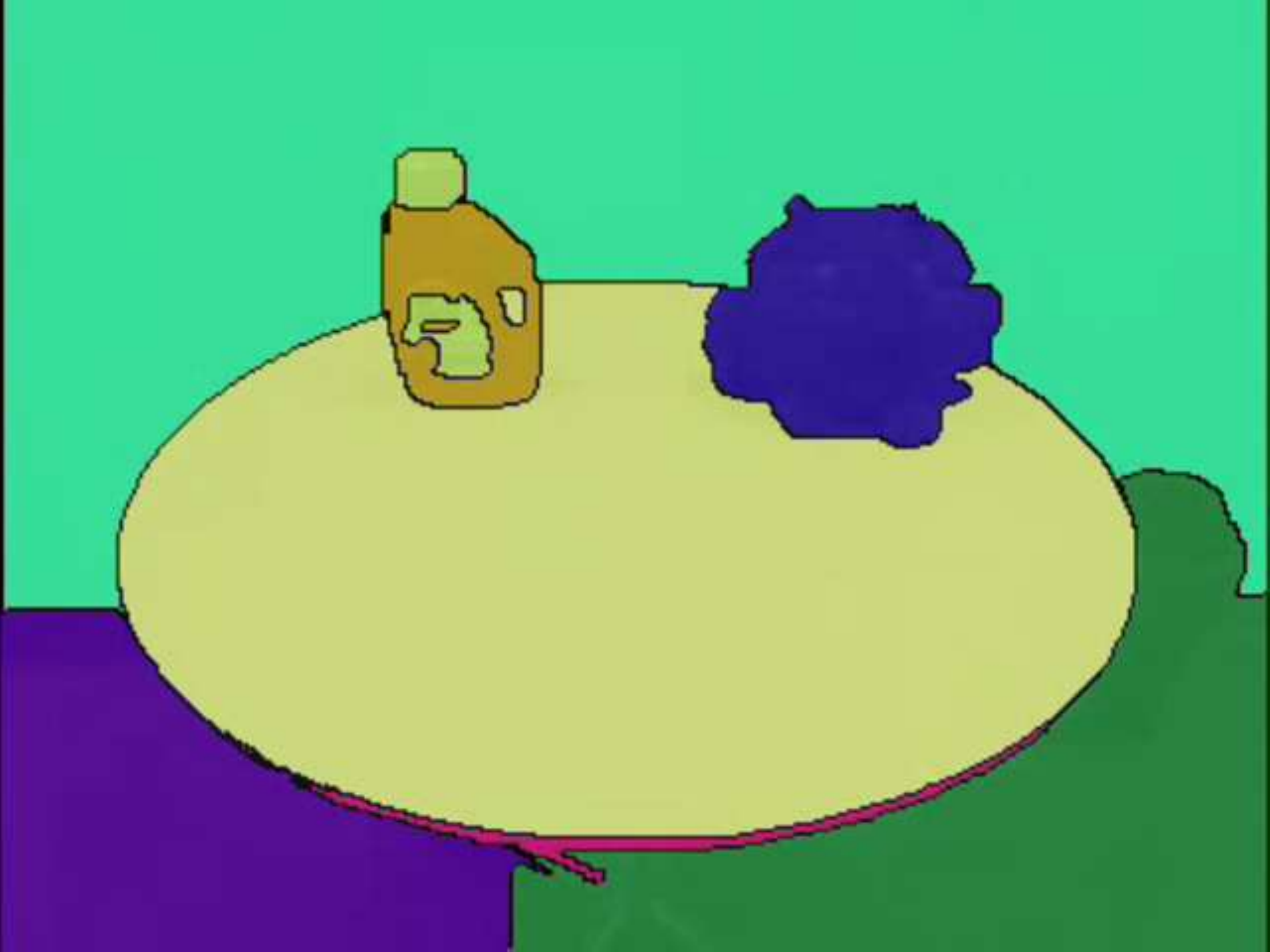} &
			\includegraphics[width=0.16\linewidth]{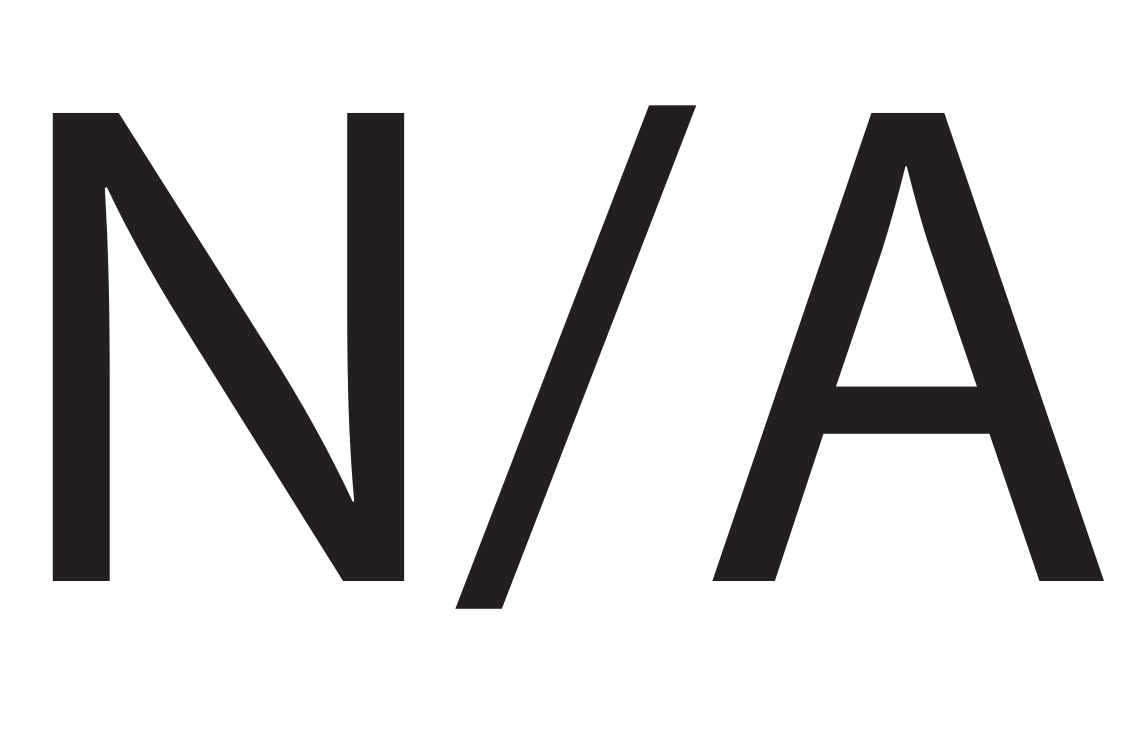}
    \end{tabular}
		\begin{tabular}{rccccc}
      \textbf{(D)} & \includegraphics[width=0.16\linewidth]{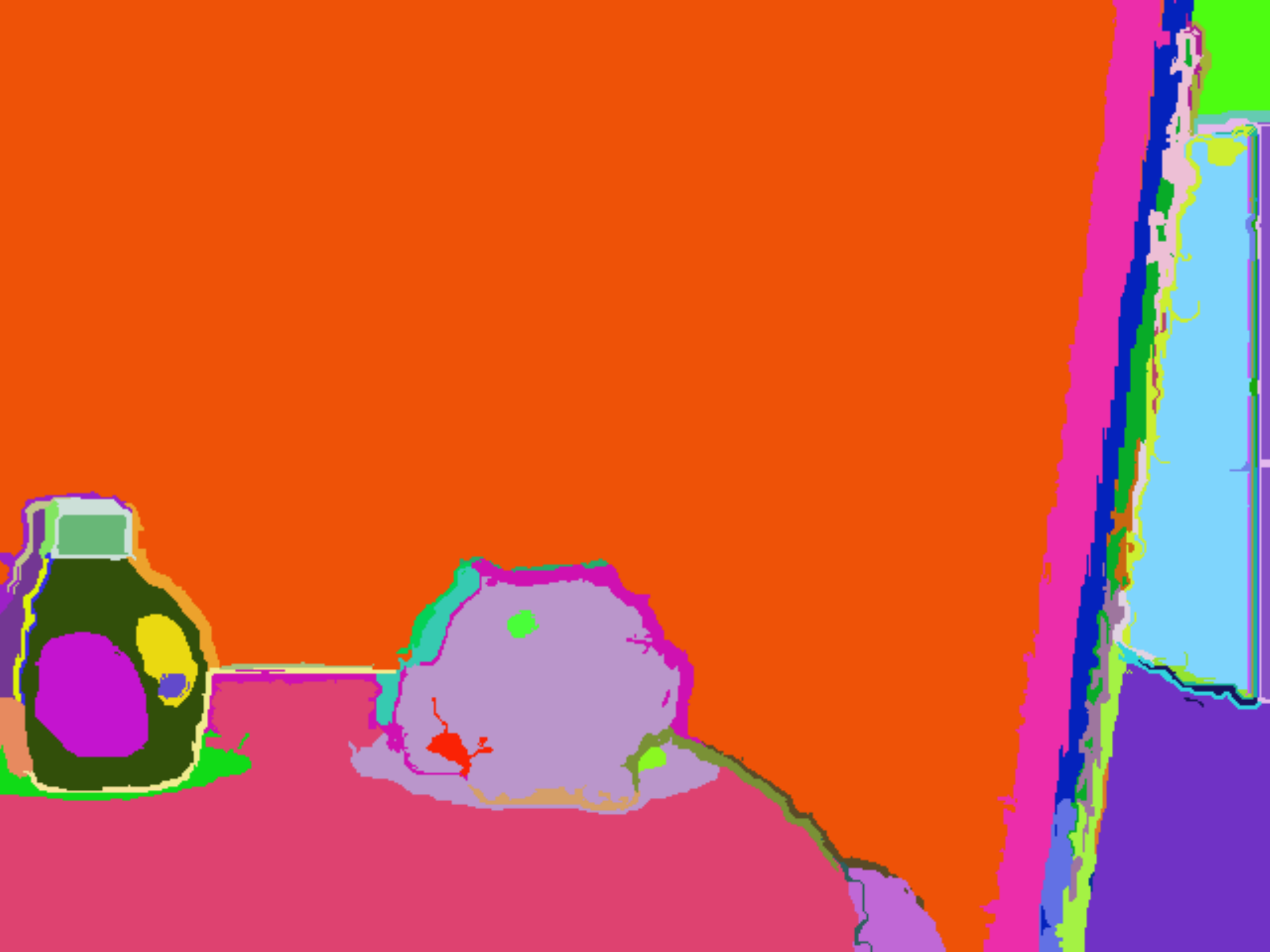} &
      \includegraphics[width=0.16\linewidth]{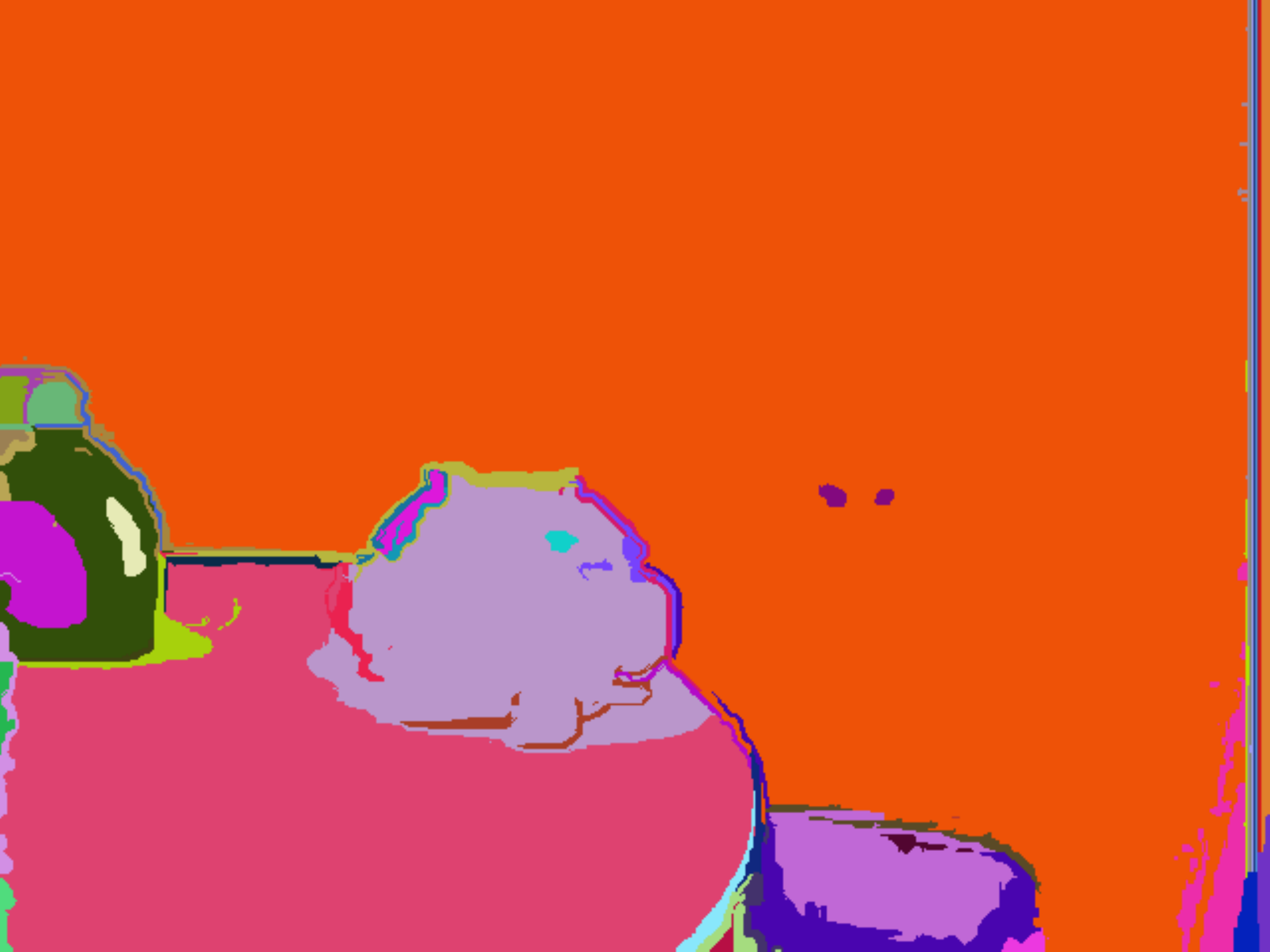} &
      \includegraphics[width=0.16\linewidth]{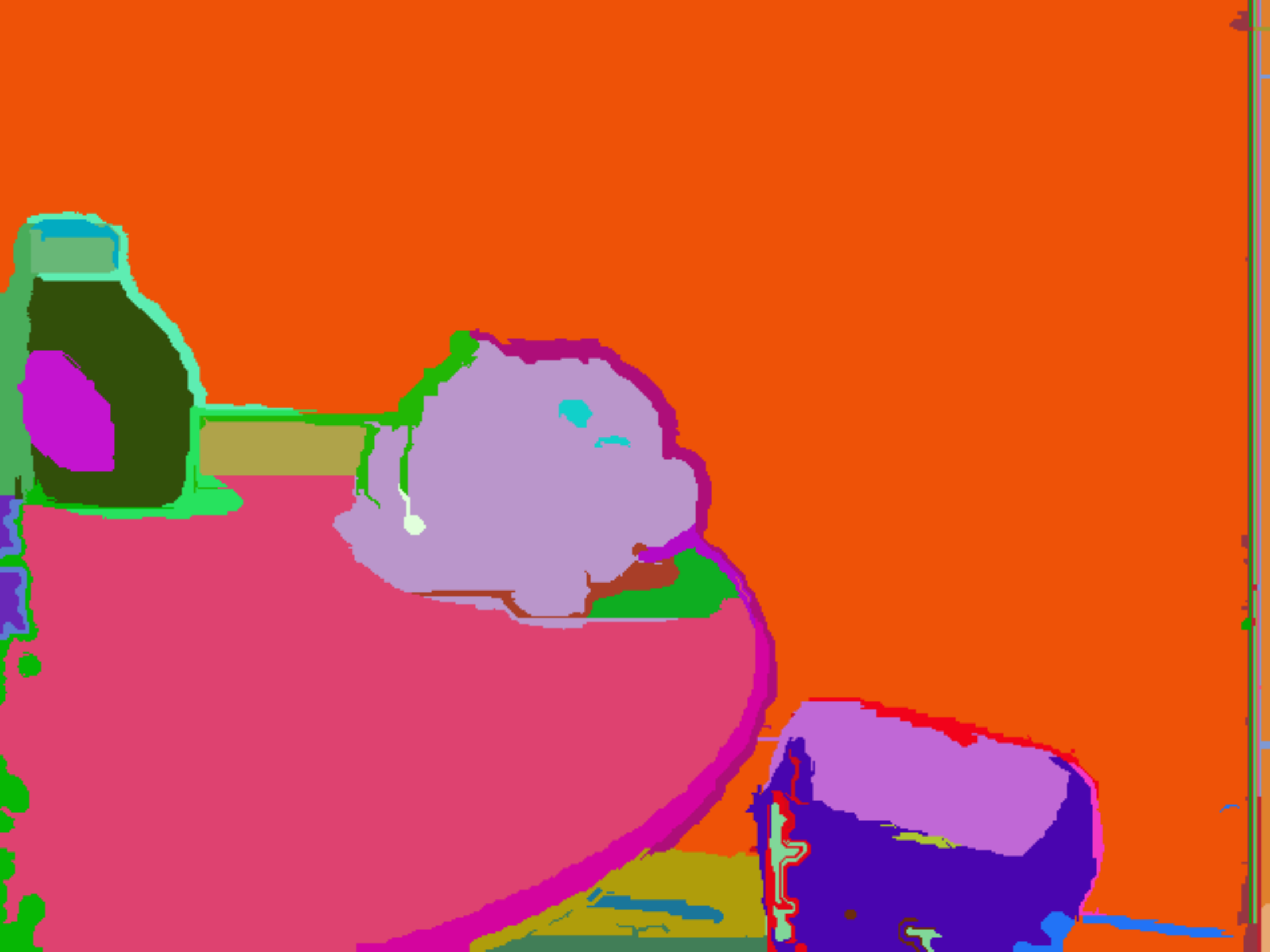} &
      \includegraphics[width=0.16\linewidth]{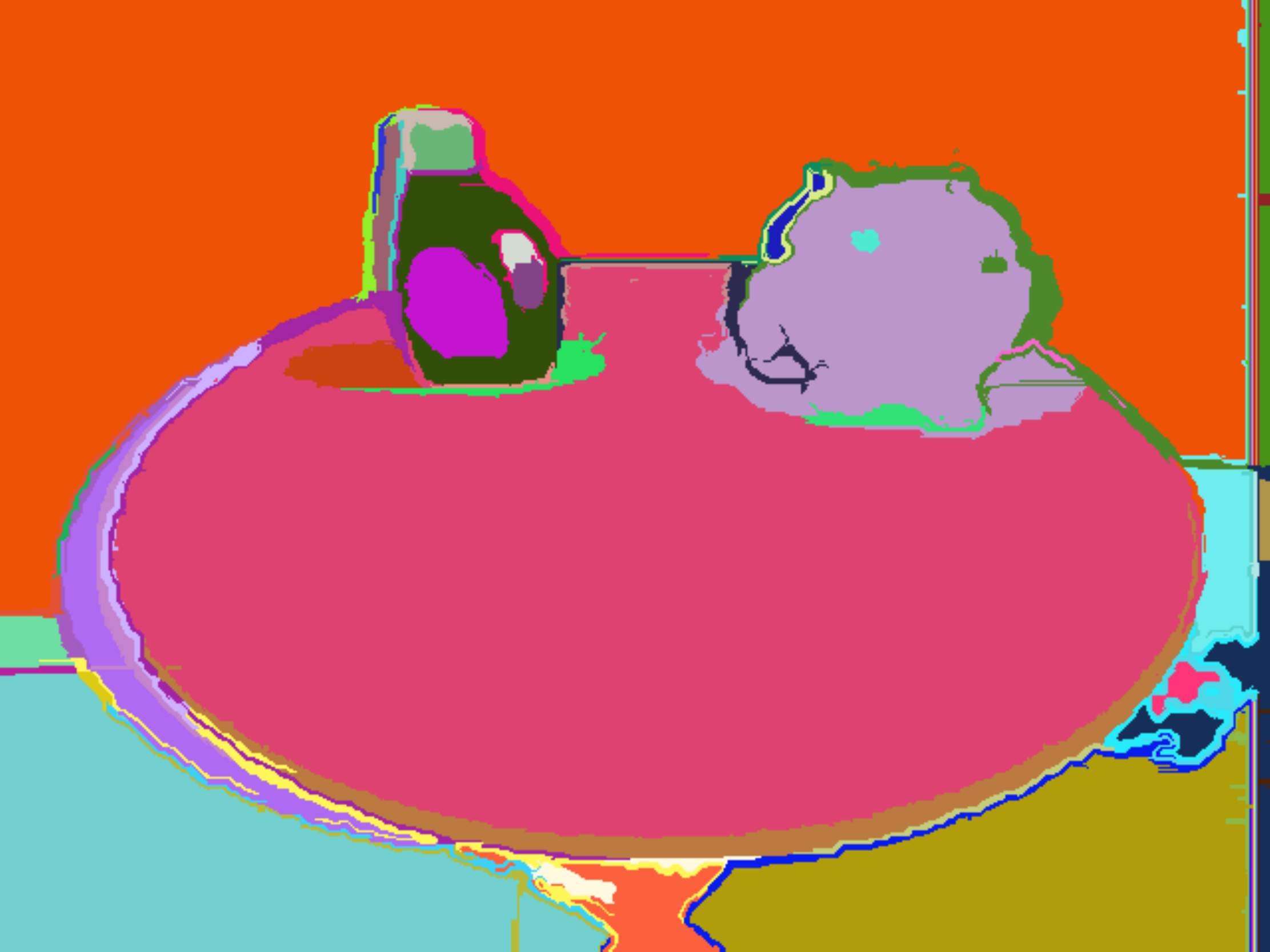} &
			\includegraphics[width=0.16\linewidth]{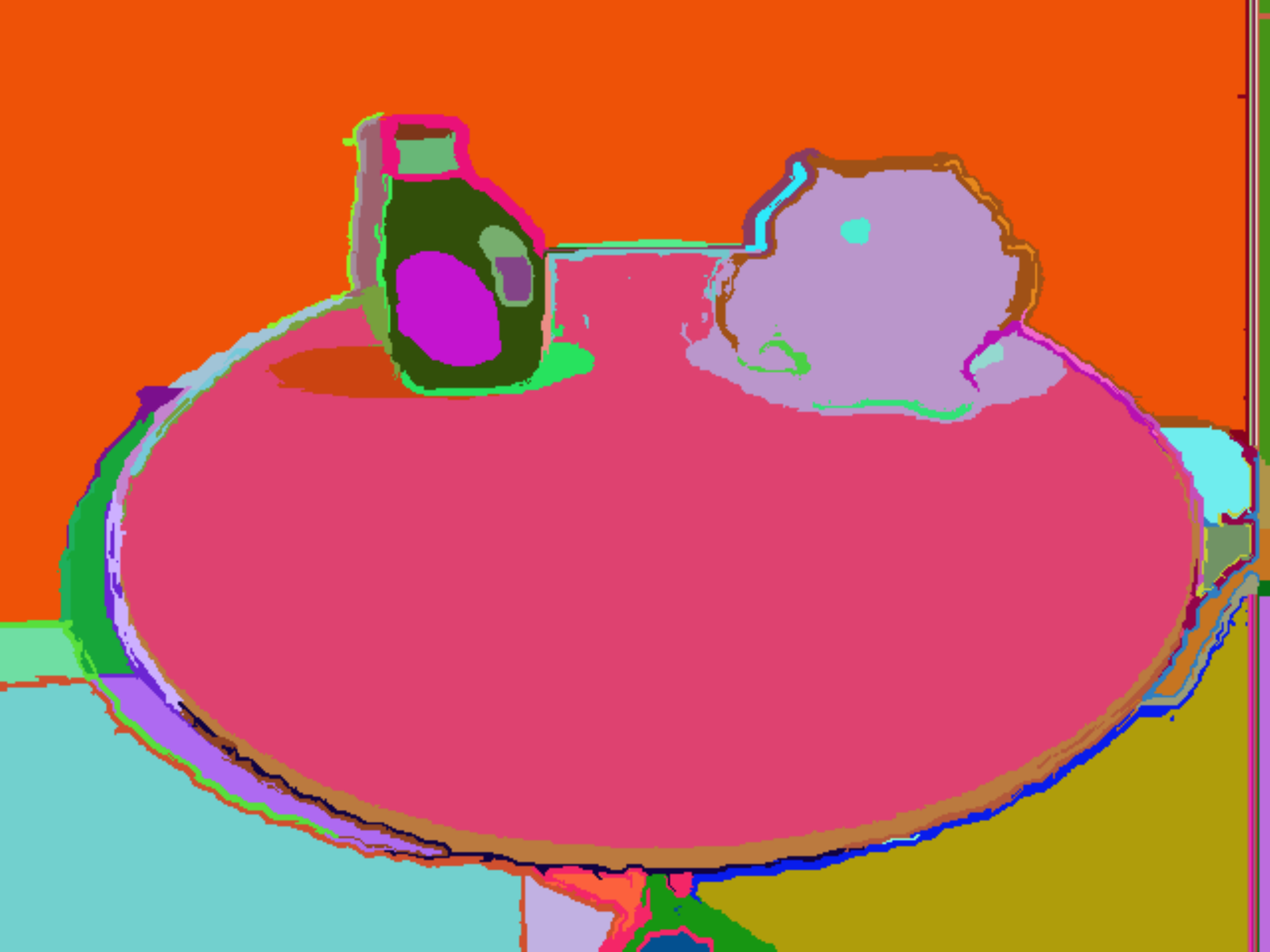}
    \end{tabular}
		\caption{Sequence 1, (A) : RGB image, (B): Results from \protect\cite{Xuy2012}, (C): Results from \protect\cite{Grundmann2010} (N/A is used when the frame is beyond the length it can run), (D): Our Results}
    \label{fig_fig13a}
\end{figure*}

\begin{figure*}  [t]
 \centering   
    \begin{tabular}{rccccc}
      \textbf{(A)} & \includegraphics[width=0.16\linewidth]{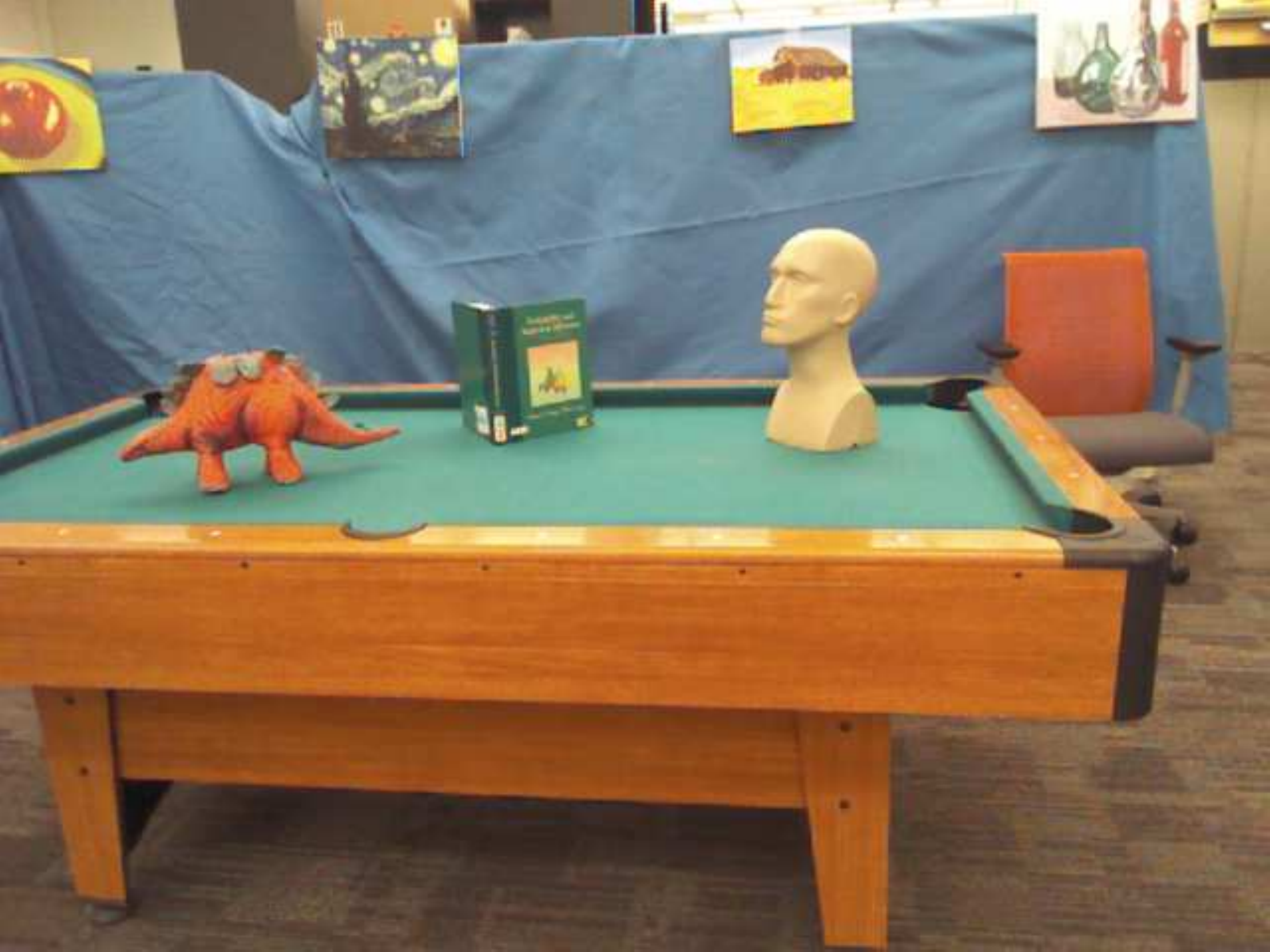} &
      \includegraphics[width=0.16\linewidth]{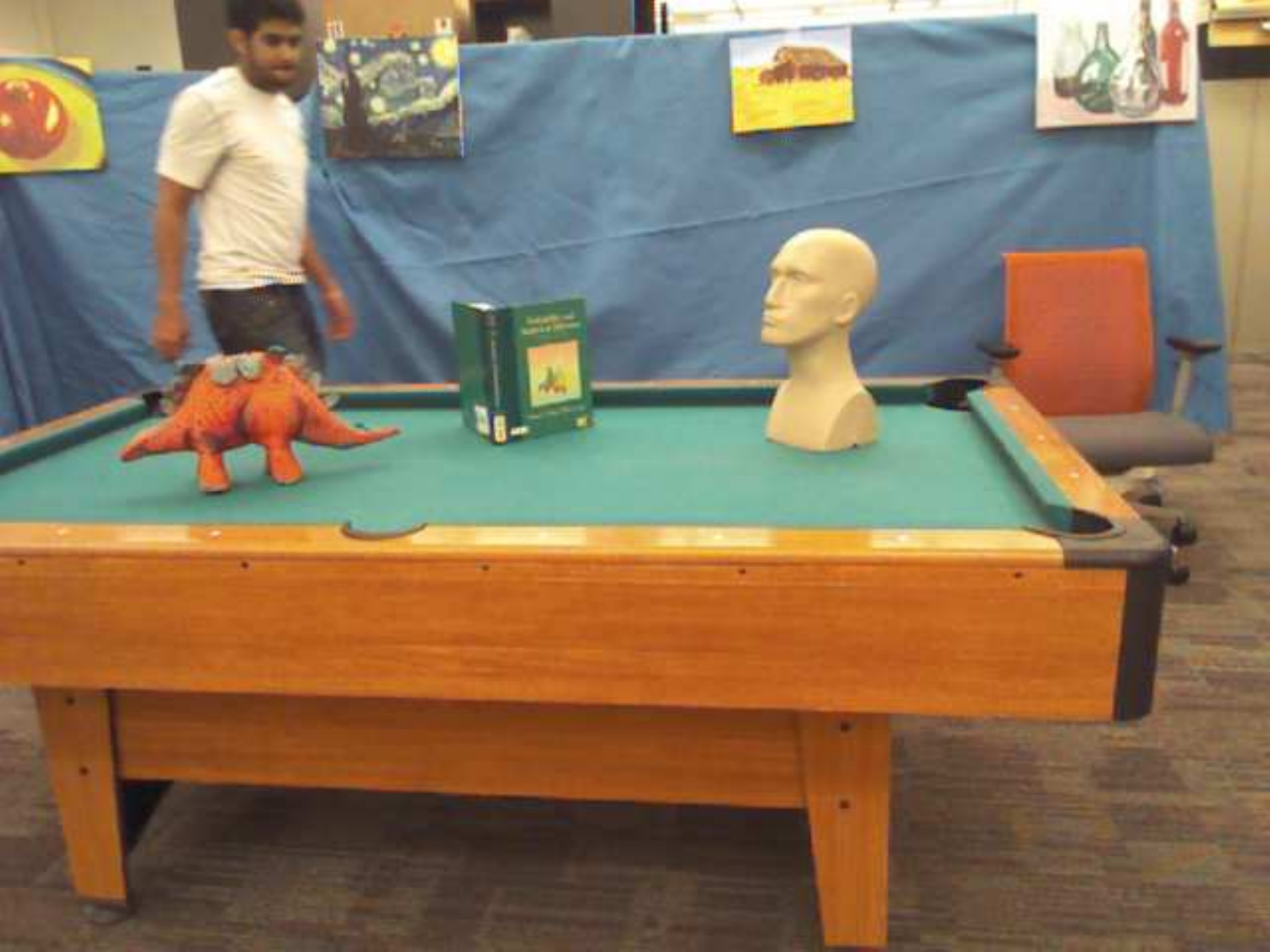} &
      \includegraphics[width=0.16\linewidth]{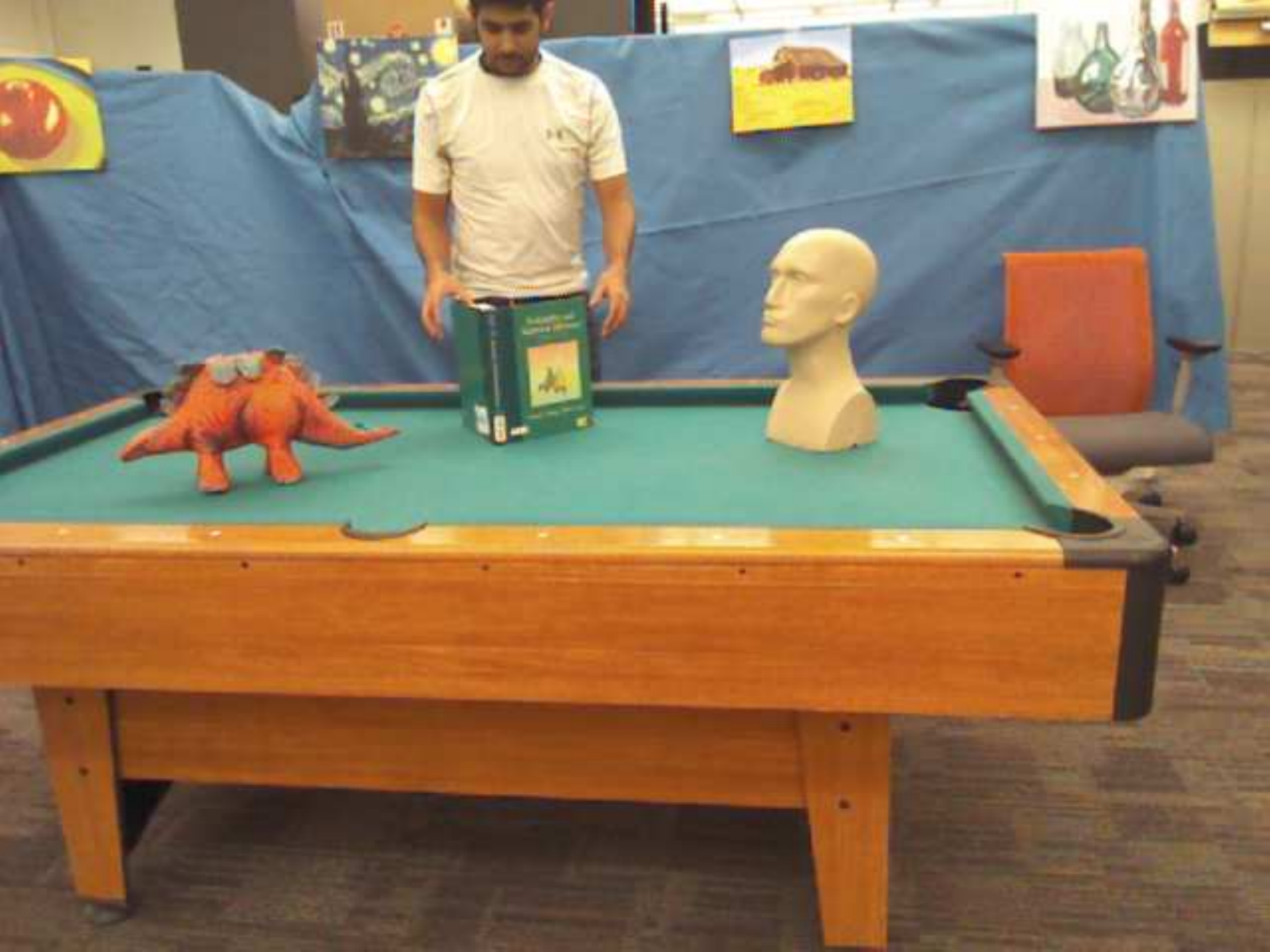} &
      \includegraphics[width=0.16\linewidth]{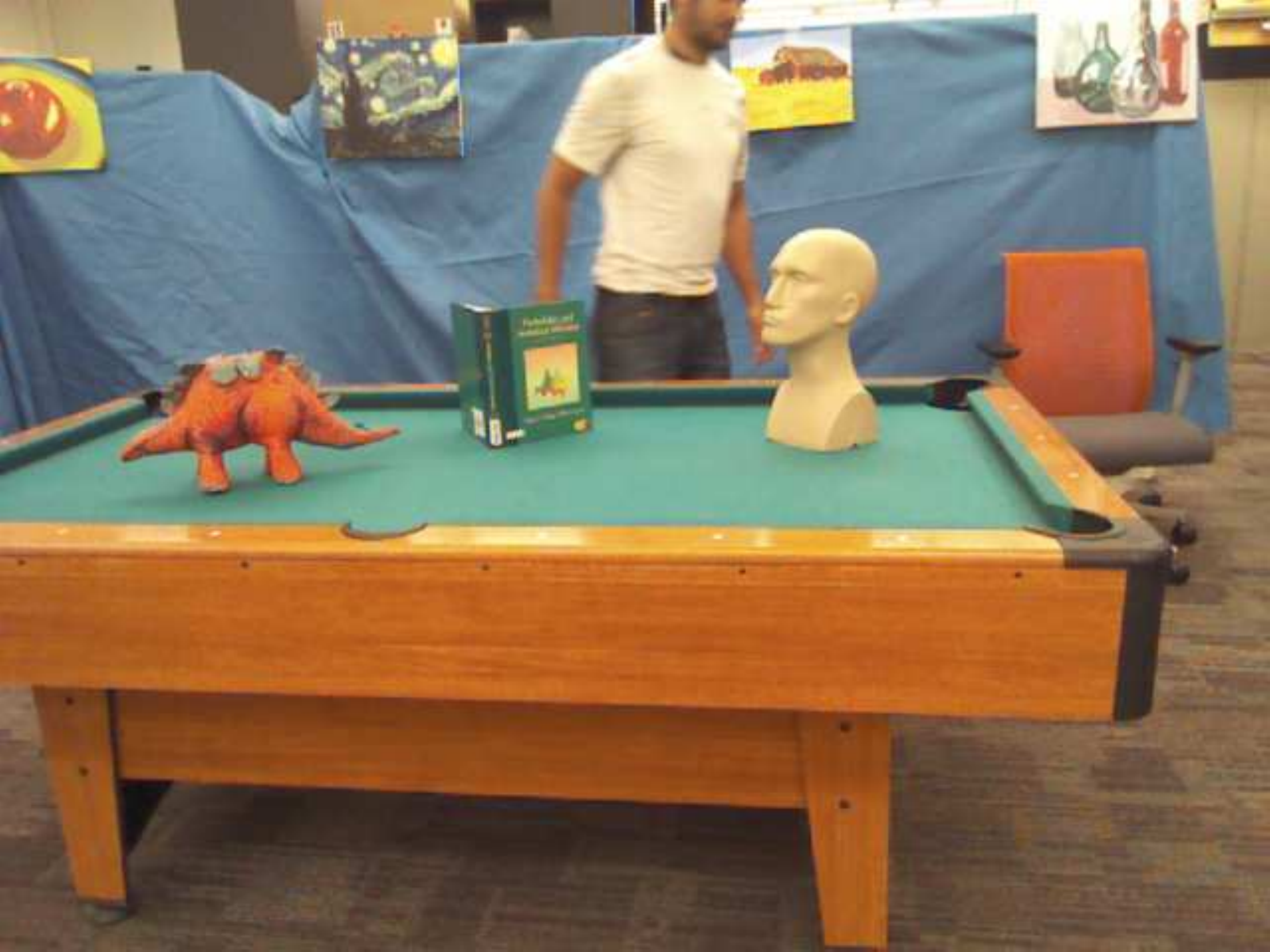} &
			\includegraphics[width=0.16\linewidth]{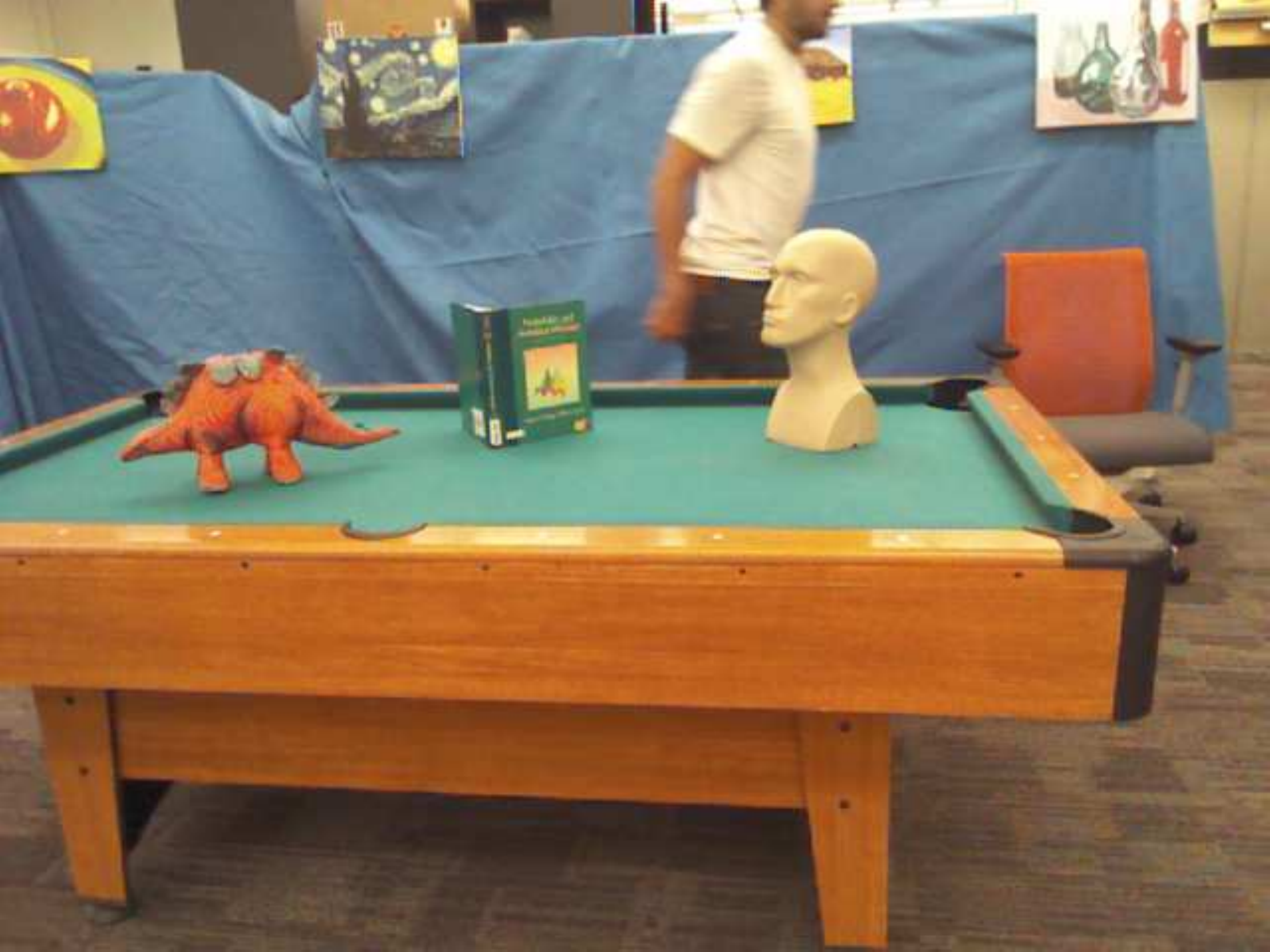}
    \end{tabular}
		\begin{tabular}{rccccc}
      \textbf{(B)} & \includegraphics[width=0.16\linewidth]{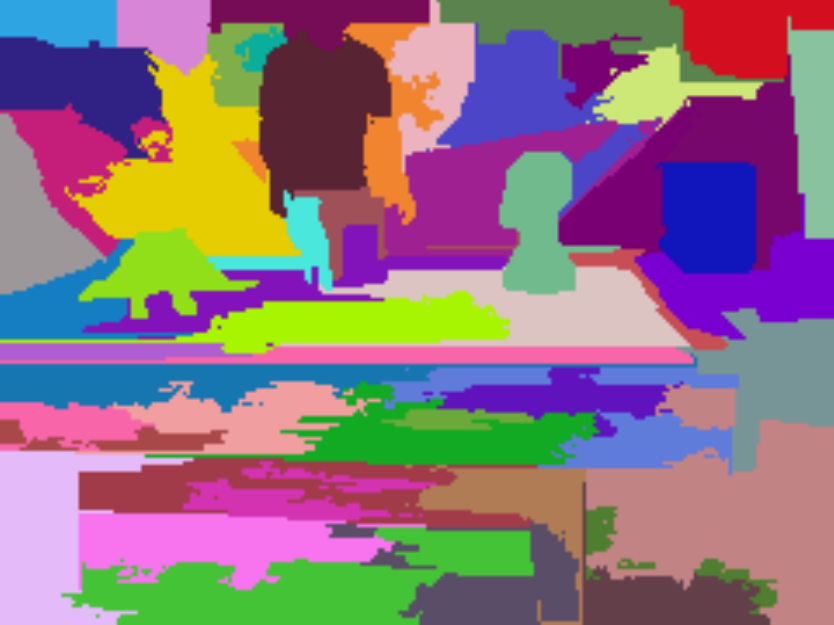} &
      \includegraphics[width=0.16\linewidth]{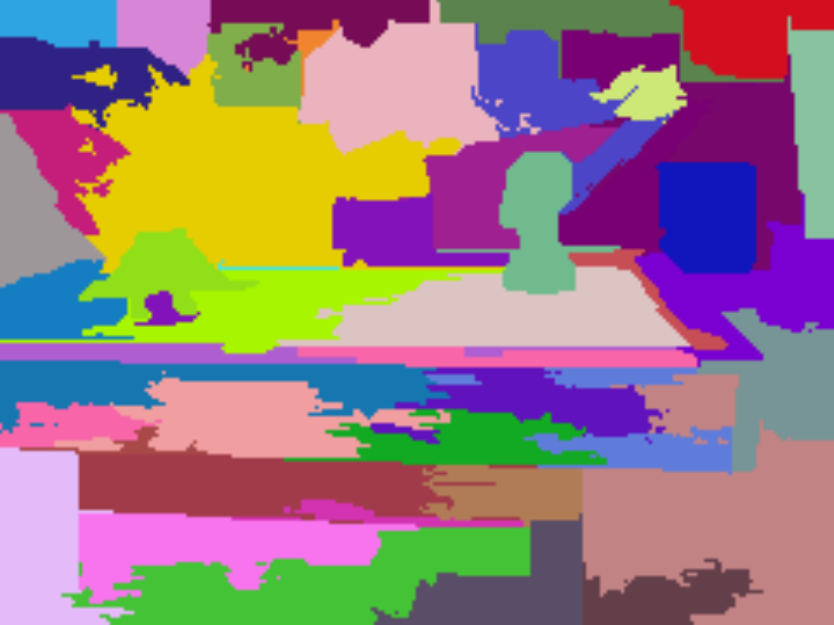} &
      \includegraphics[width=0.16\linewidth]{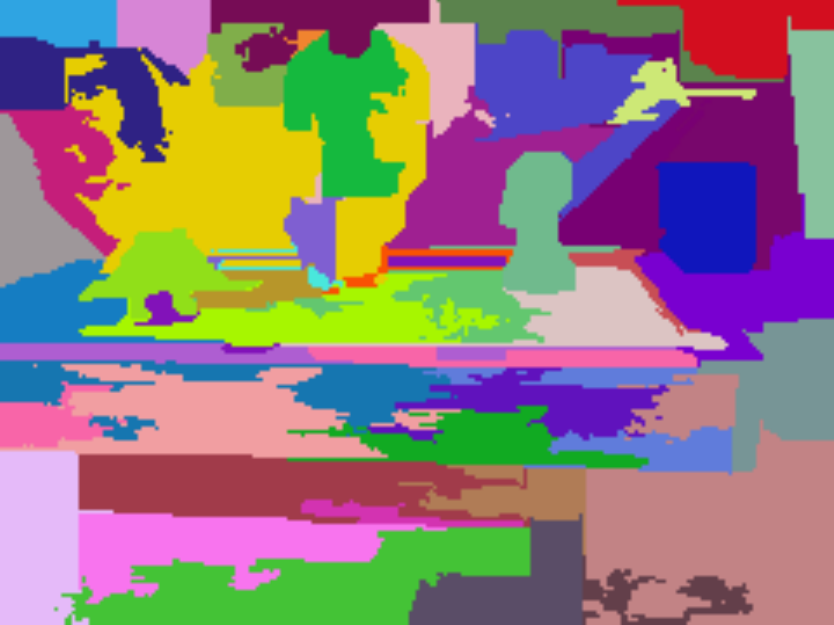} &
      \includegraphics[width=0.16\linewidth]{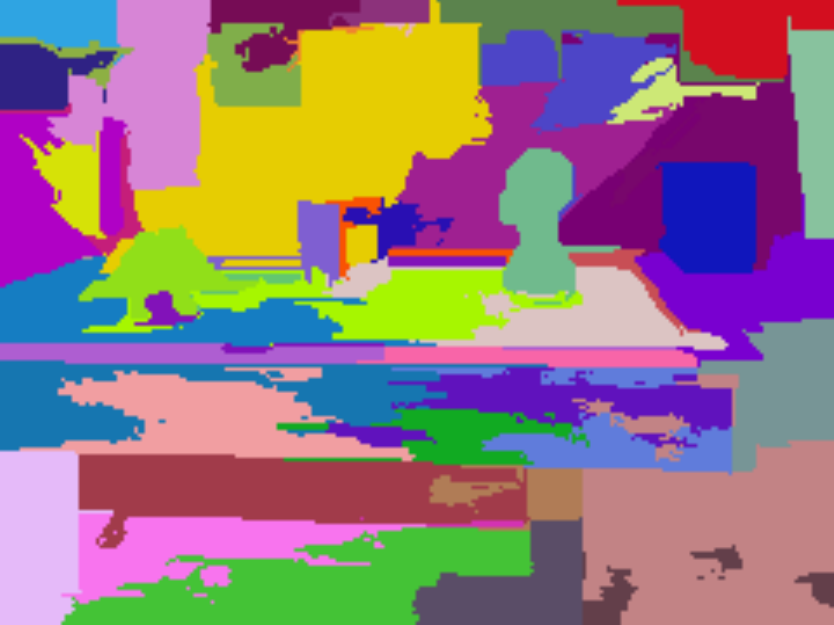} &
			\includegraphics[width=0.16\linewidth]{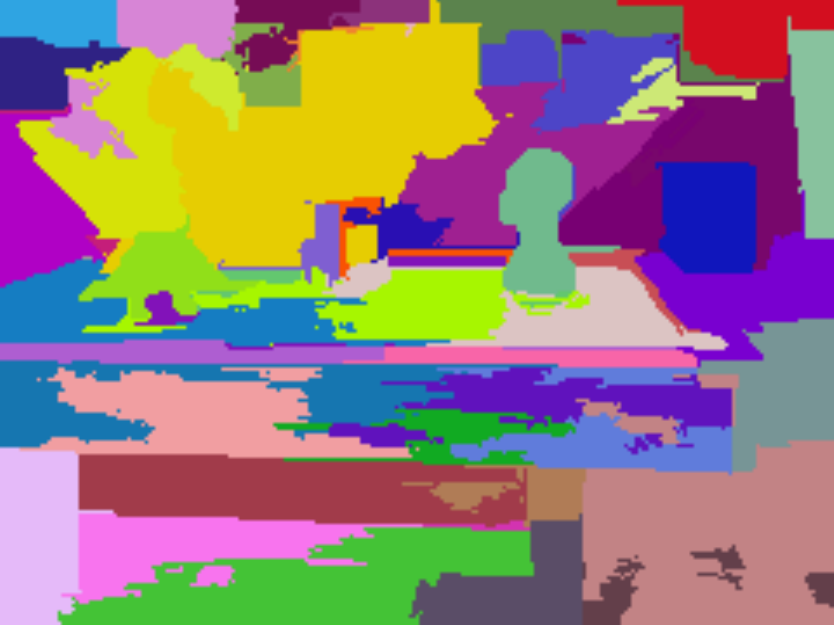}
    \end{tabular}
		\begin{tabular}{rccccc}
      \textbf{(C)} & \includegraphics[width=0.16\linewidth]{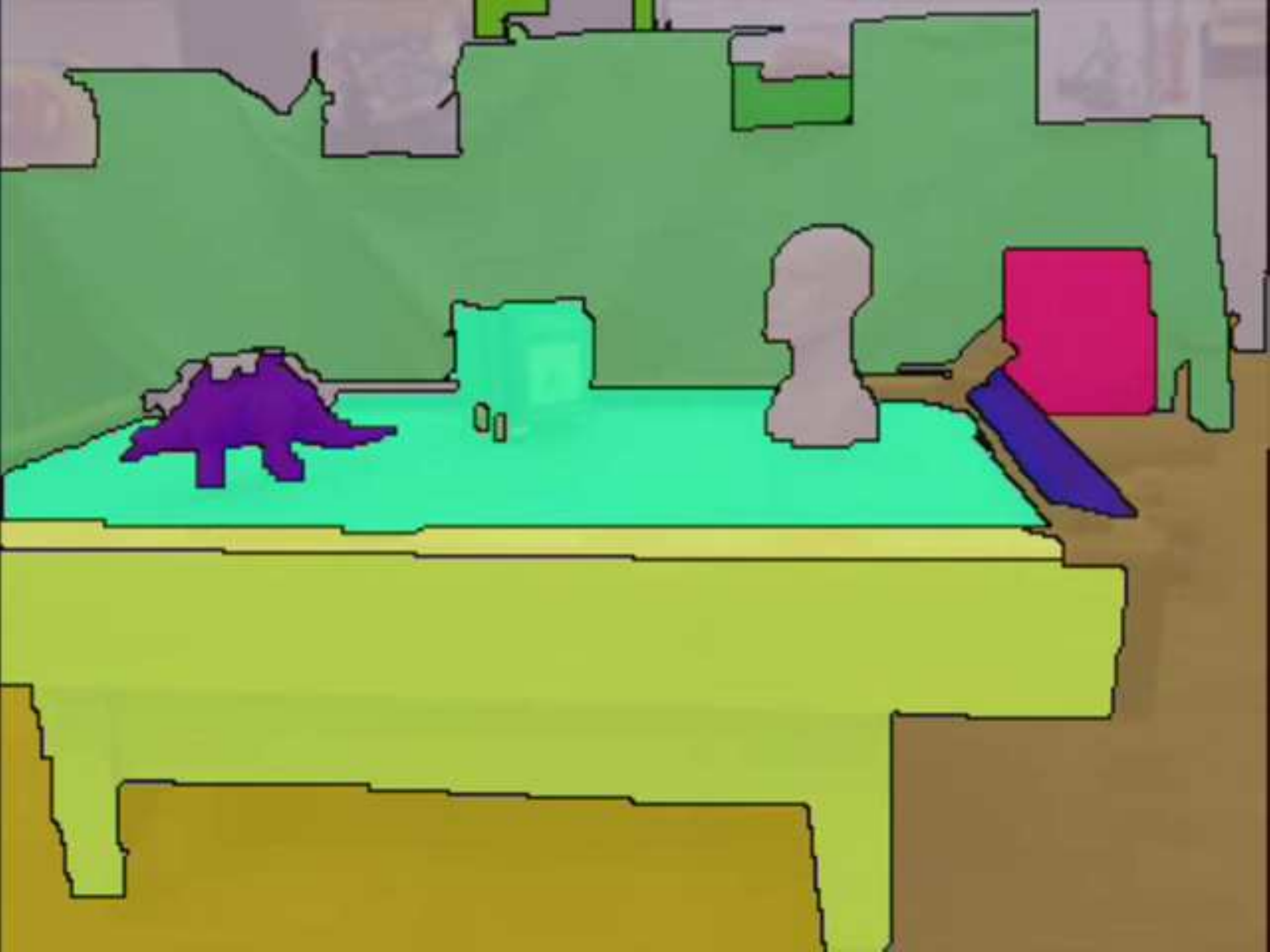} &
      \includegraphics[width=0.16\linewidth]{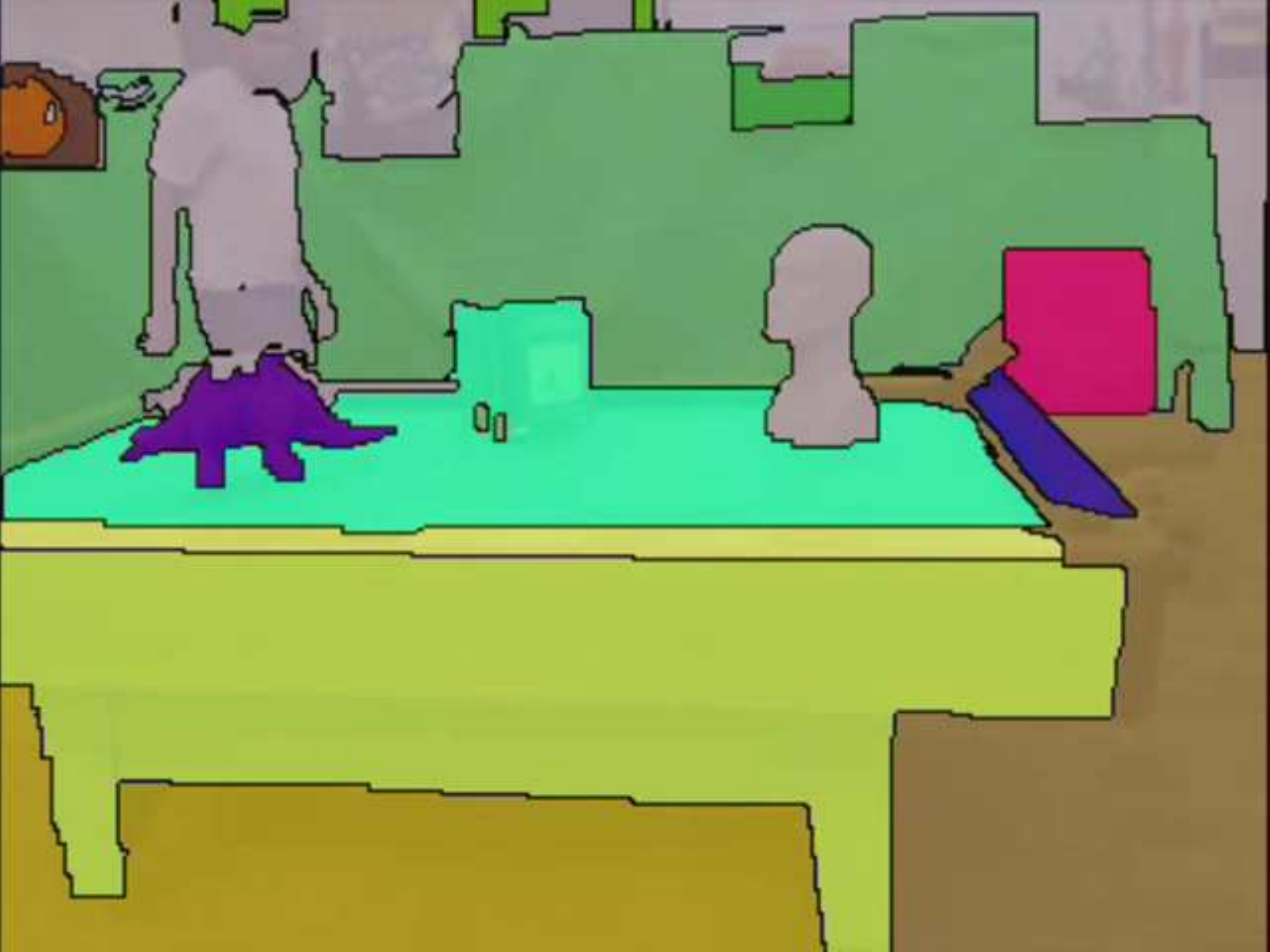} &
      \includegraphics[width=0.16\linewidth]{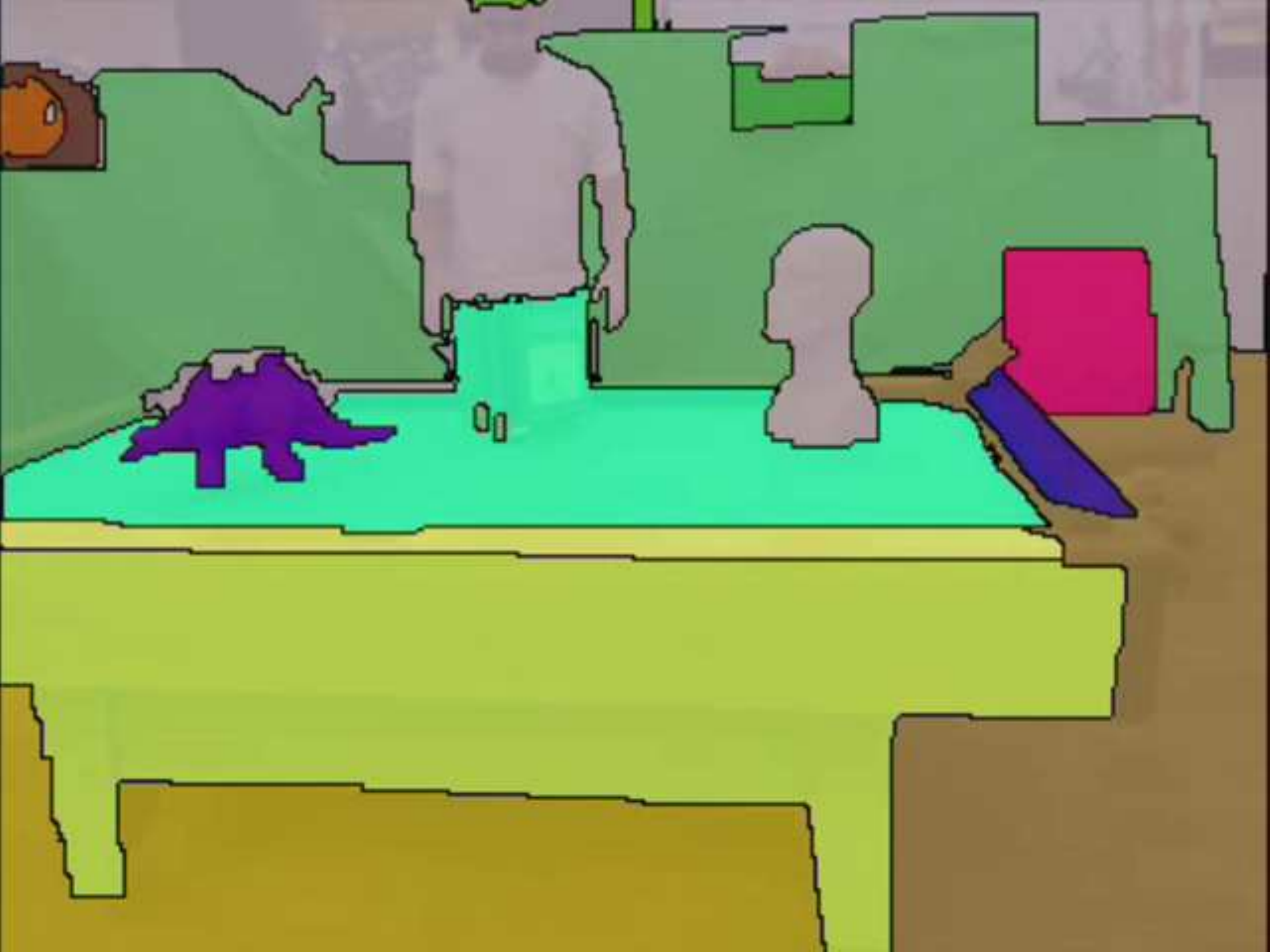} &
      \includegraphics[width=0.16\linewidth]{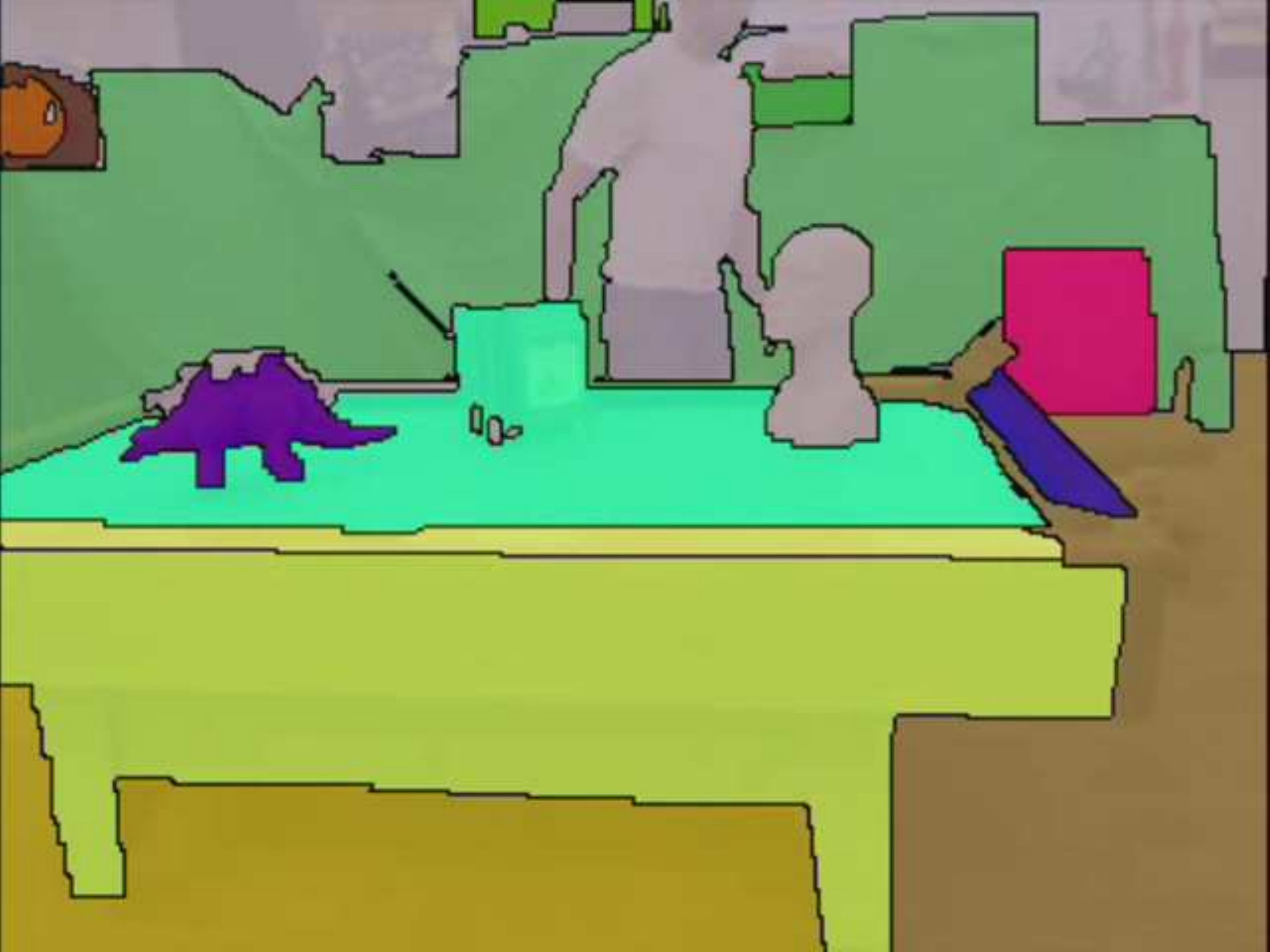} &
			\includegraphics[width=0.16\linewidth]{fig/na.eps}
    \end{tabular}
		\begin{tabular}{rccccc}
      \textbf{(D)} & \includegraphics[width=0.16\linewidth]{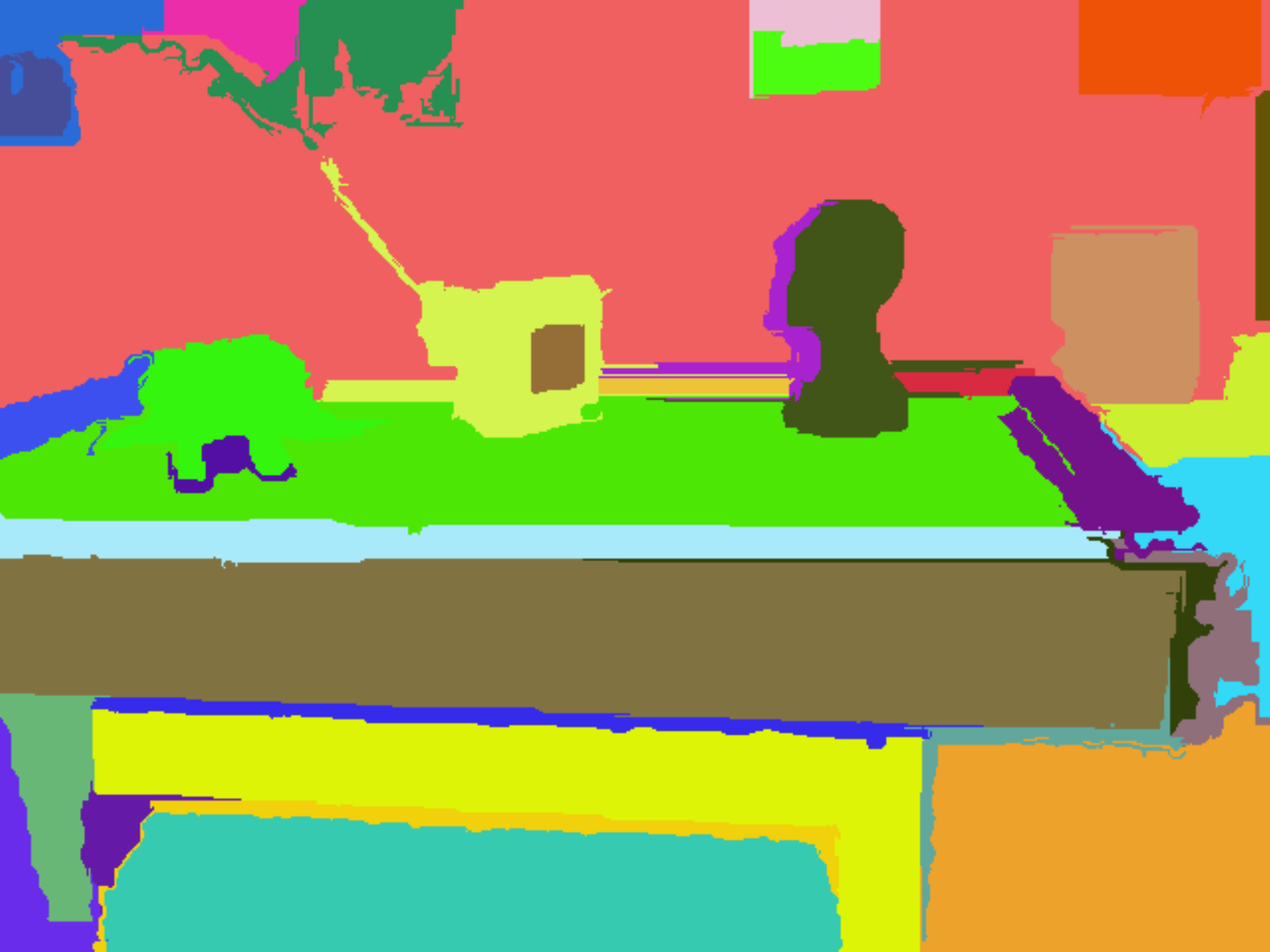} &
      \includegraphics[width=0.16\linewidth]{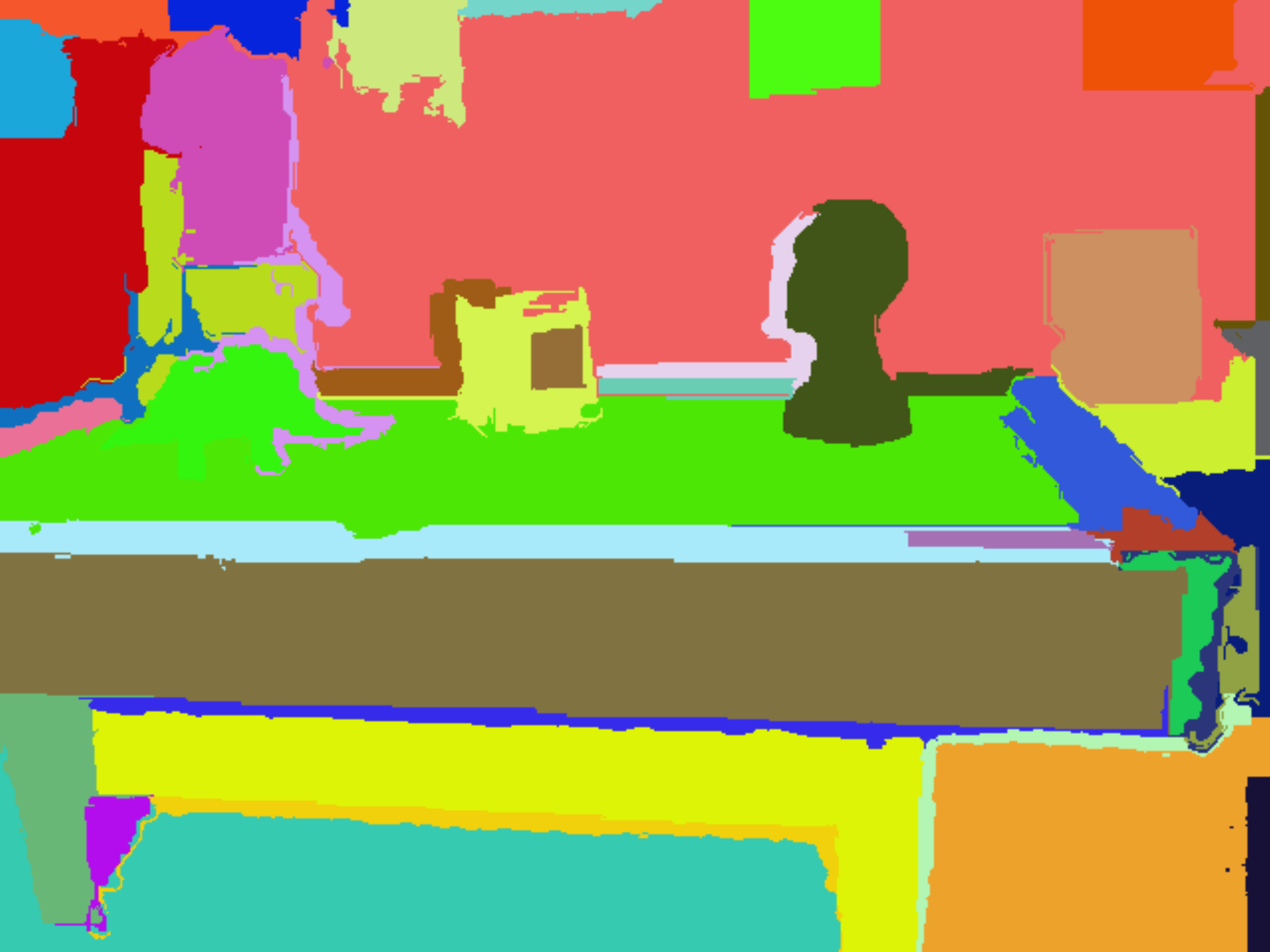} &
      \includegraphics[width=0.16\linewidth]{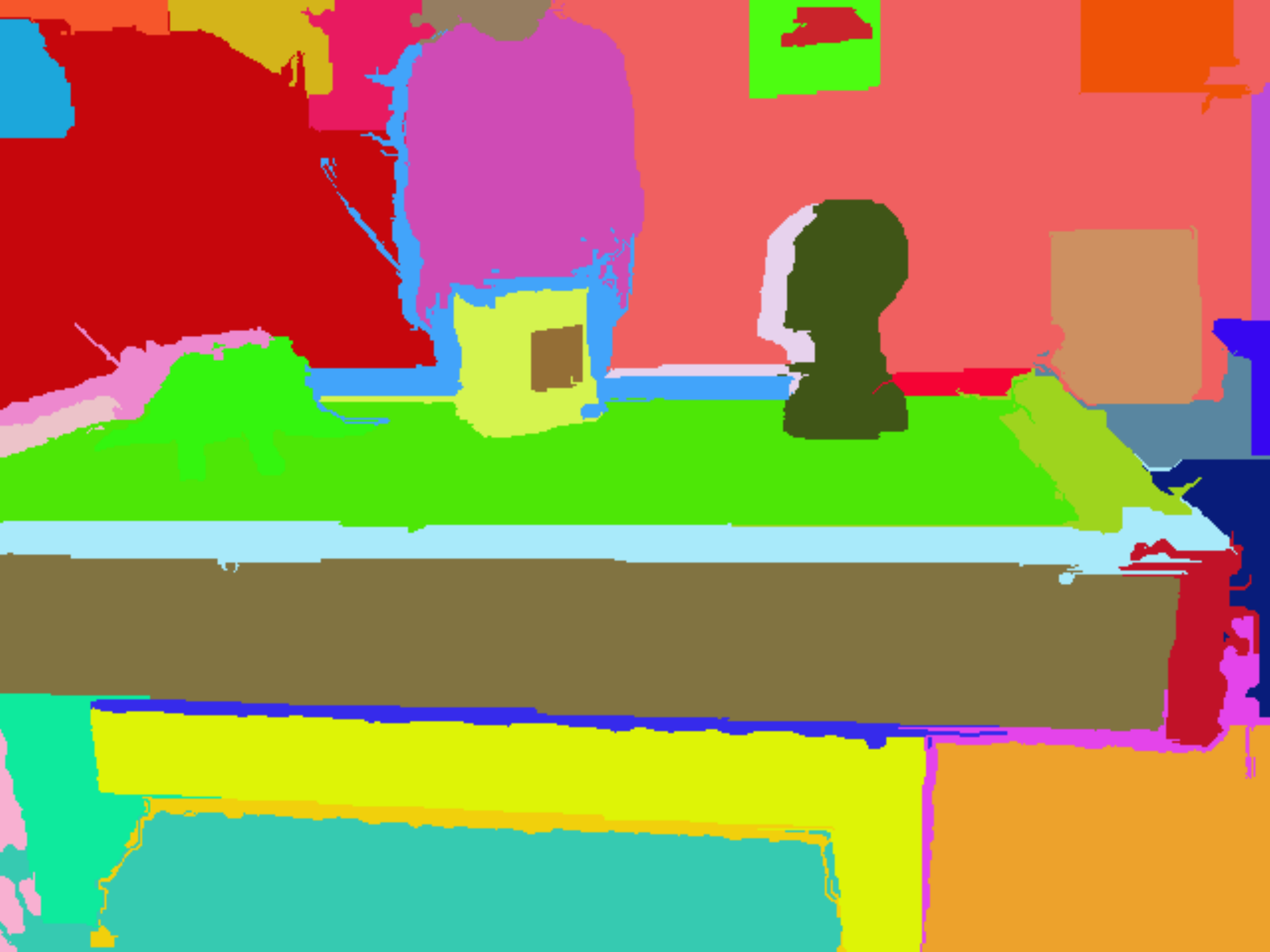} &
      \includegraphics[width=0.16\linewidth]{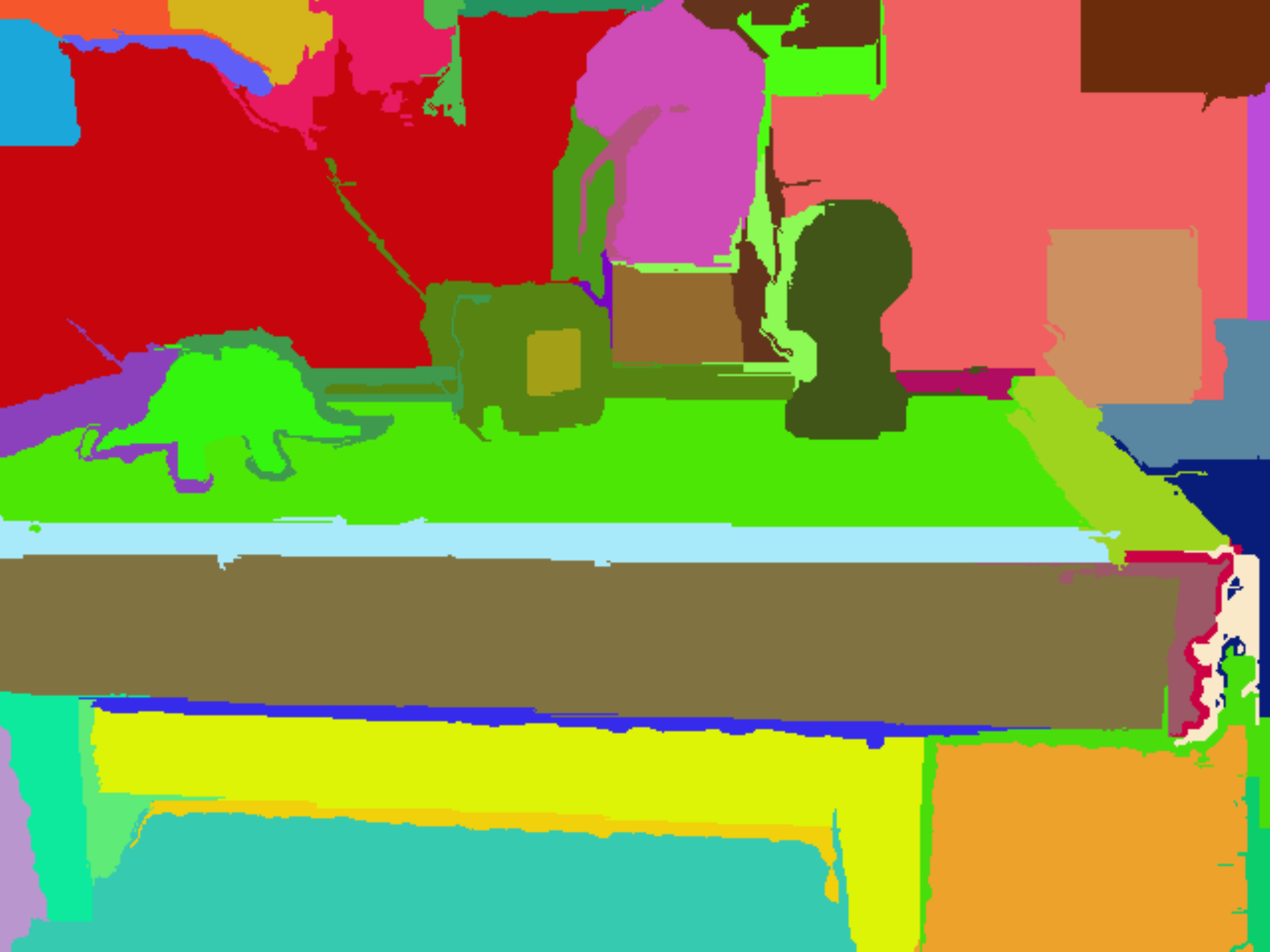} &
			\includegraphics[width=0.16\linewidth]{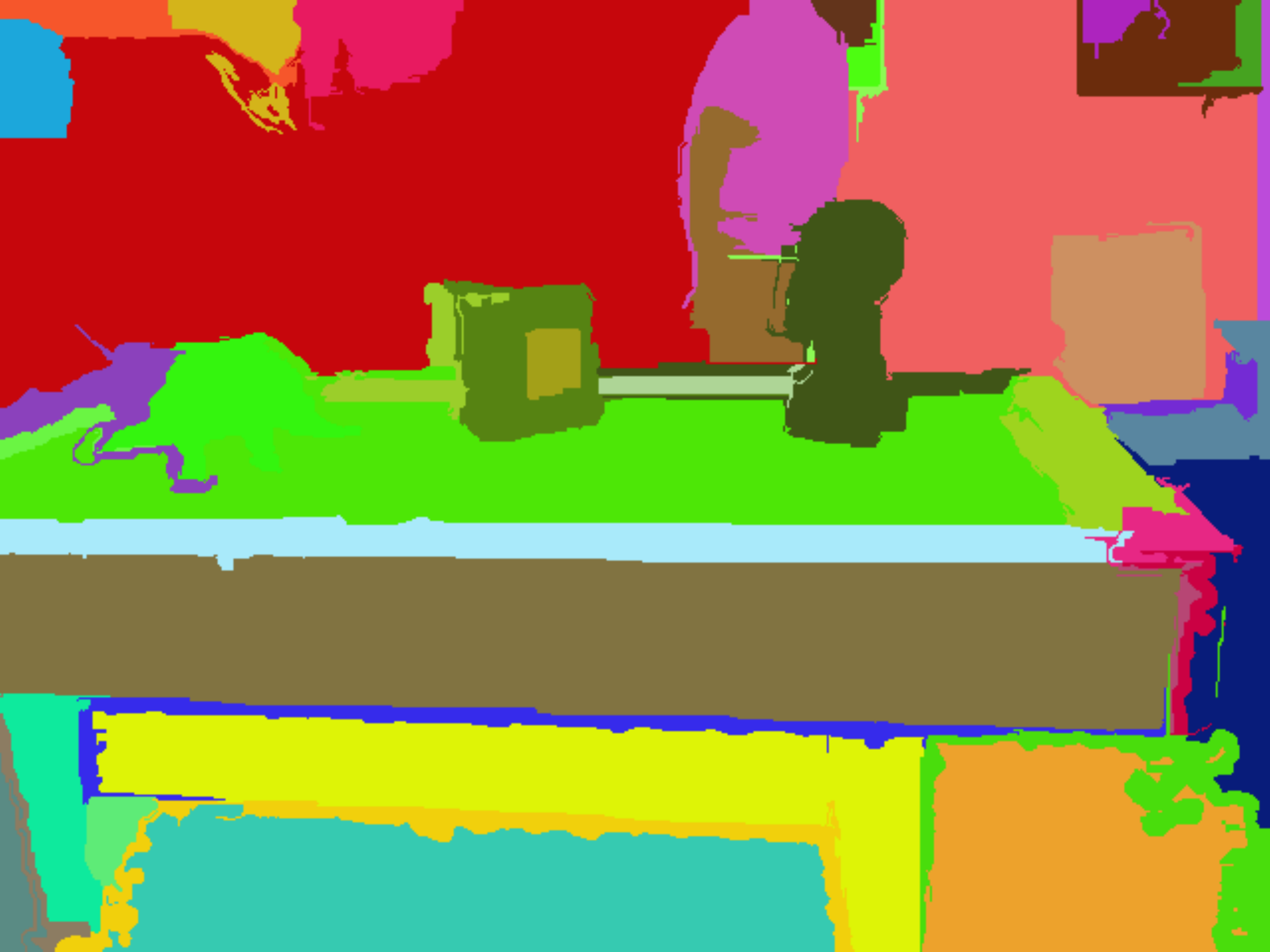}
    \end{tabular}
		\caption{Sequence 2, (A) : RGB image, (B): Results from \protect\cite{Xuy2012}, (C): Results from \protect\cite{Grundmann2010} (N/A is used when the frame is beyond the length it can run), (D): Our Results}
    \label{fig_fig13b}
\end{figure*}

\begin{figure*}  [t]
 \centering   
    \begin{tabular}{rccccc}
      \textbf{(A)} & \includegraphics[width=0.16\linewidth]{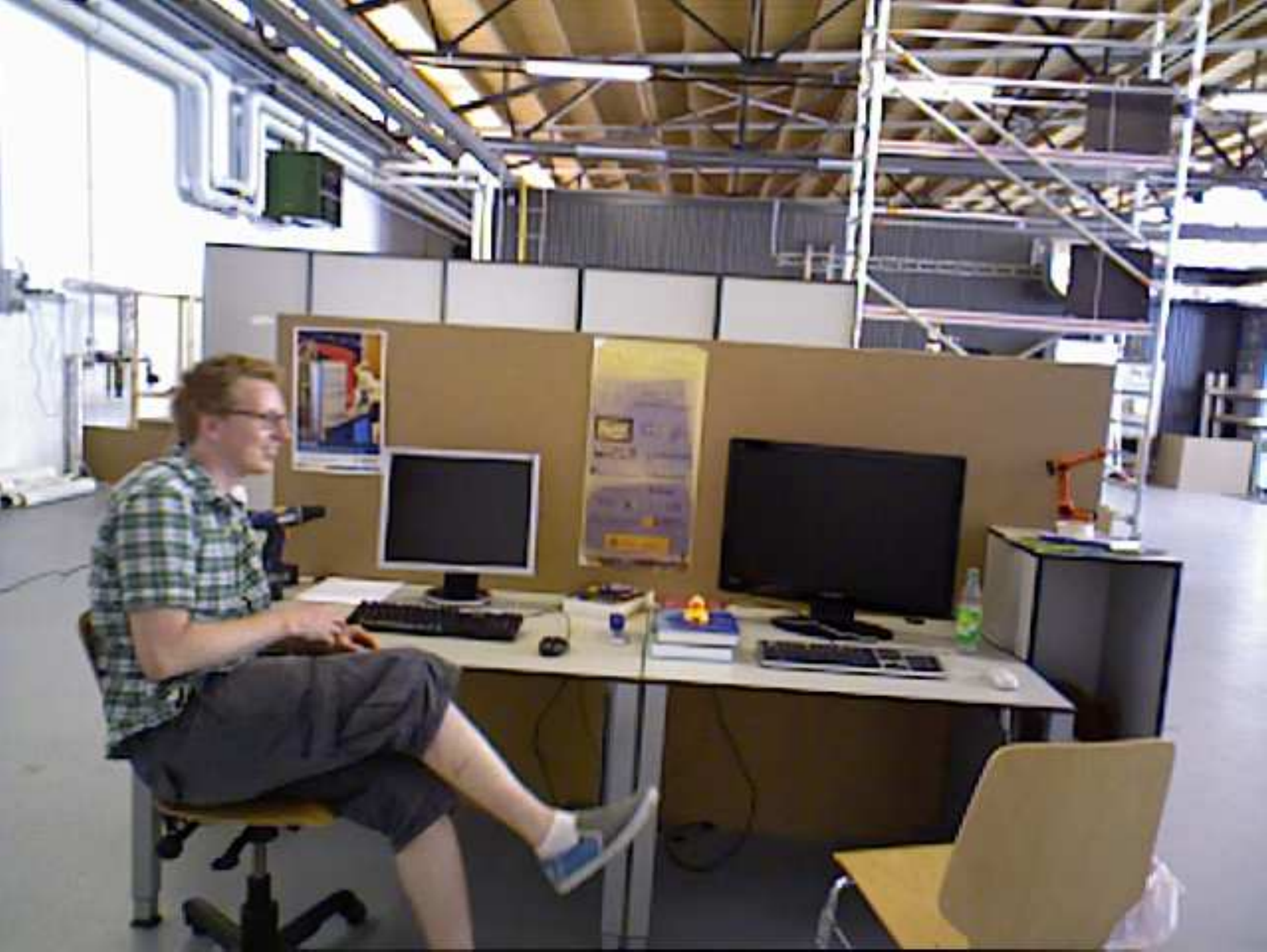} &
      \includegraphics[width=0.16\linewidth]{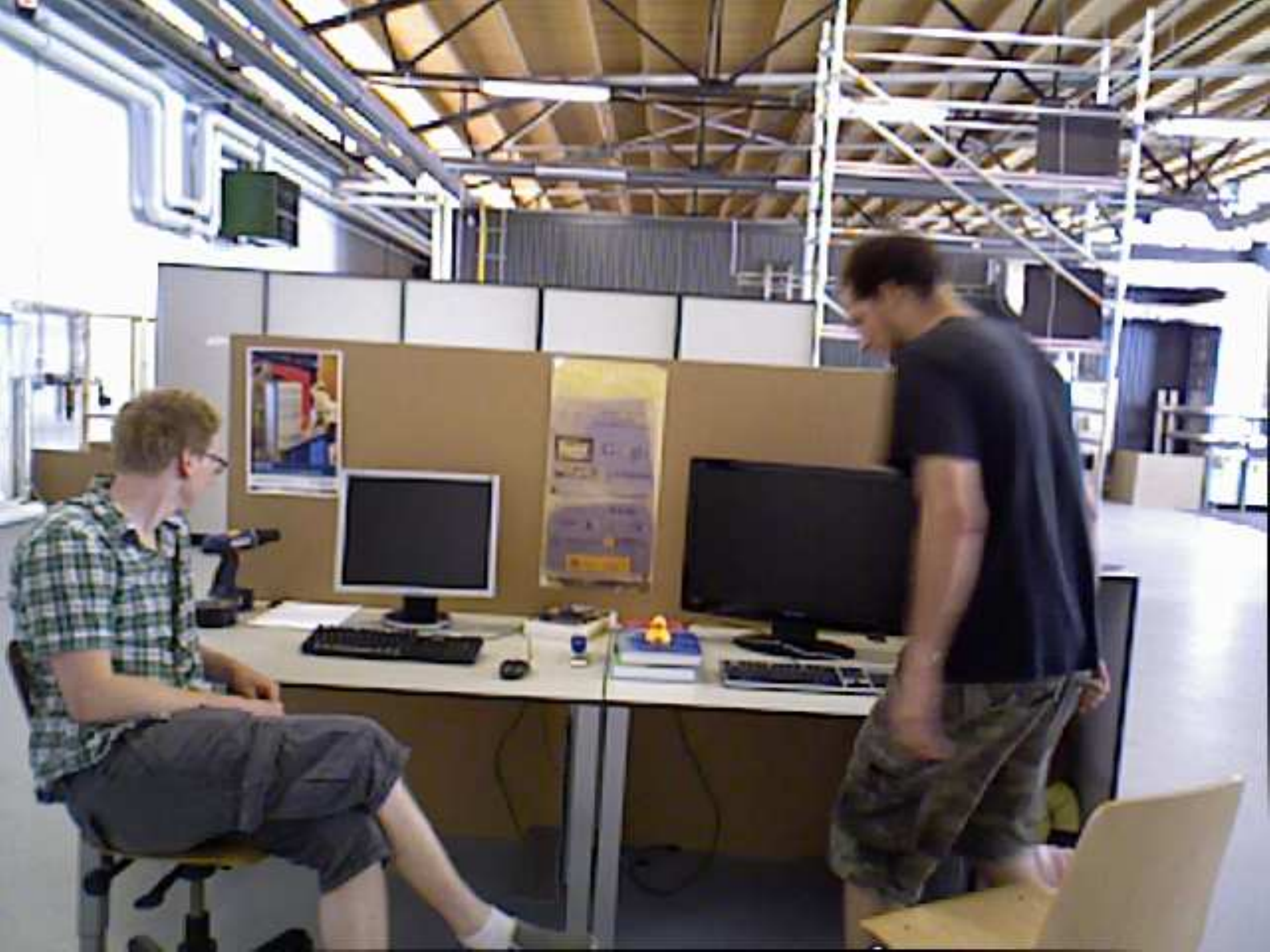} &
      \includegraphics[width=0.16\linewidth]{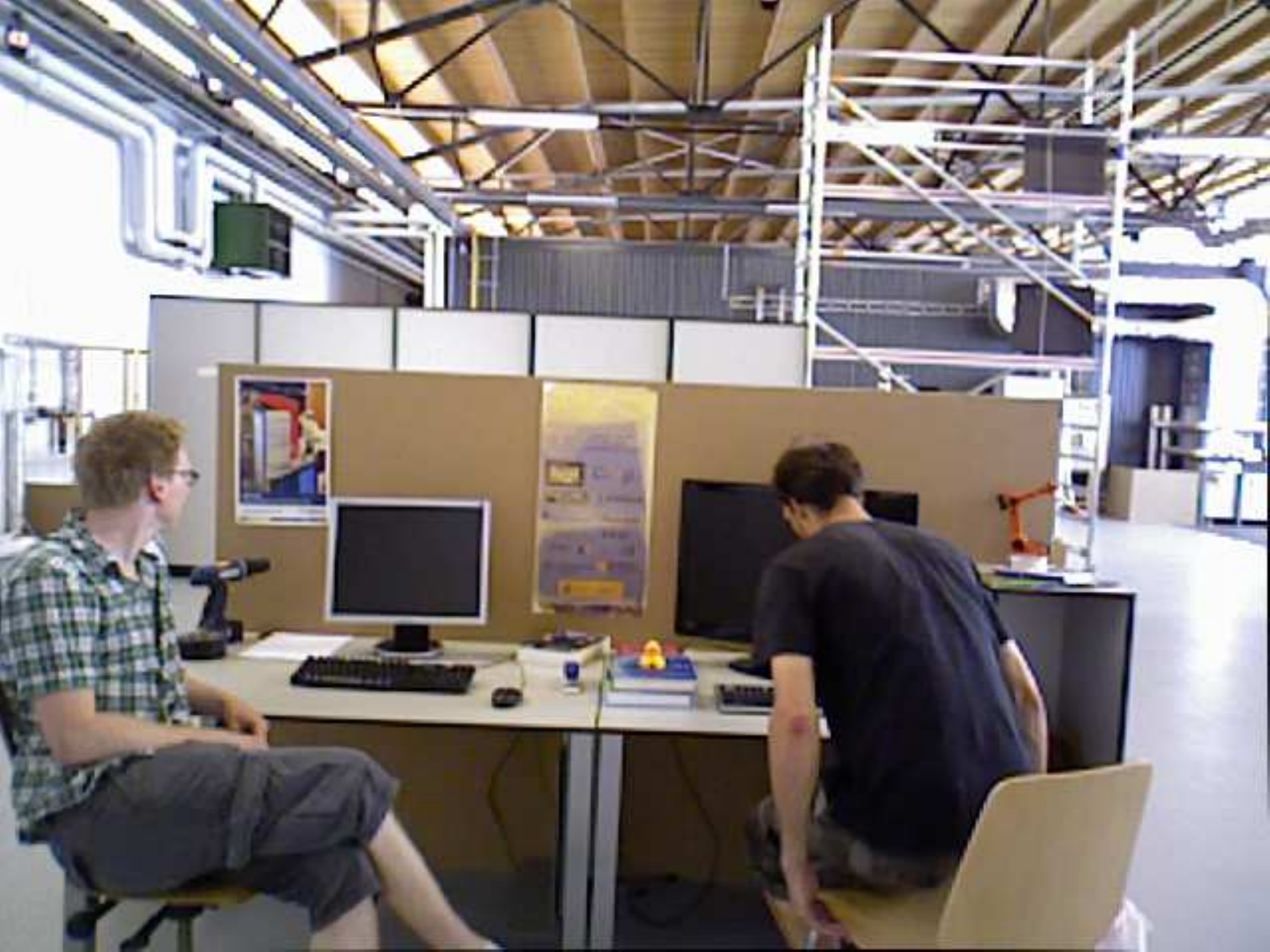} &
      \includegraphics[width=0.16\linewidth]{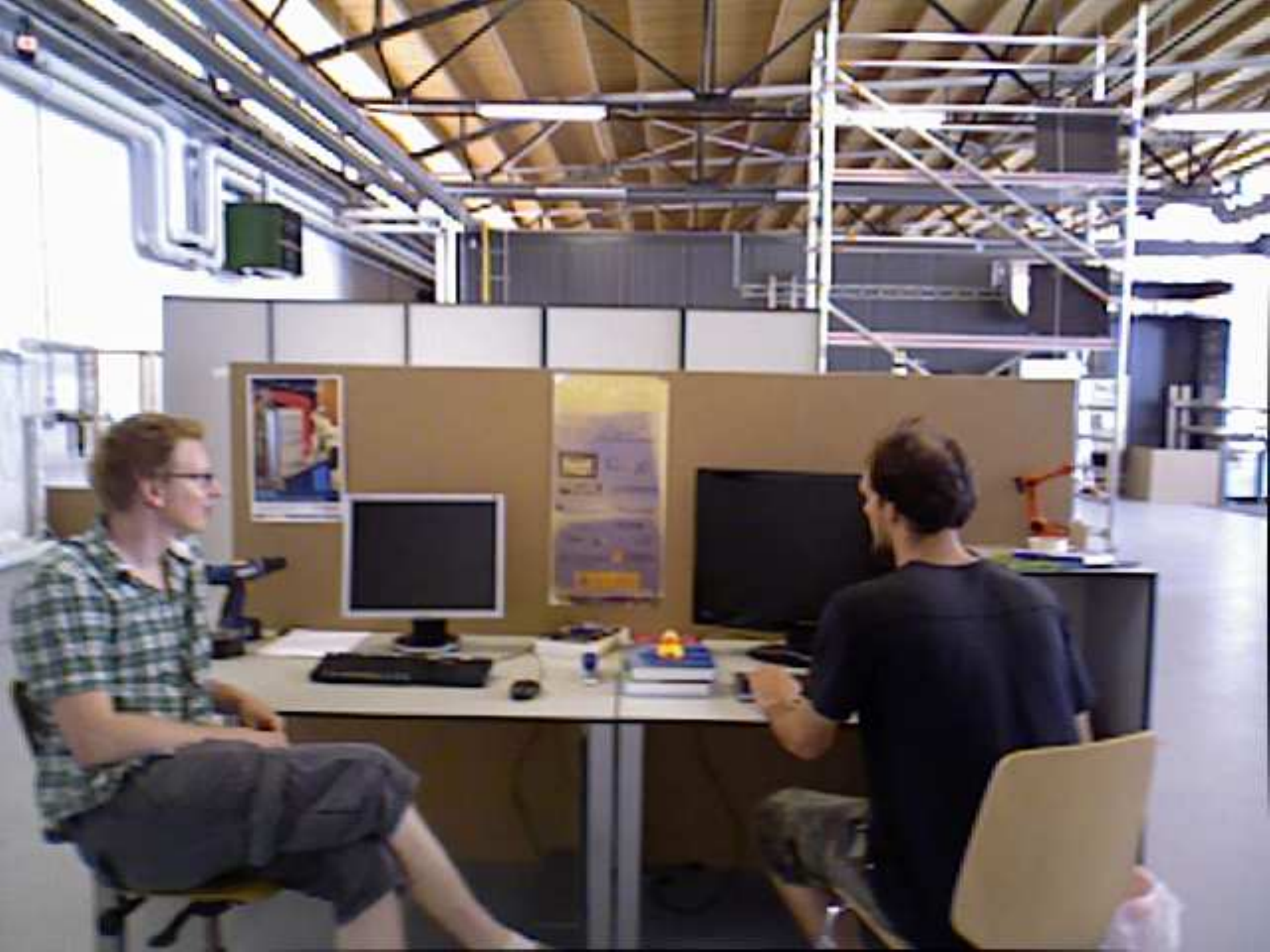} &
			\includegraphics[width=0.16\linewidth]{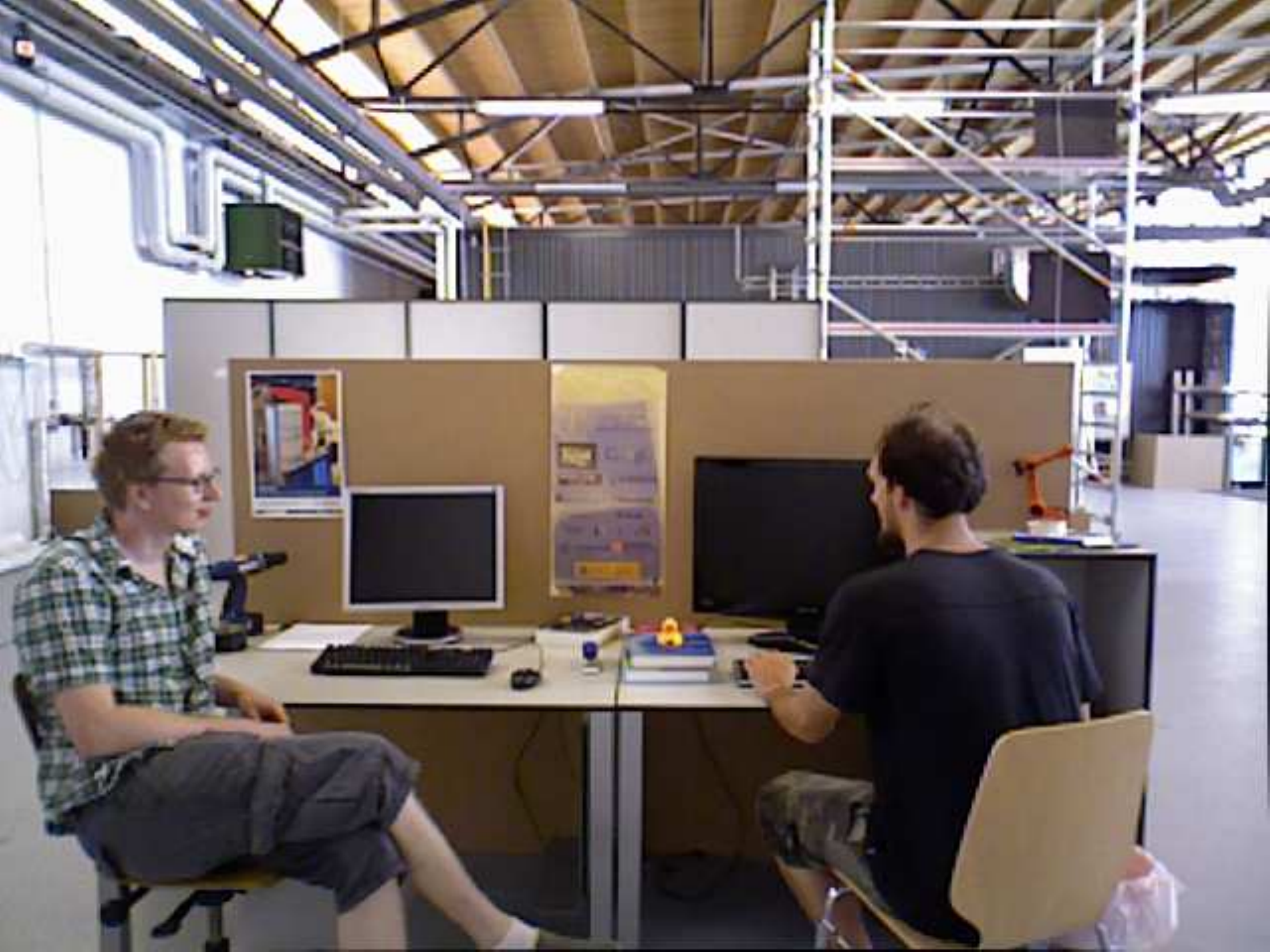}
    \end{tabular}
		\begin{tabular}{rccccc}
      \textbf{(B)} & \includegraphics[width=0.16\linewidth]{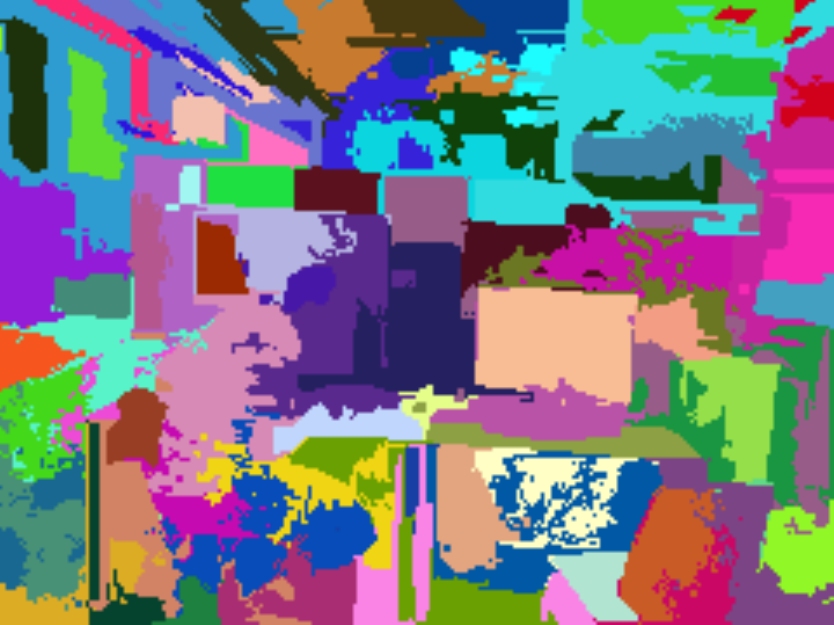} &
      \includegraphics[width=0.16\linewidth]{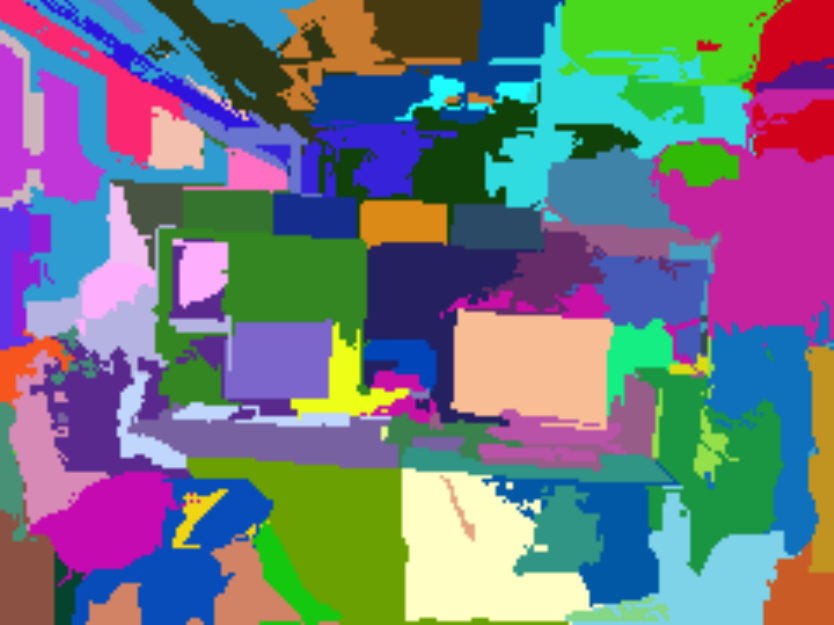} &
      \includegraphics[width=0.16\linewidth]{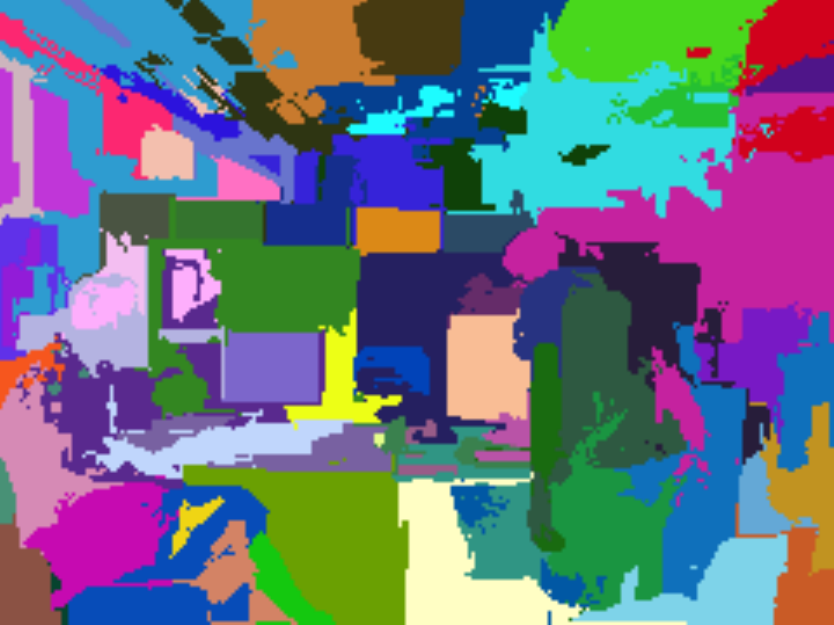} &
      \includegraphics[width=0.16\linewidth]{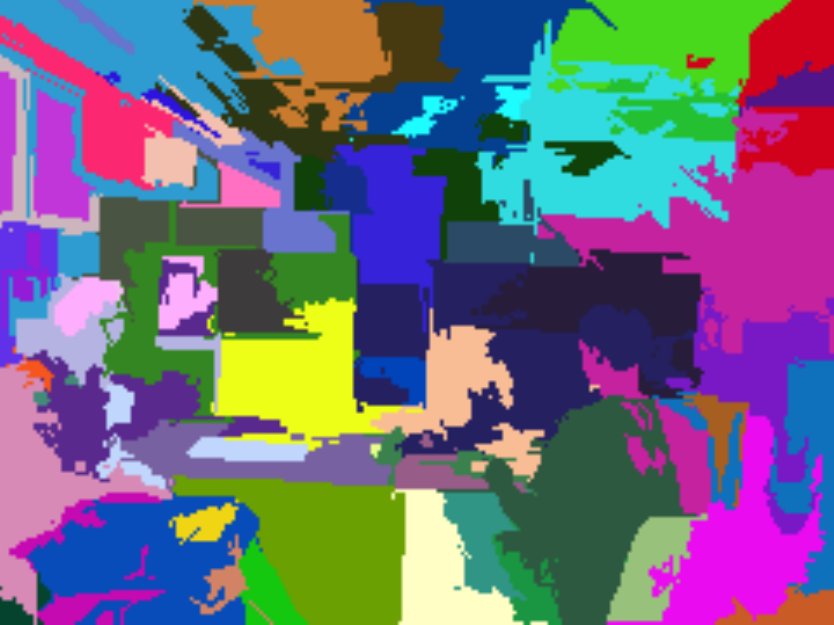} &
			\includegraphics[width=0.16\linewidth]{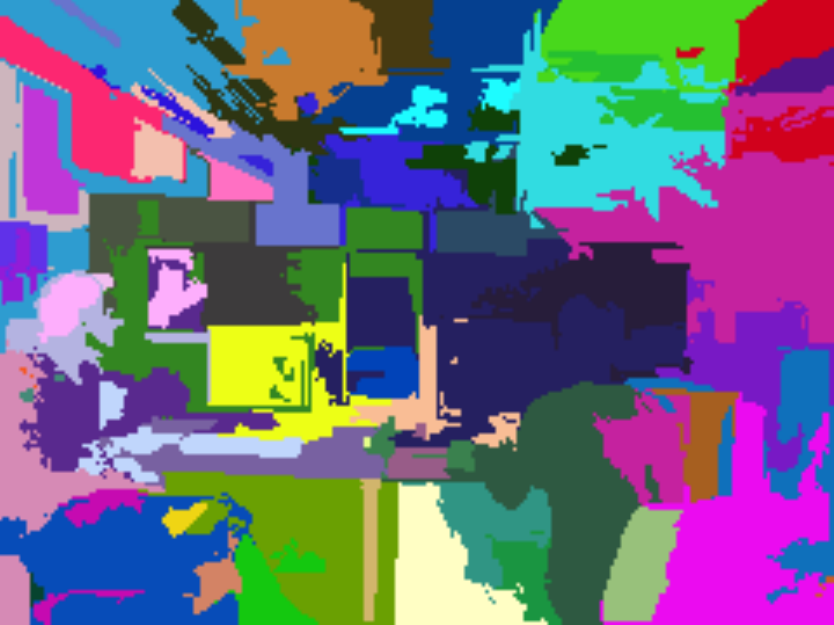}
    \end{tabular}
		\begin{tabular}{rccccc}
      \textbf{(C)}& \includegraphics[width=0.16\linewidth]{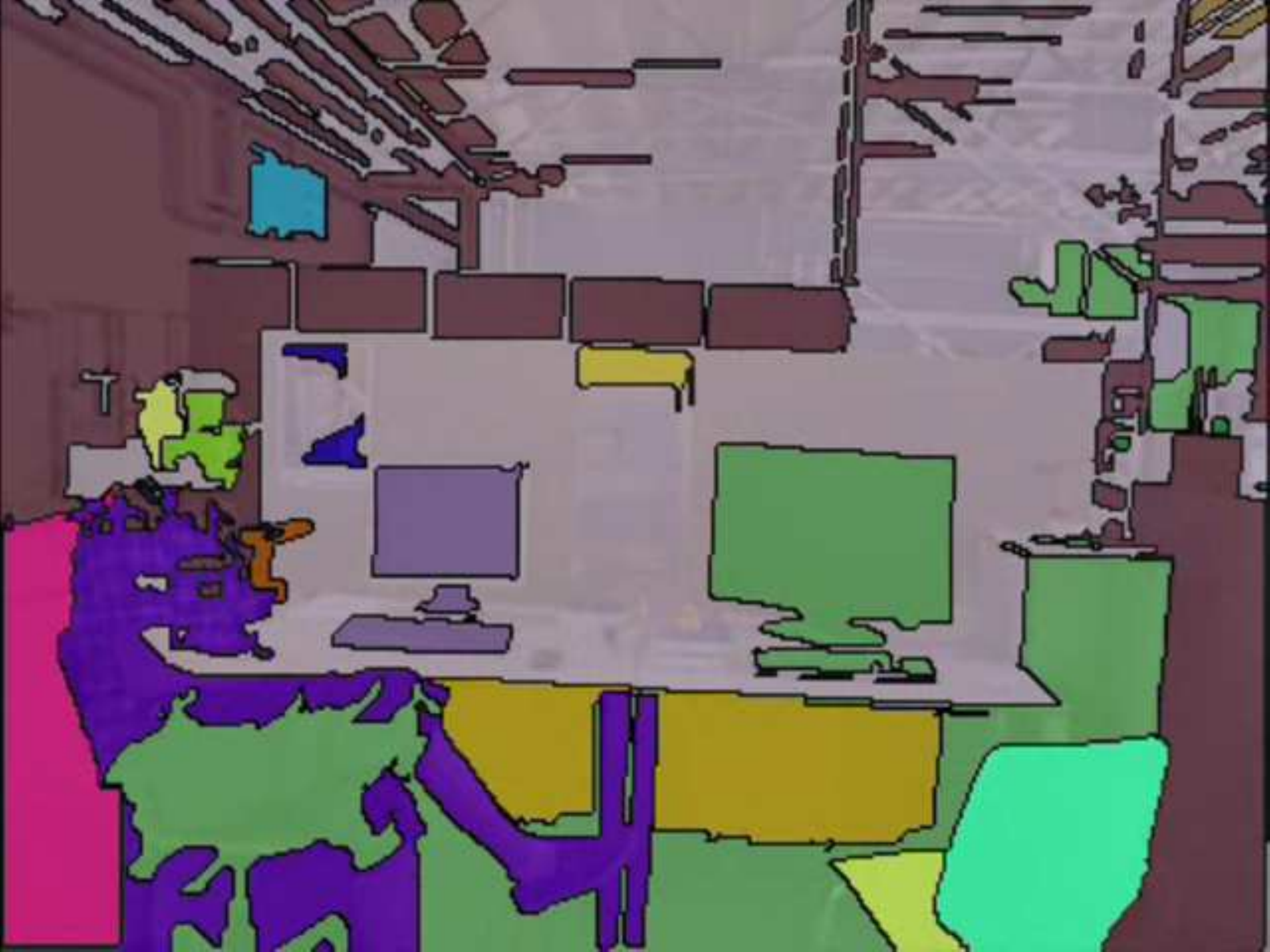} &
      \includegraphics[width=0.16\linewidth]{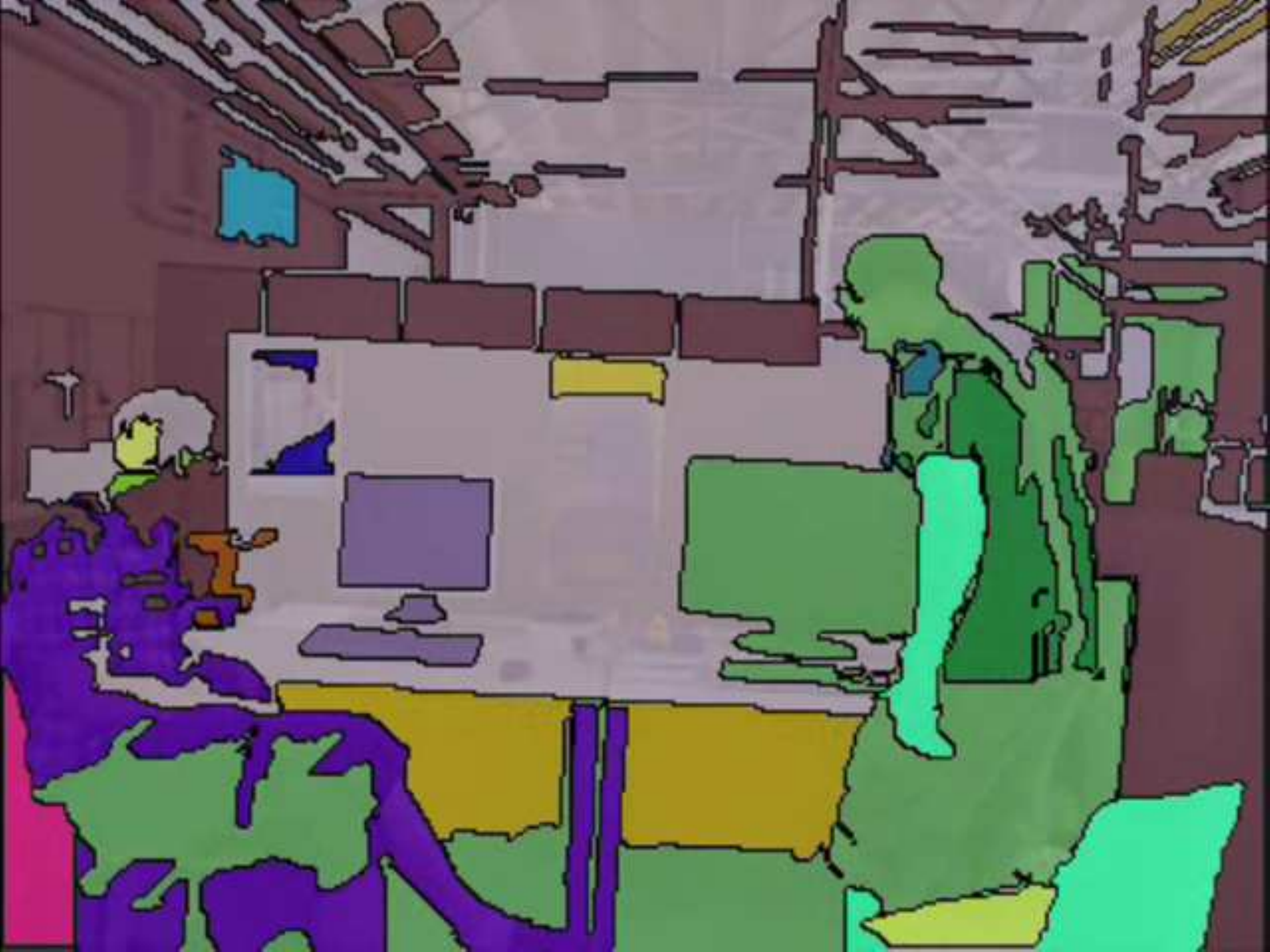} &
      \includegraphics[width=0.16\linewidth]{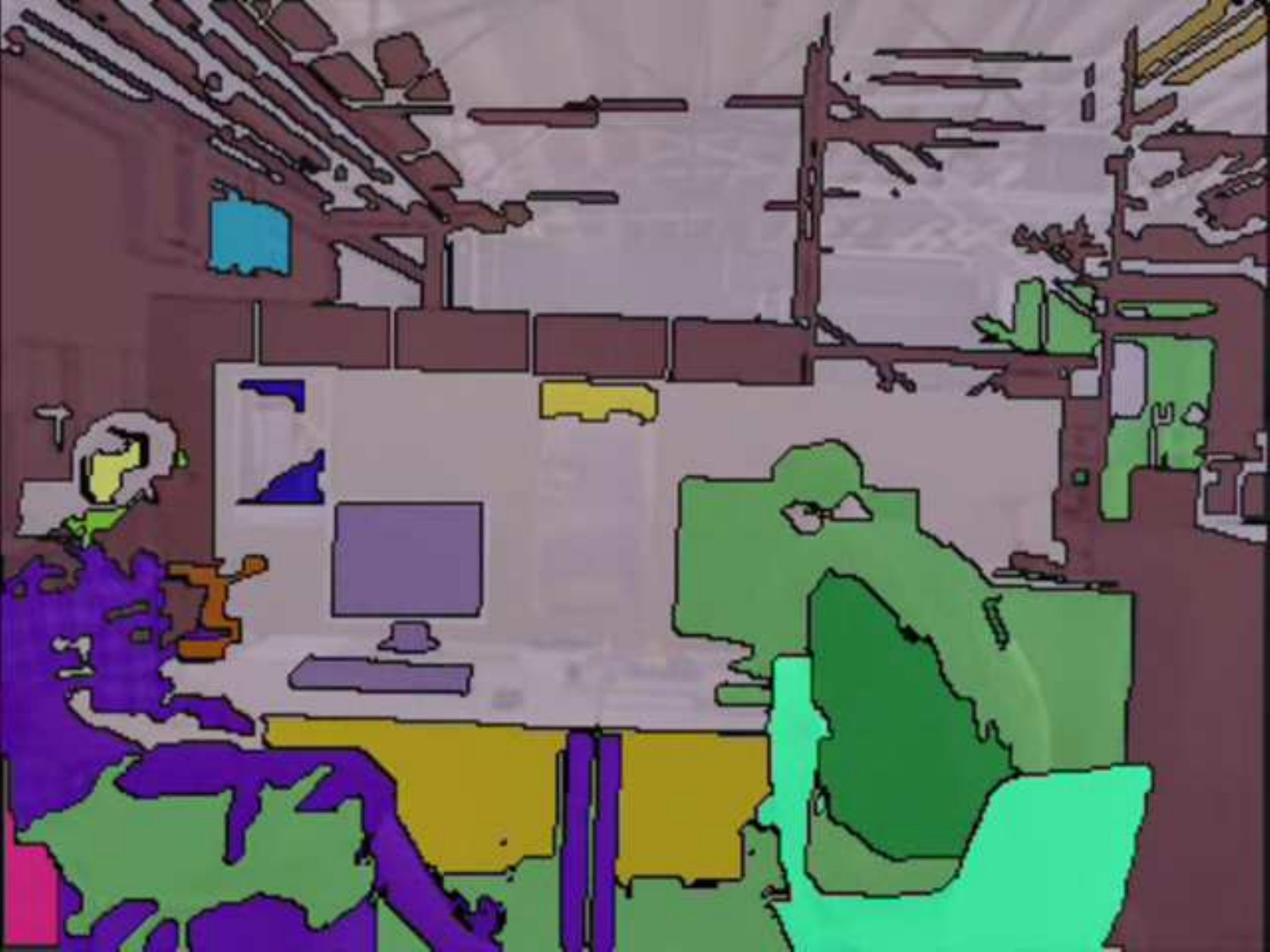} &
      \includegraphics[width=0.16\linewidth]{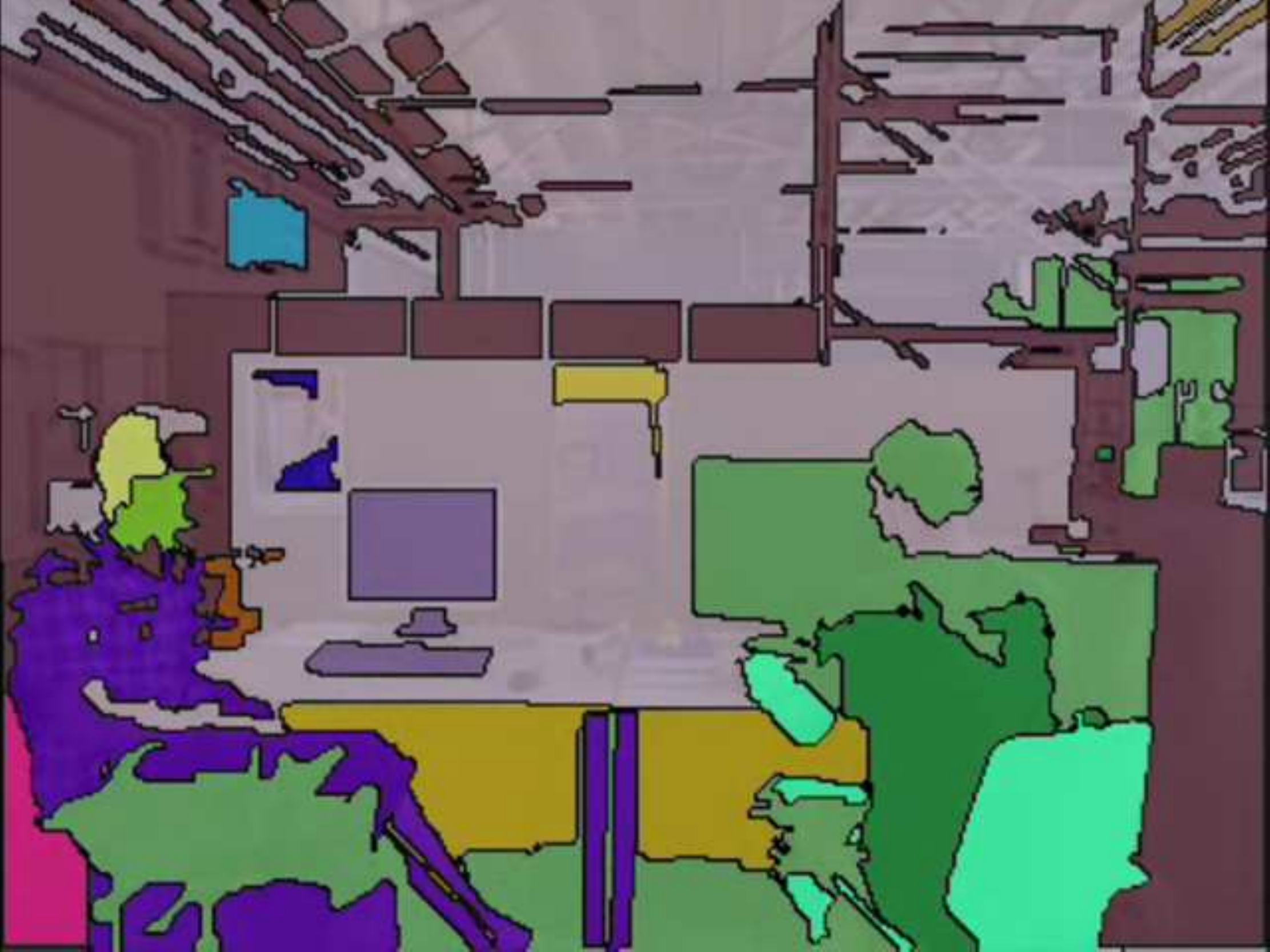} &
			\includegraphics[width=0.16\linewidth]{fig/na.eps}
    \end{tabular}
		\begin{tabular}{rccccc}
      \textbf{(D)}&\includegraphics[width=0.16\linewidth]{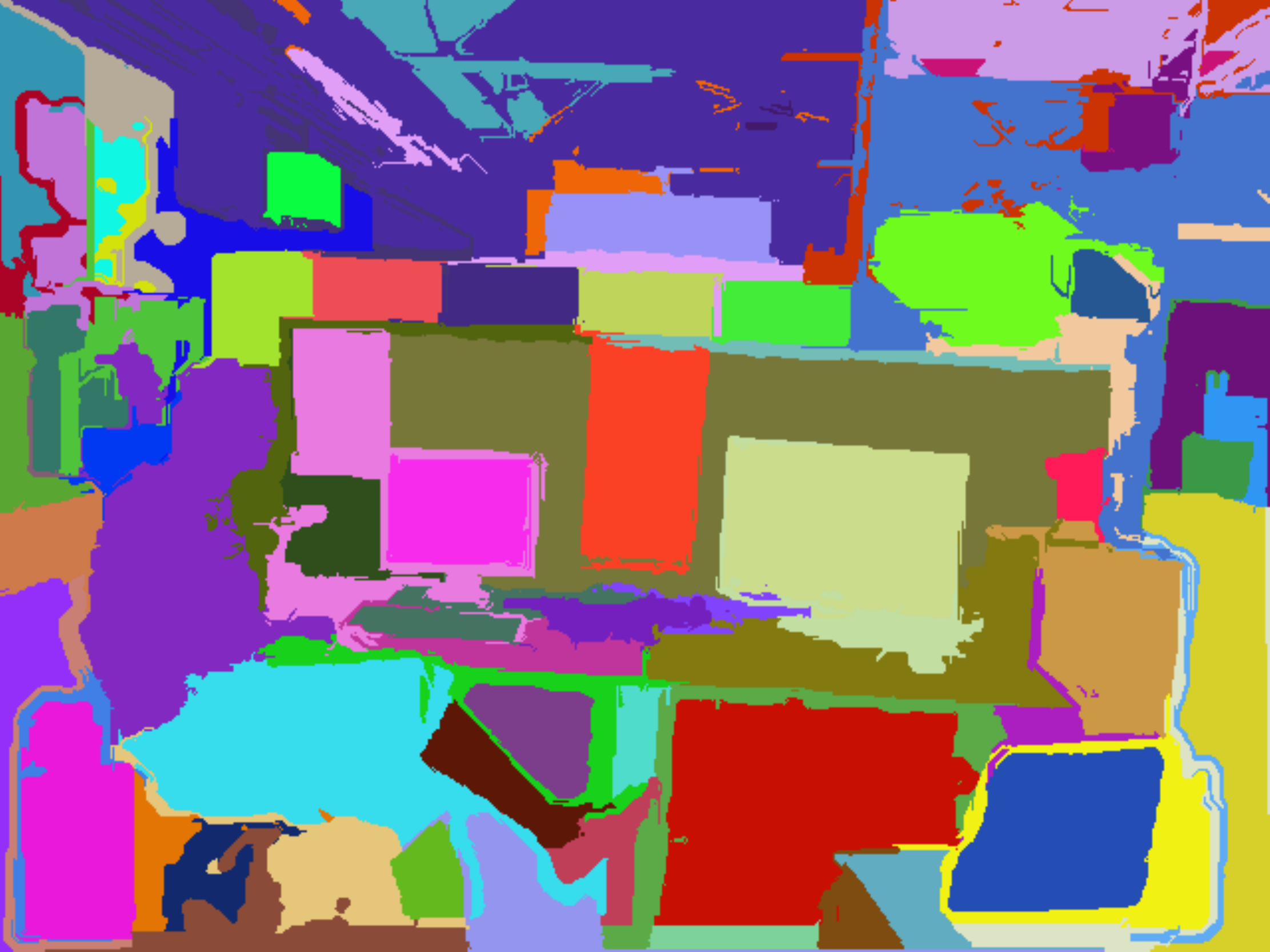} &
      \includegraphics[width=0.16\linewidth]{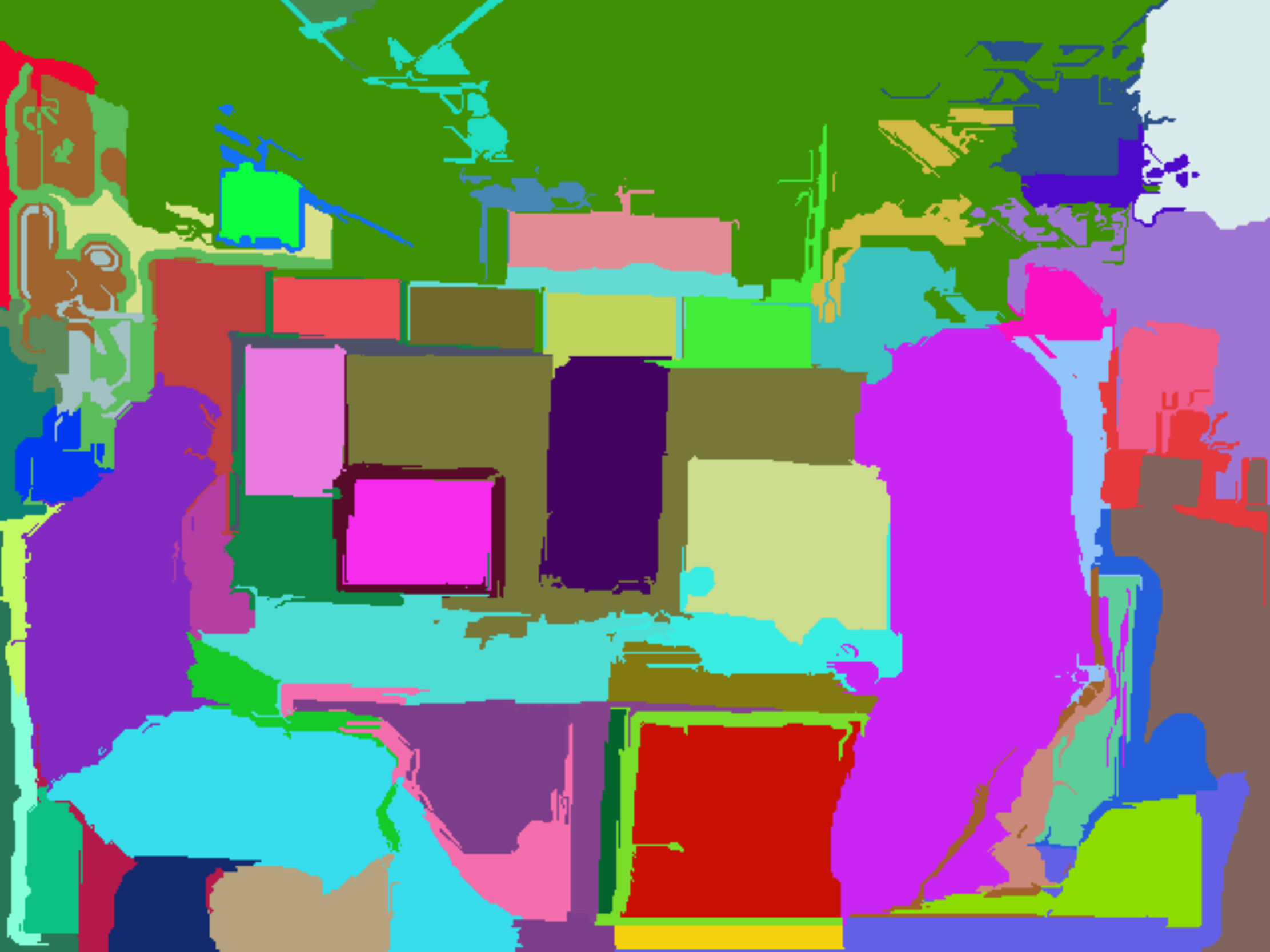} &
      \includegraphics[width=0.16\linewidth]{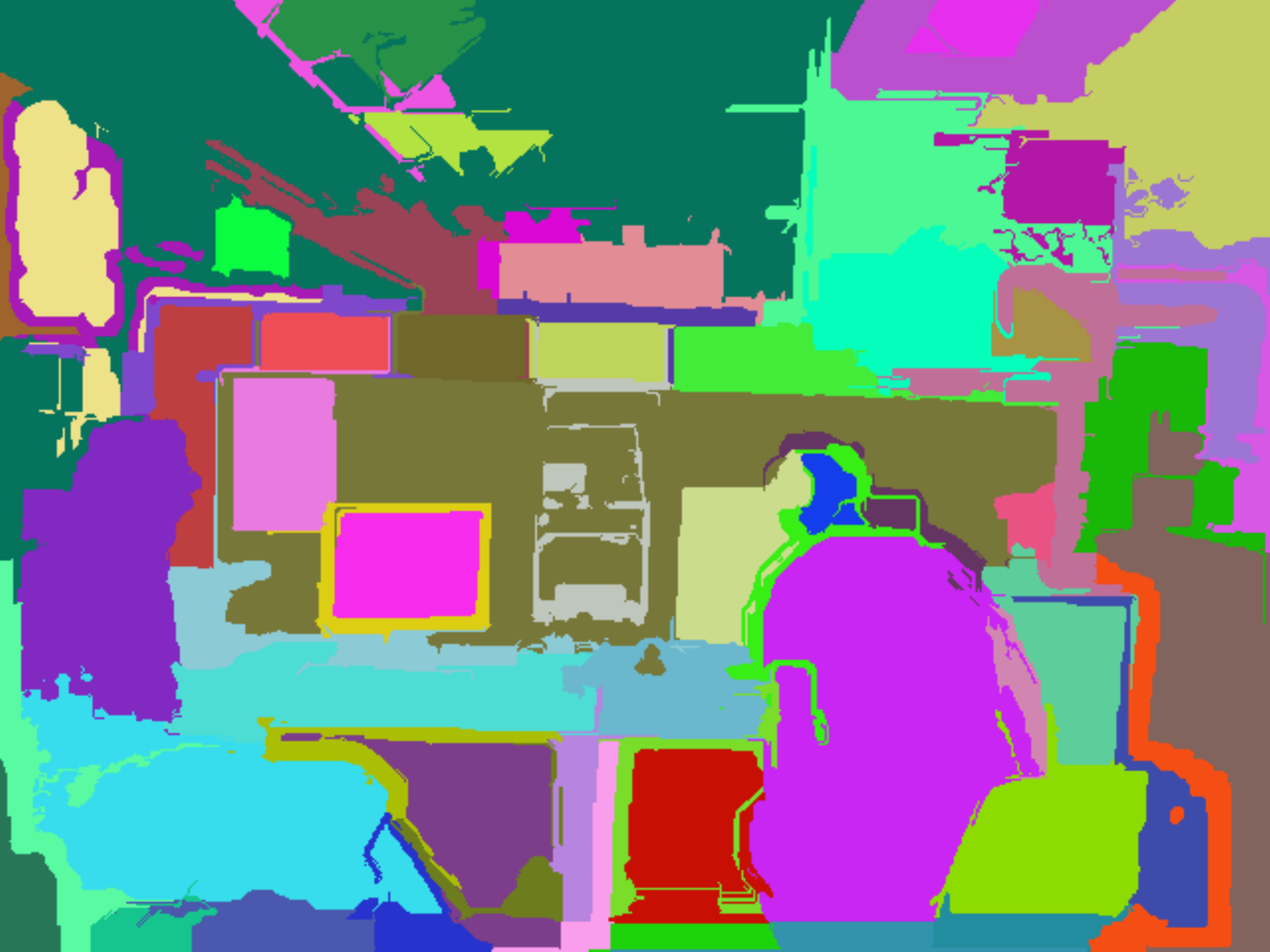} &
      \includegraphics[width=0.16\linewidth]{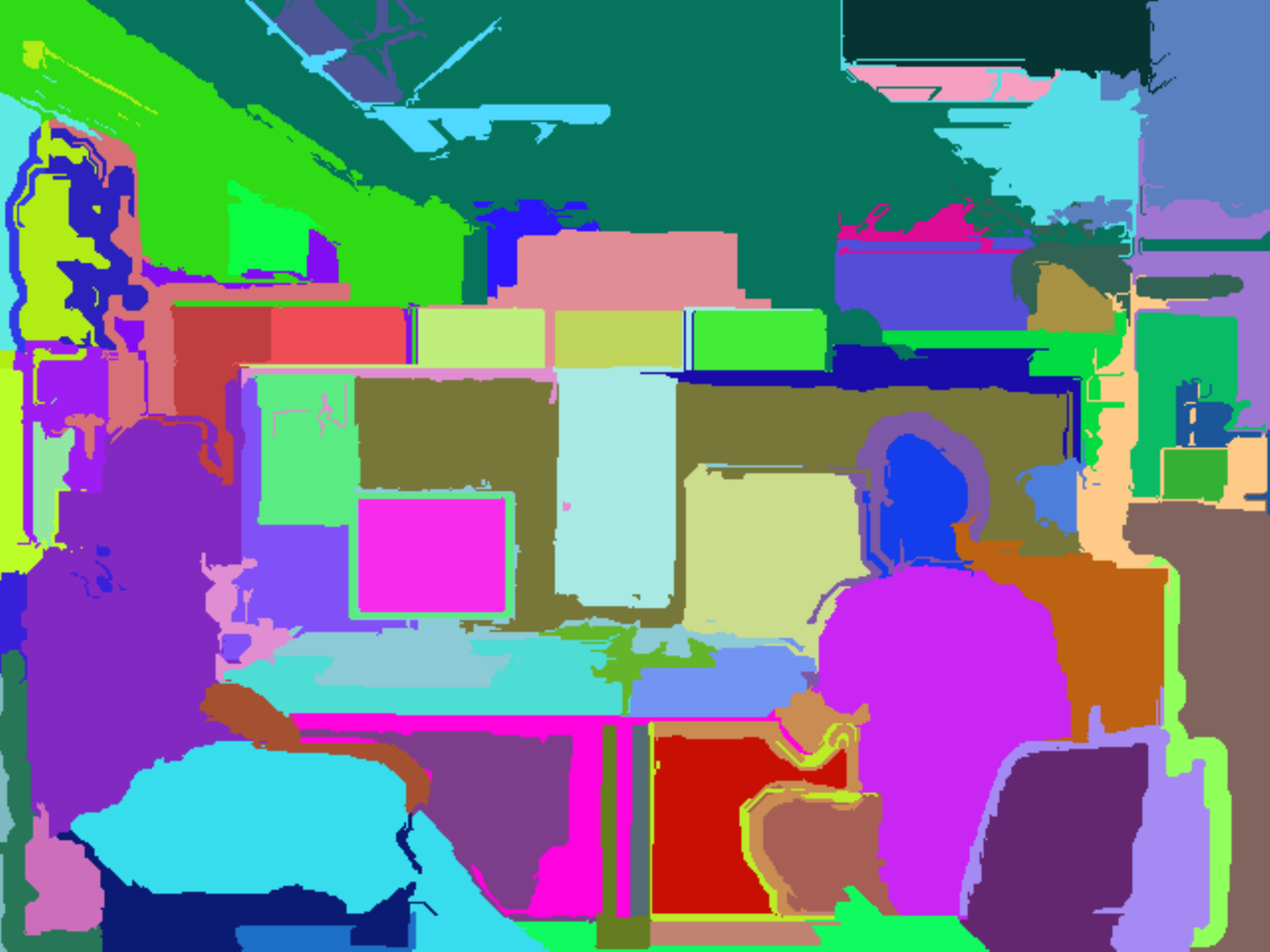} &
			\includegraphics[width=0.16\linewidth]{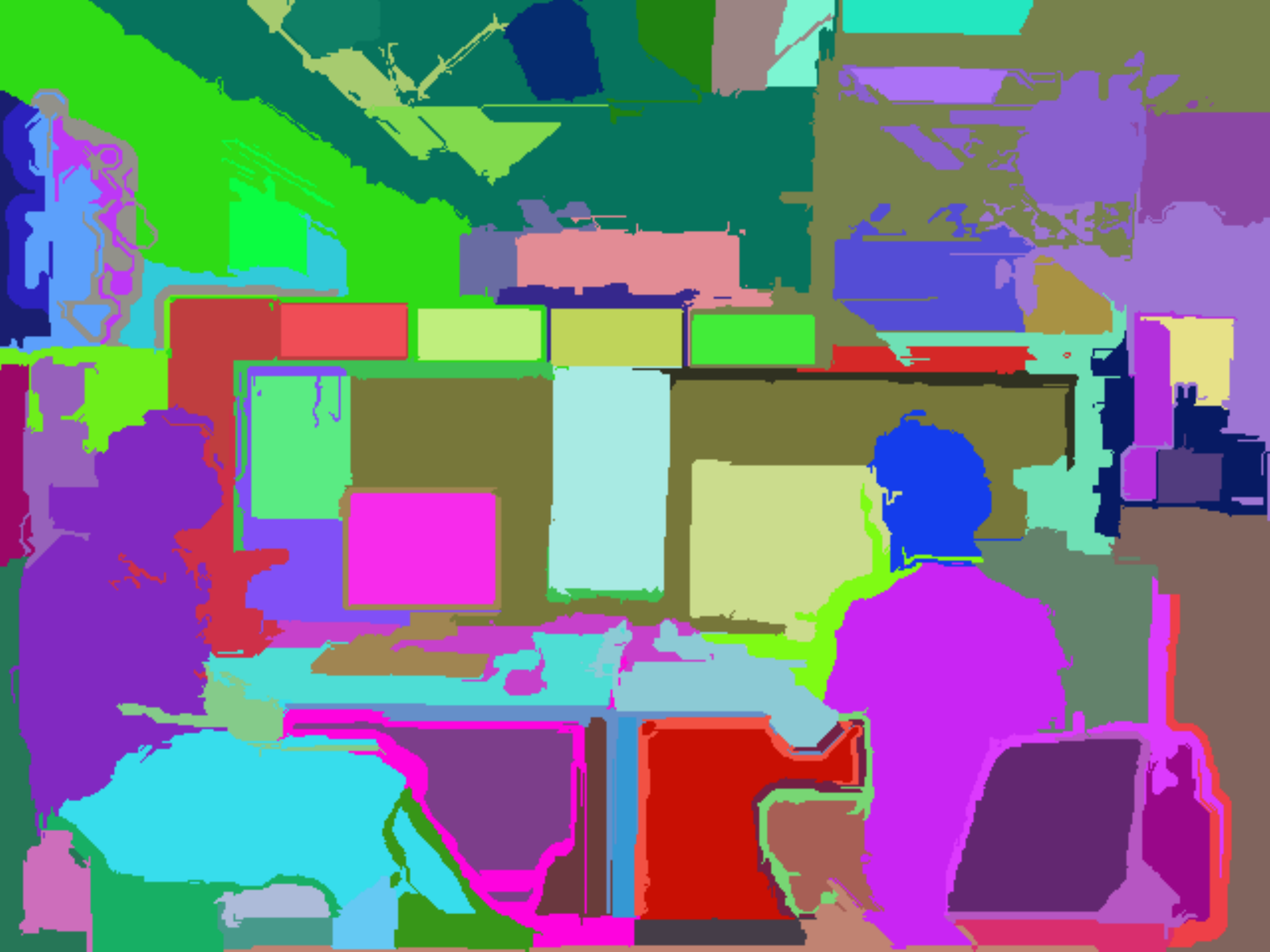}
    \end{tabular}
    \caption{Sequence 3, (A) : RGB image, (B): Results from \protect\cite{Xuy2012}, (C): Results from \protect\cite{Grundmann2010} (N/A is used when the frame is beyond the length it can run), (D): Our Results}
    \label{fig_fig13c}
\end{figure*}

\noindent
To quantify these results, we evaluated the output of the graph-based algorithm of~\cite{Felzenszwalb2004} for different values of $\alpha$ for all three sequences (shown in Figures \ref{fig_fig13a}-\ref{fig_fig13c} corresponding to S1-S3).  As shown in Figure~\ref{fig_fig7}, better results are obtained when color is weighted higher than depth ($\alpha$ is small).  Nevertheless, no value of $\alpha$ yields an acceptable segmentation.  Also shown is the initial over-segmentation step of our method.  Since our method is independent of $\alpha$, these results are plotted as horizontal lines.  Note that the segmentation error of our multistage segmentation algorithm is lower than the output from the linear combination approach for any value of $\alpha$. 

\begin{figure}
	\centering
	\begin{tabular}{cc}
		\includegraphics[width=0.5\linewidth]{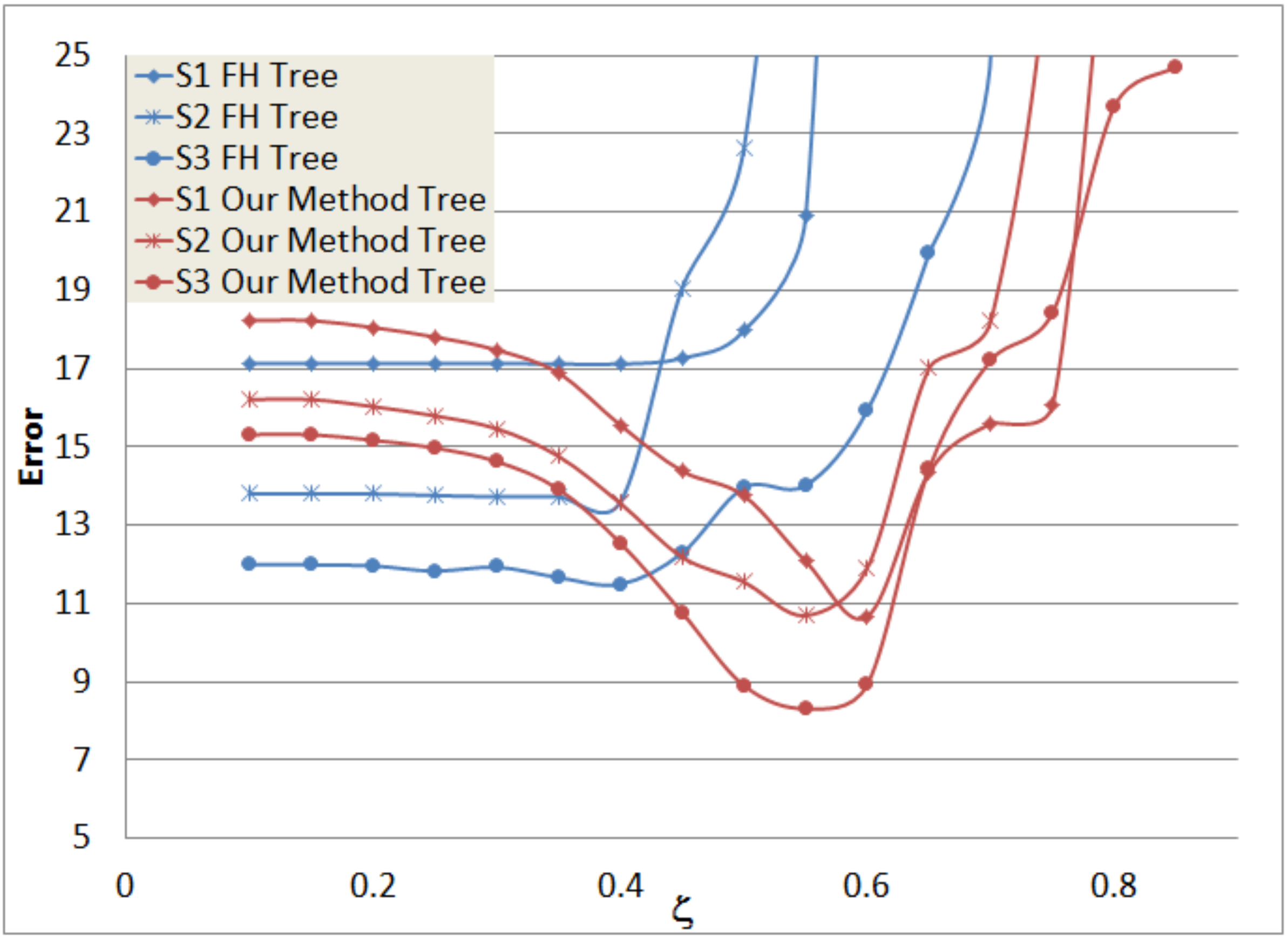} &
		\includegraphics[width=0.5\linewidth]{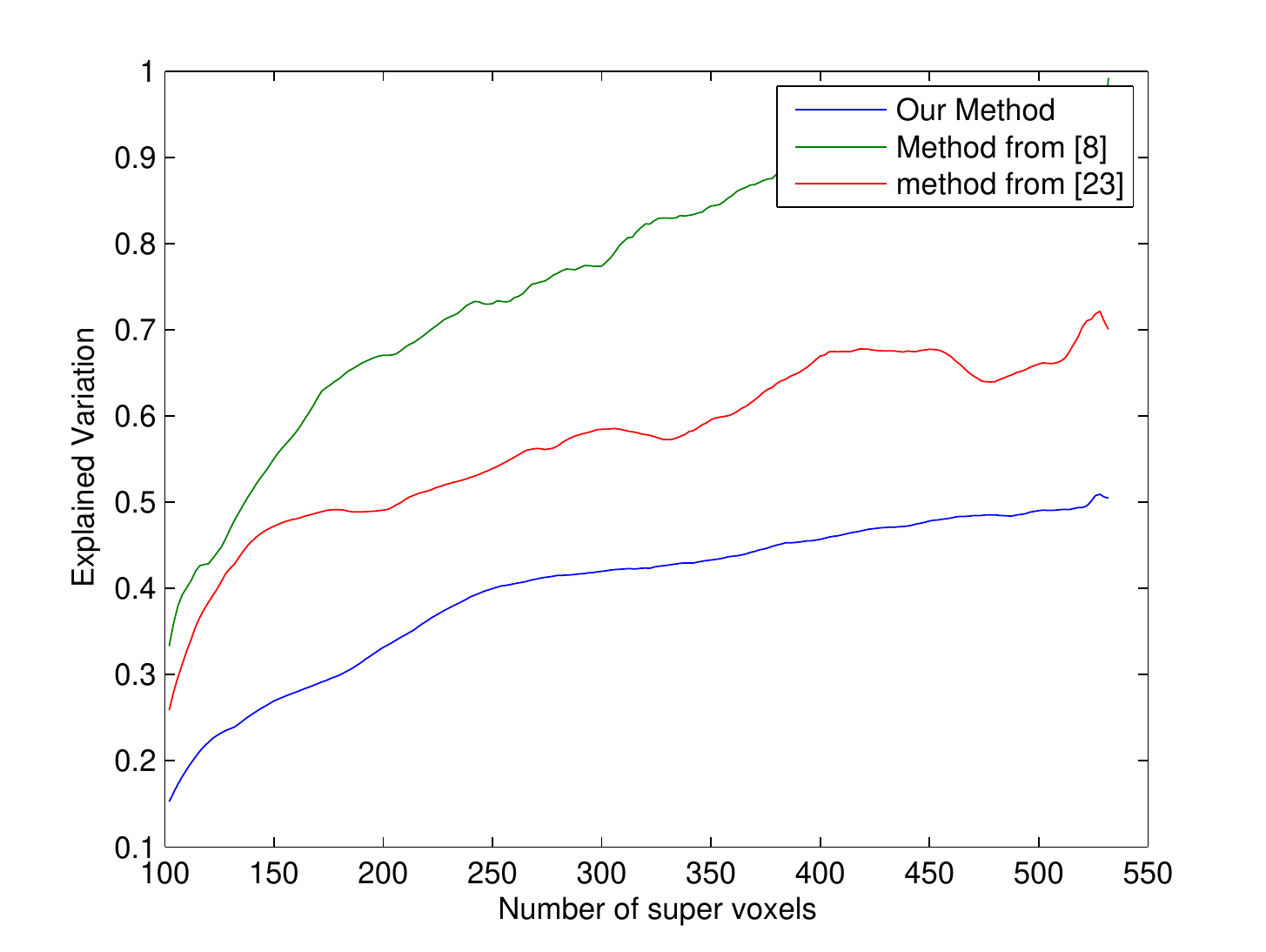}
	\end{tabular}
	\caption{Left:  Effect of varying $\zeta$ on the output of our hierarchical clustering applied to the output of either our initial over-segmentation step or that of~\cite{Felzenszwalb2004} (with the best $\alpha$ and $k$ possible).
	  Right:  The explained variation from \cite{Xuy2012} of our method, \cite{Xuy2012}, and \cite{Grundmann2010} on Sequence 2.  }
	\label{fig_fig9b}
\end{figure}

\noindent
Even though our algorithm does not depend on $\alpha$, it does initially depend on $k_{depth}$ and $k_{color}$. The left of Figure~\ref{fig_fig7} shows the error (measured by Equation~\ref{eq_err} where $k$ is the parameter from \cite{Felzenszwalb2004}) obtained from segmenting just based on depth as the parameter $k_{depth}$ is changed. Note that on all three images the output gets worse as this value is increased due to under-segmentation.  On the right is a plot of all three sequences for both approaches discussed in \ref{fig_fig6}, where our algorithm includes only the initial over-segmentation step.  With a low $k_{depth}$, these plots show the effect of changing the value of $k_{color}$ for our algorithm, or similarly of $k$ for~\cite{Felzenszwalb2004}.  For nearly all values our method achieves less error.

\begin{figure*} [t]
    \centering   
    \begin{tabular}{cccccc}
      \includegraphics[width=0.15\linewidth]{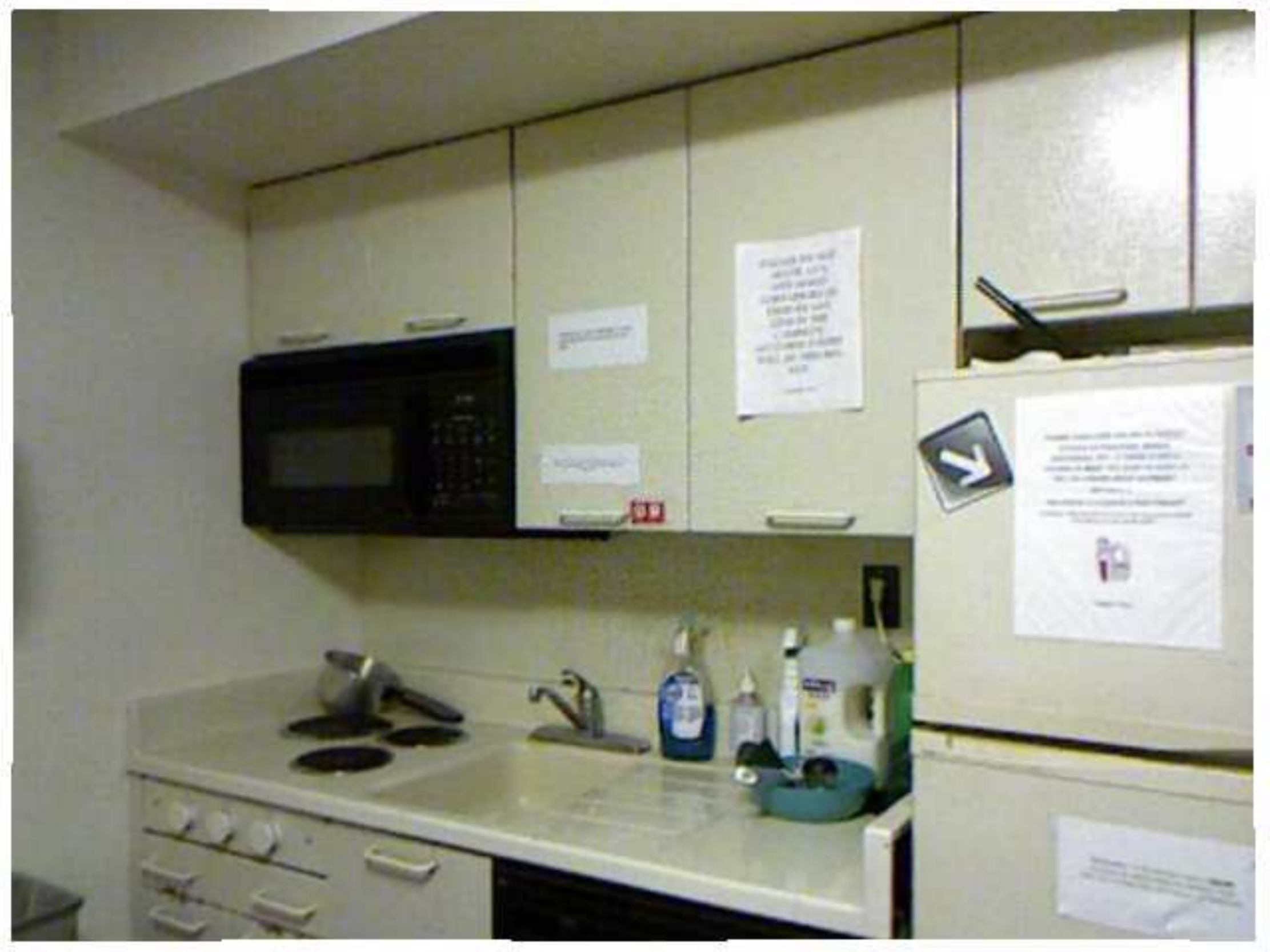} &
      \includegraphics[width=0.15\linewidth]{fig/Seq8/9.eps} &
      \includegraphics[width=0.15\linewidth]{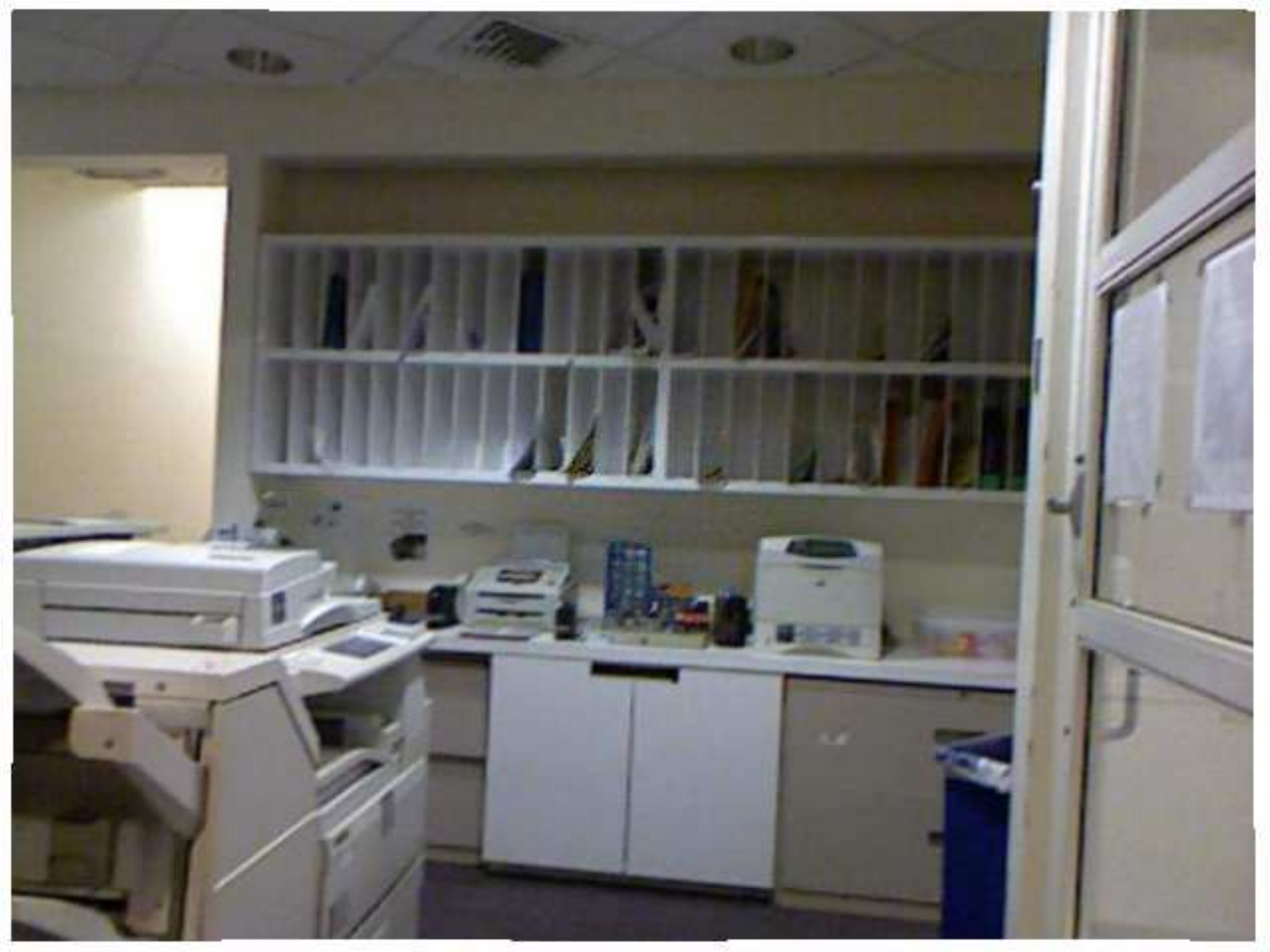} &
      \includegraphics[width=0.15\linewidth]{fig/Seq8/47.eps} &
			\includegraphics[width=0.15\linewidth]{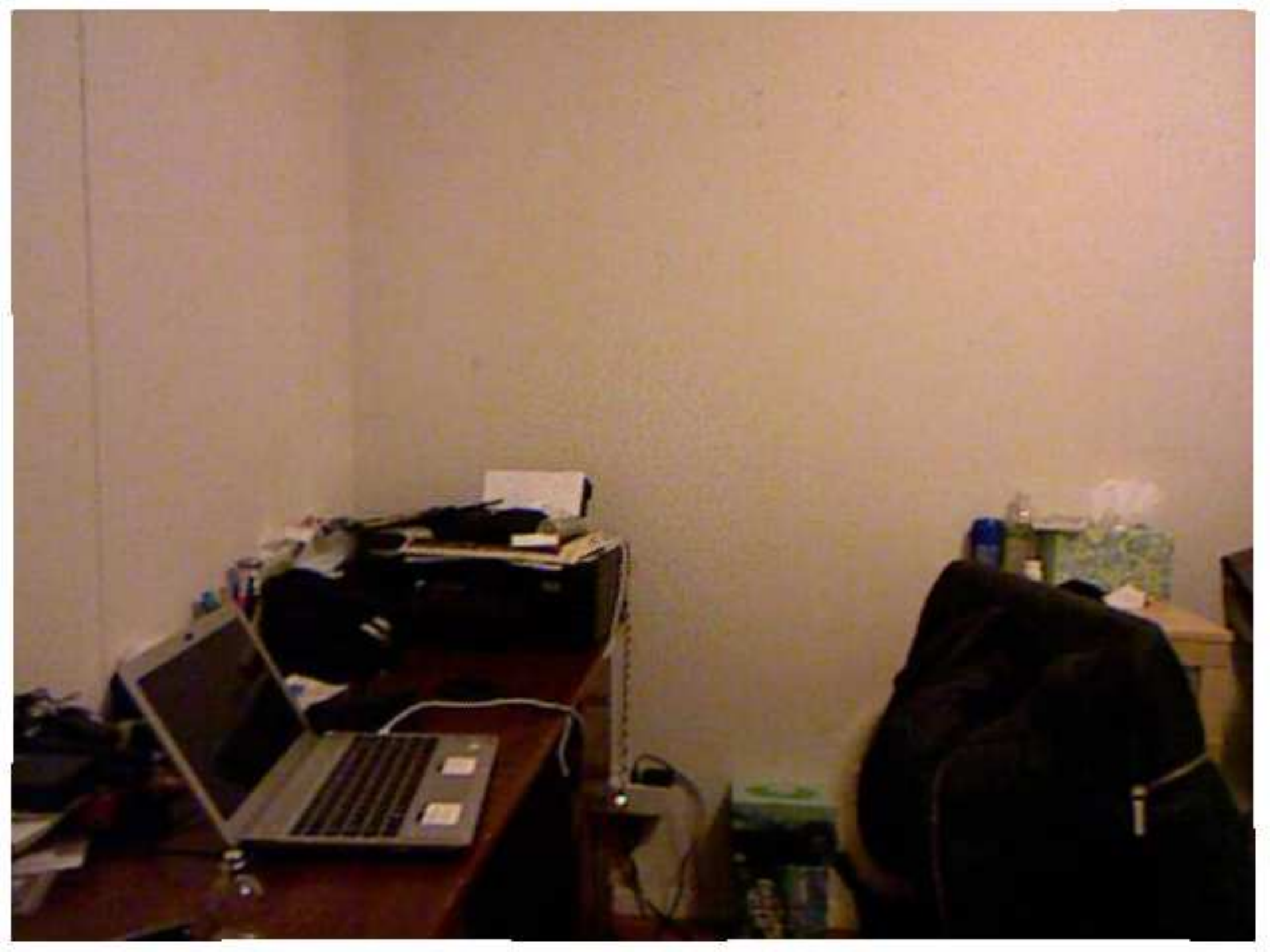} &
			\includegraphics[width=0.15\linewidth]{fig/Seq8/62.eps}
    \end{tabular}
    \begin{tabular}{cccccc}
      \includegraphics[width=0.15\linewidth]{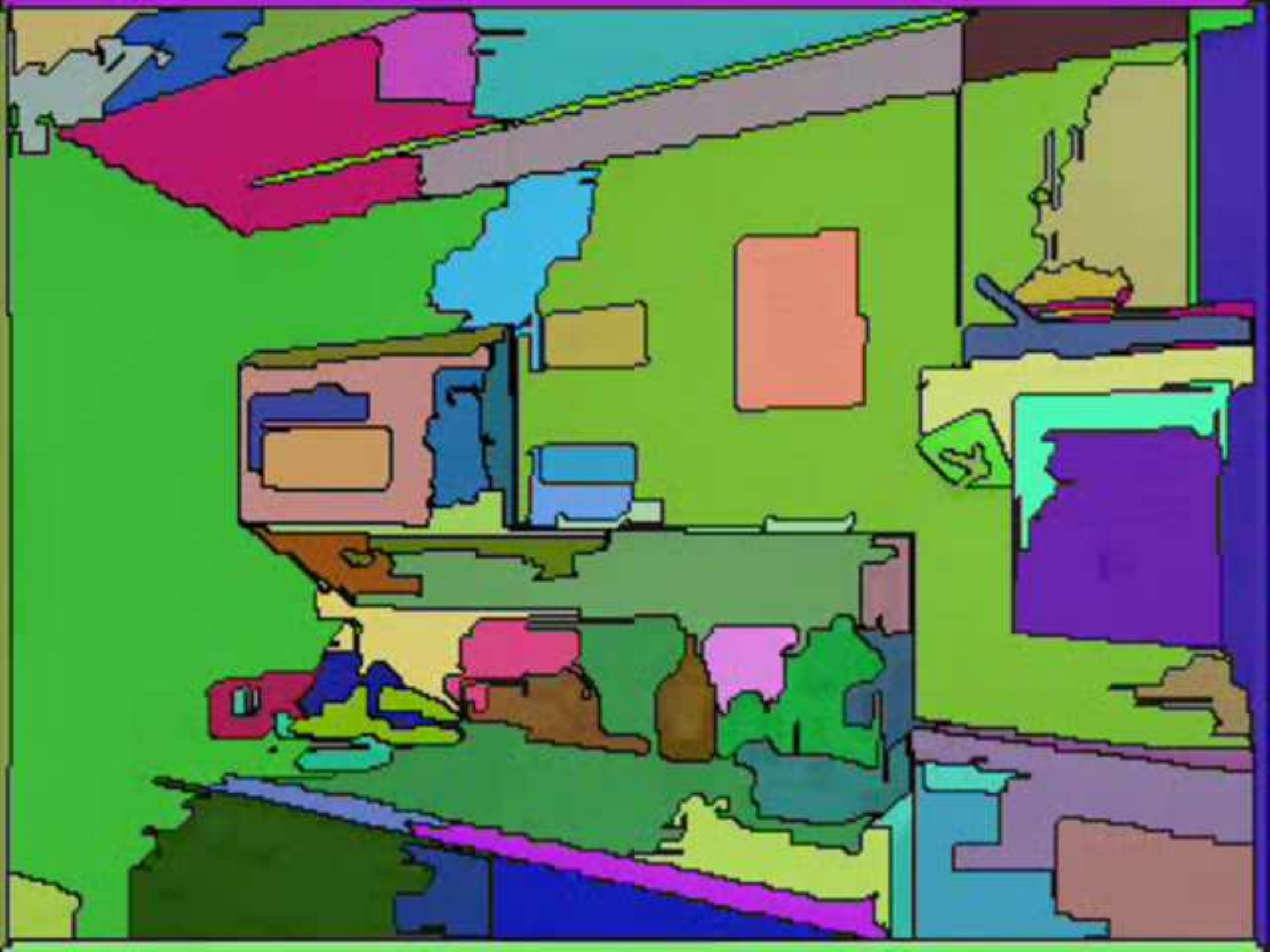} &
      \includegraphics[width=0.15\linewidth]{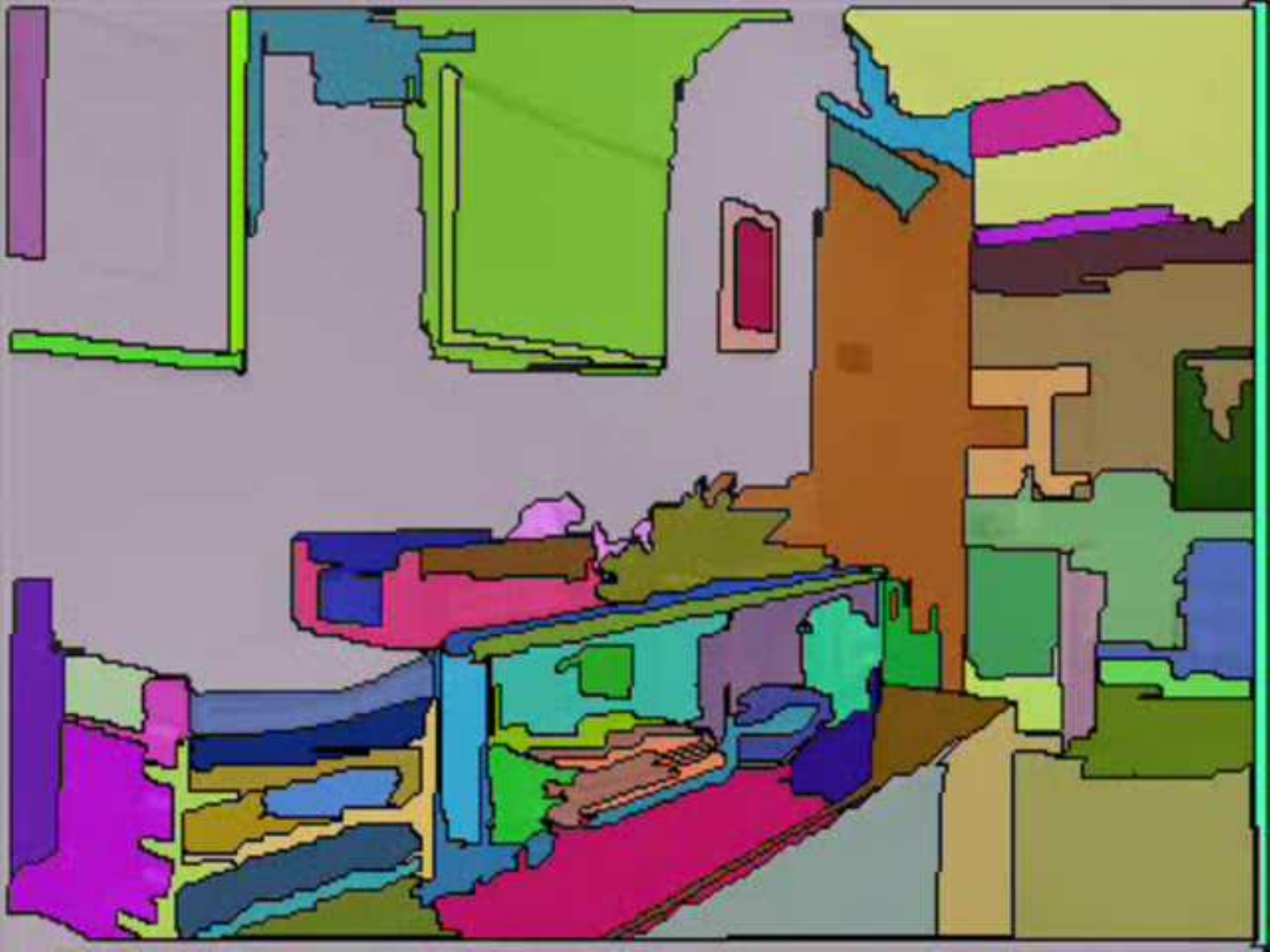} &
      \includegraphics[width=0.15\linewidth]{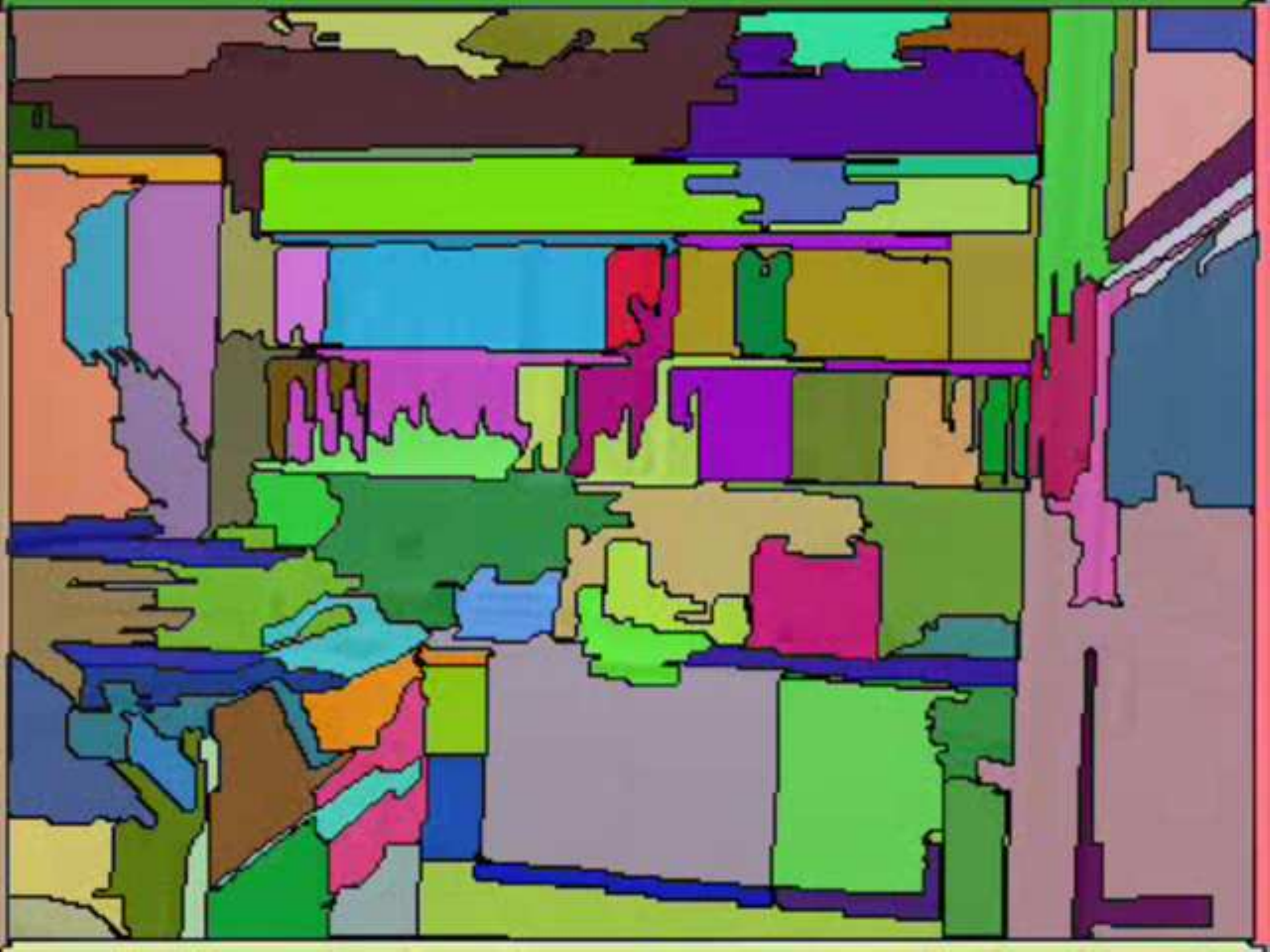} &
      \includegraphics[width=0.15\linewidth]{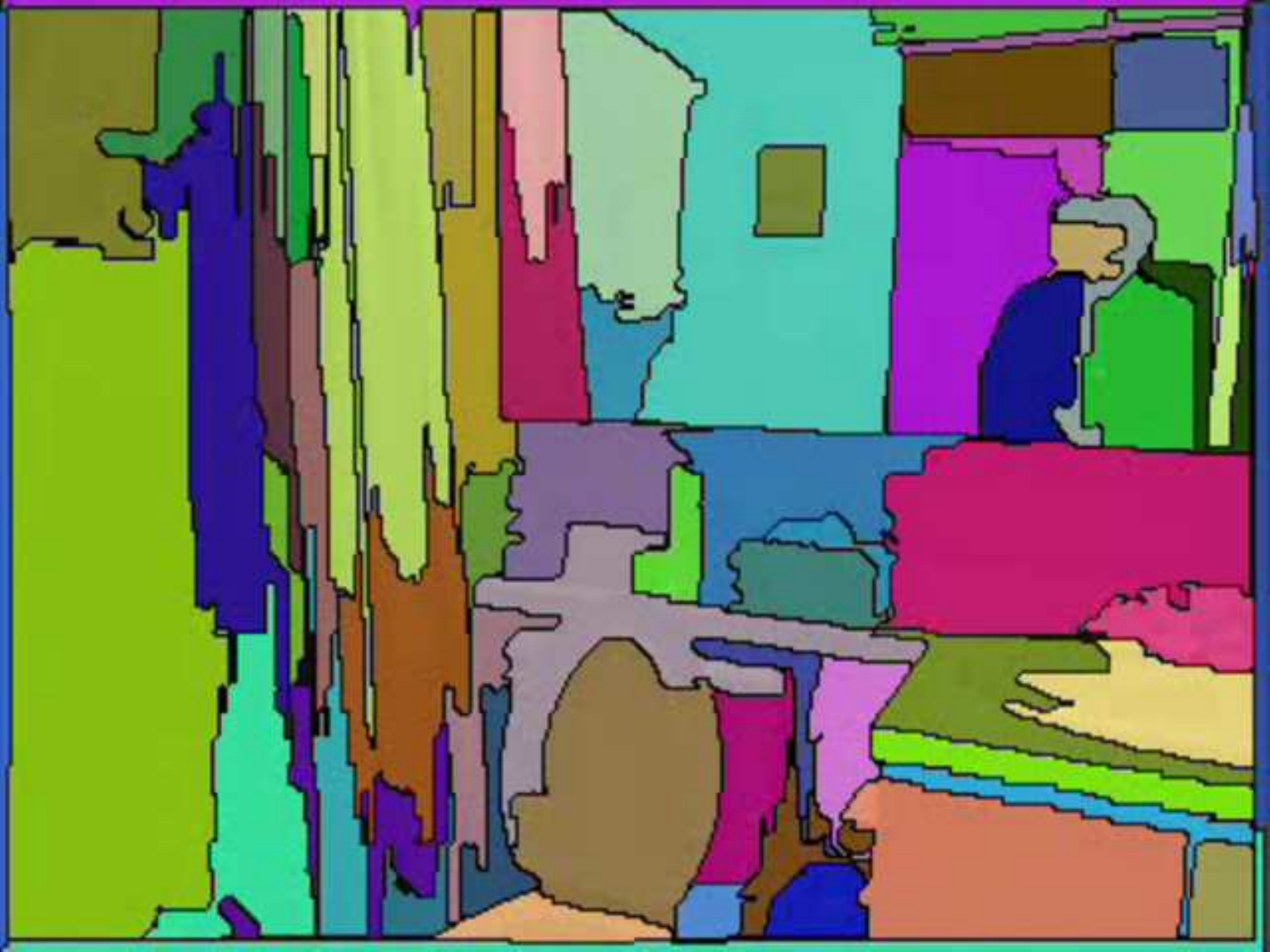} &
			\includegraphics[width=0.15\linewidth]{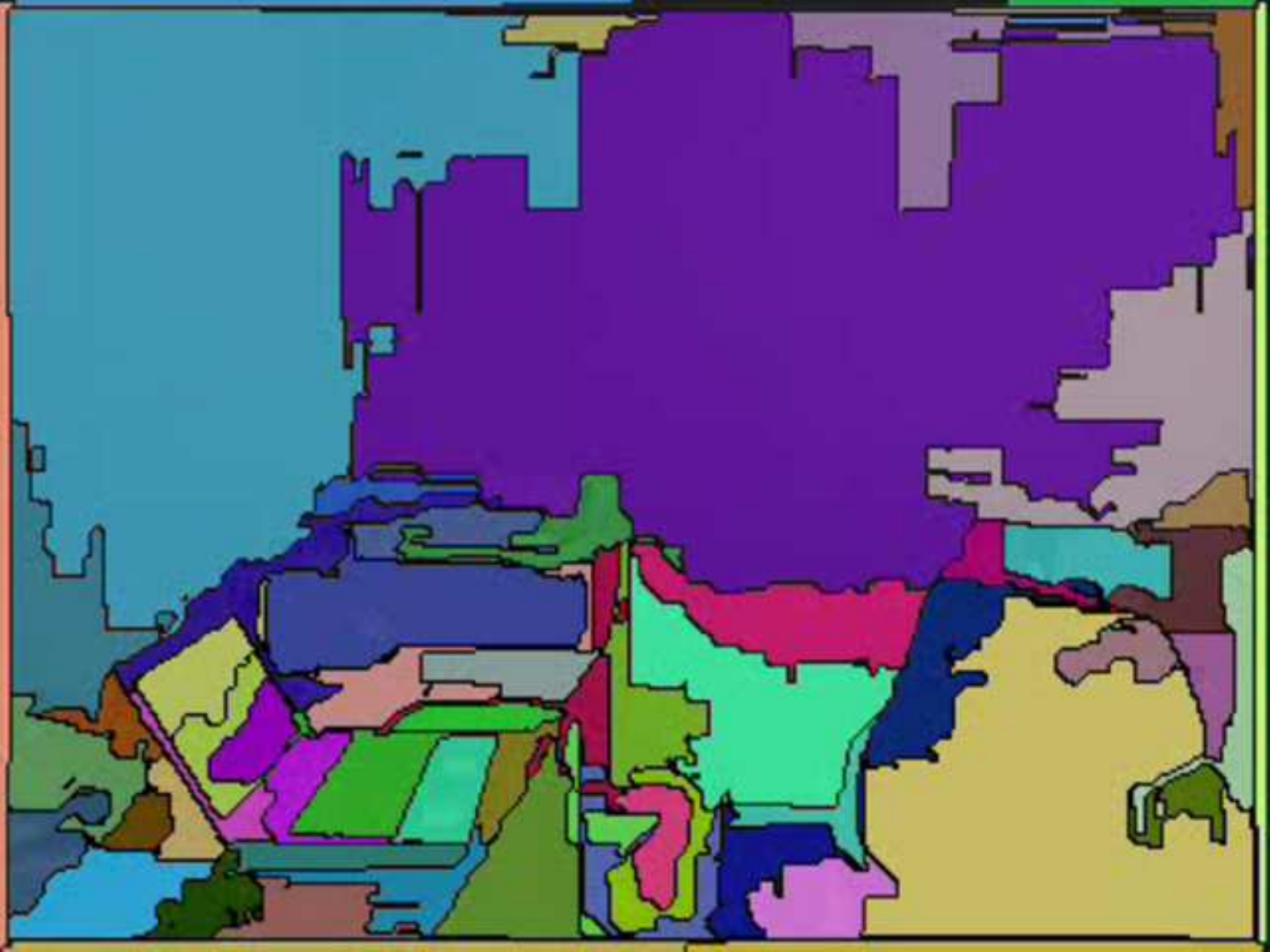} &
			\includegraphics[width=0.15\linewidth]{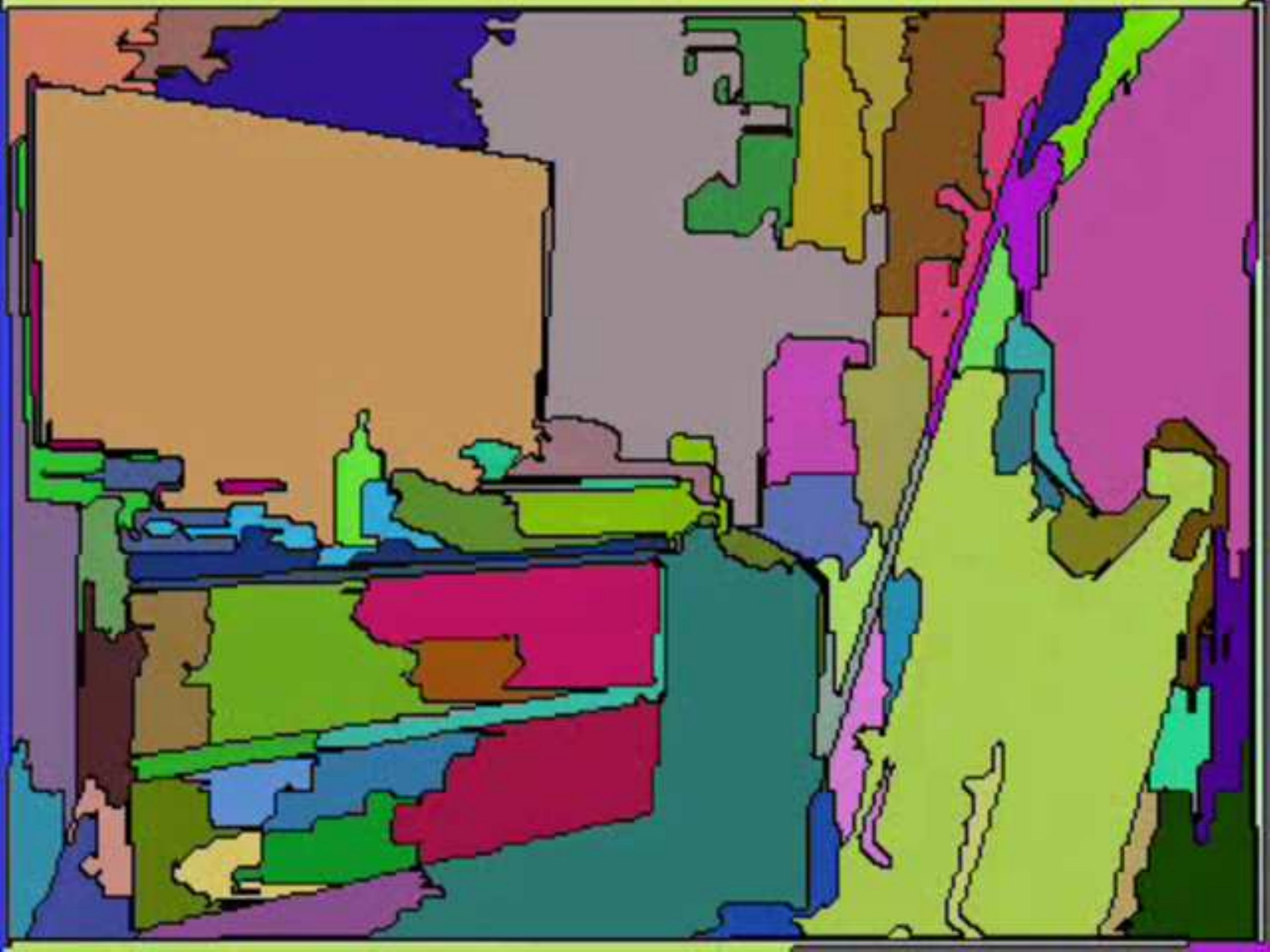}
    \end{tabular}
		\begin{tabular}{cccccc}
      \includegraphics[width=0.15\linewidth]{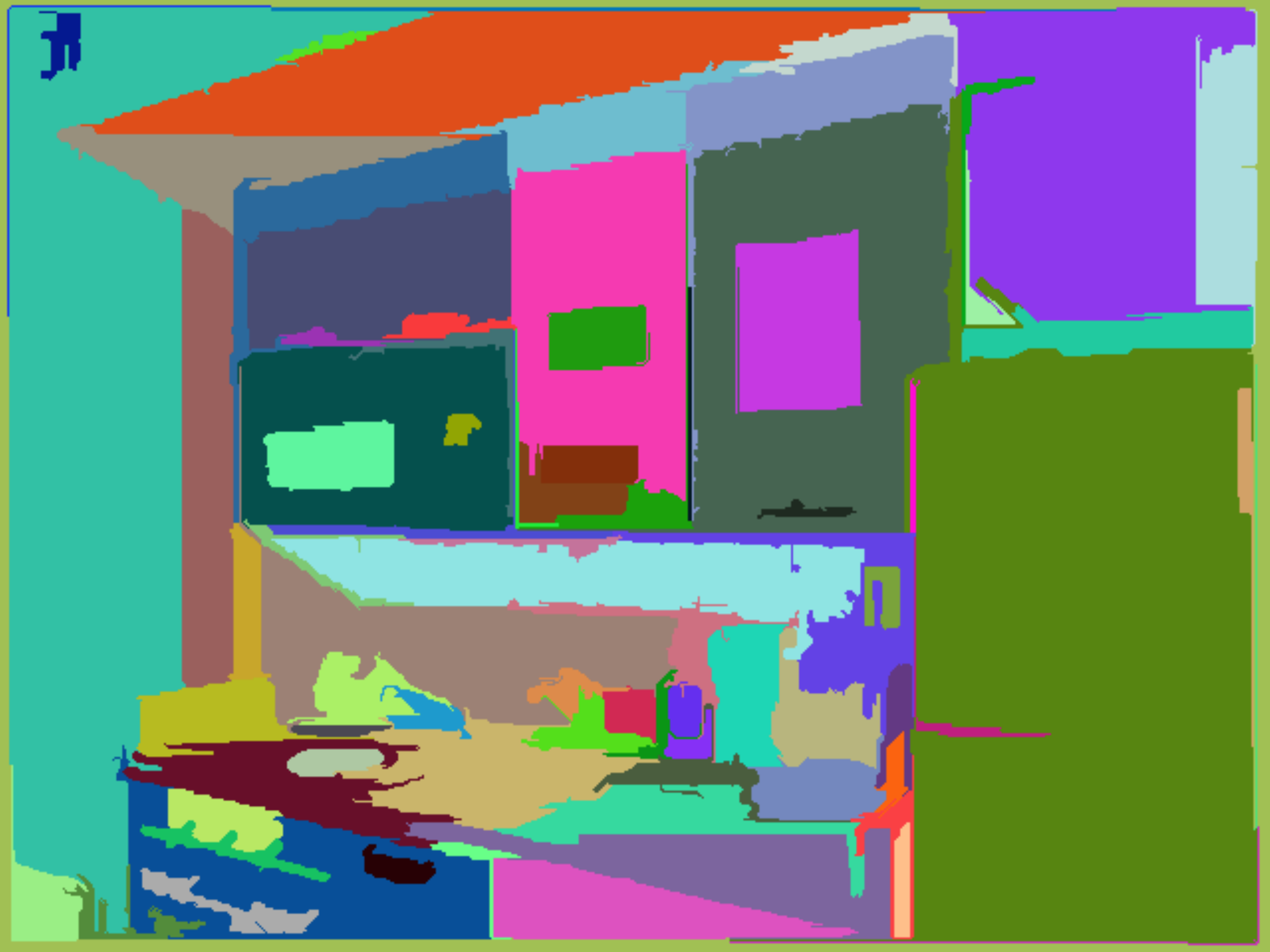} &
      \includegraphics[width=0.15\linewidth]{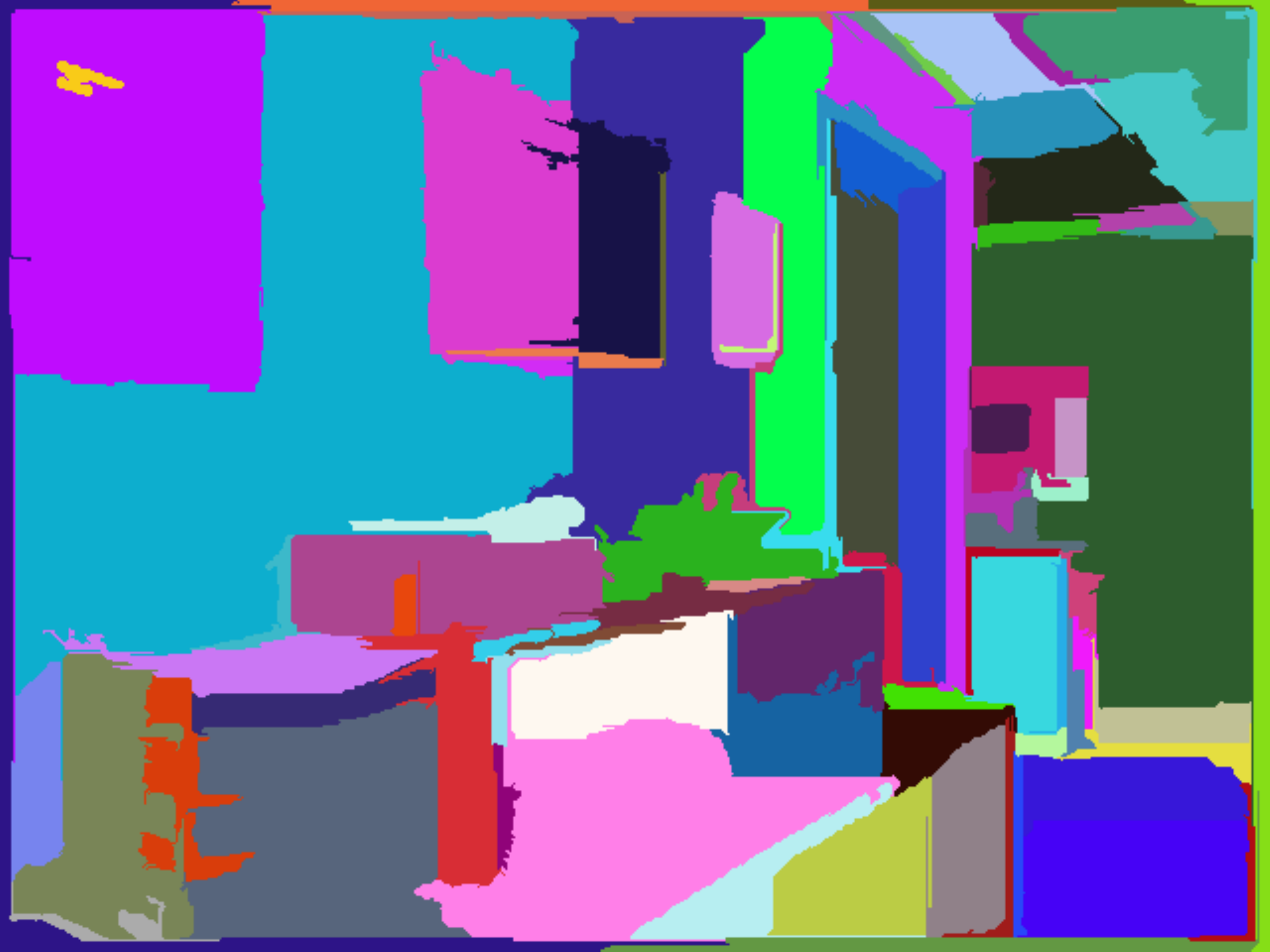} &
      \includegraphics[width=0.15\linewidth]{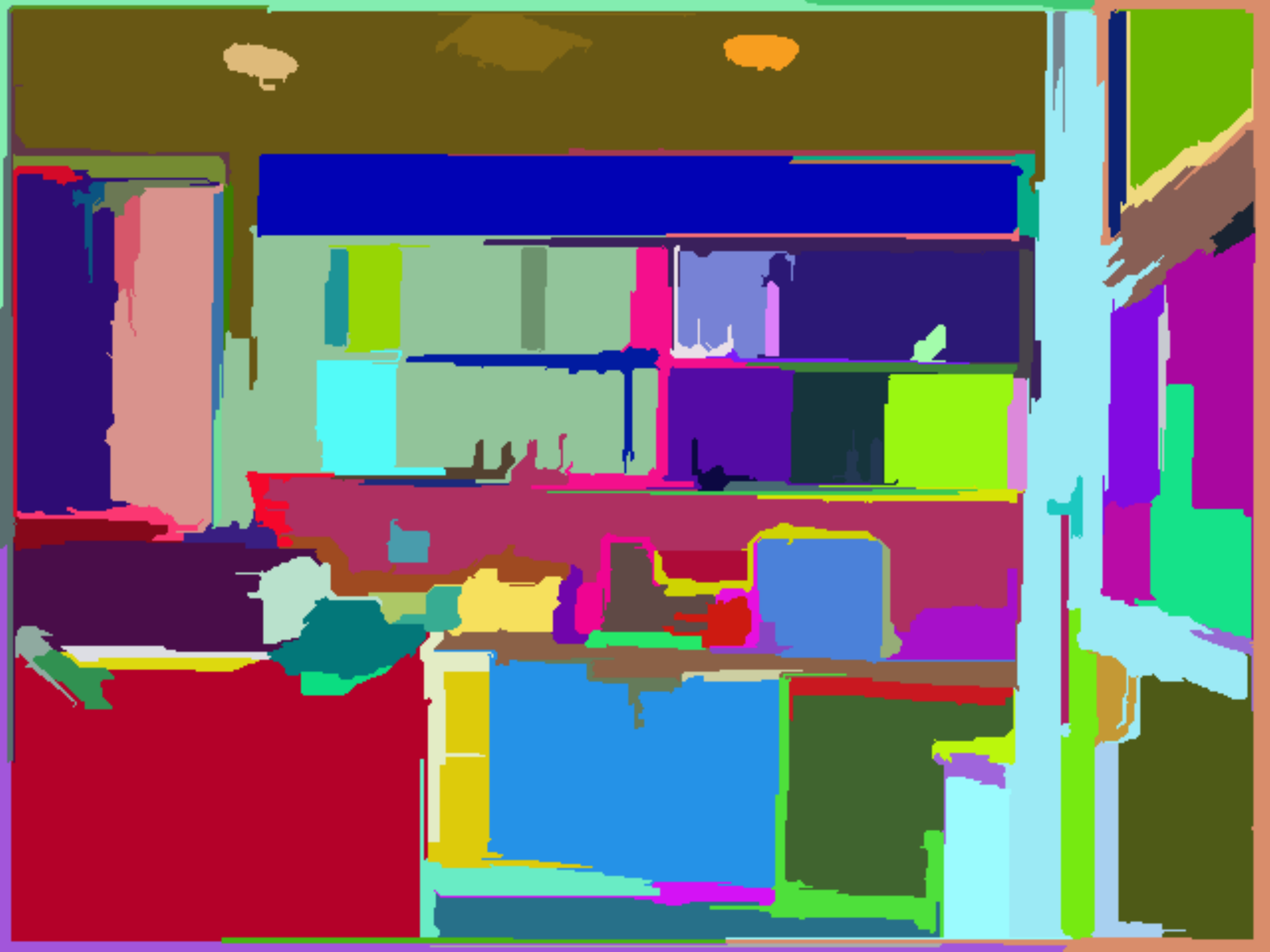} &
      \includegraphics[width=0.15\linewidth]{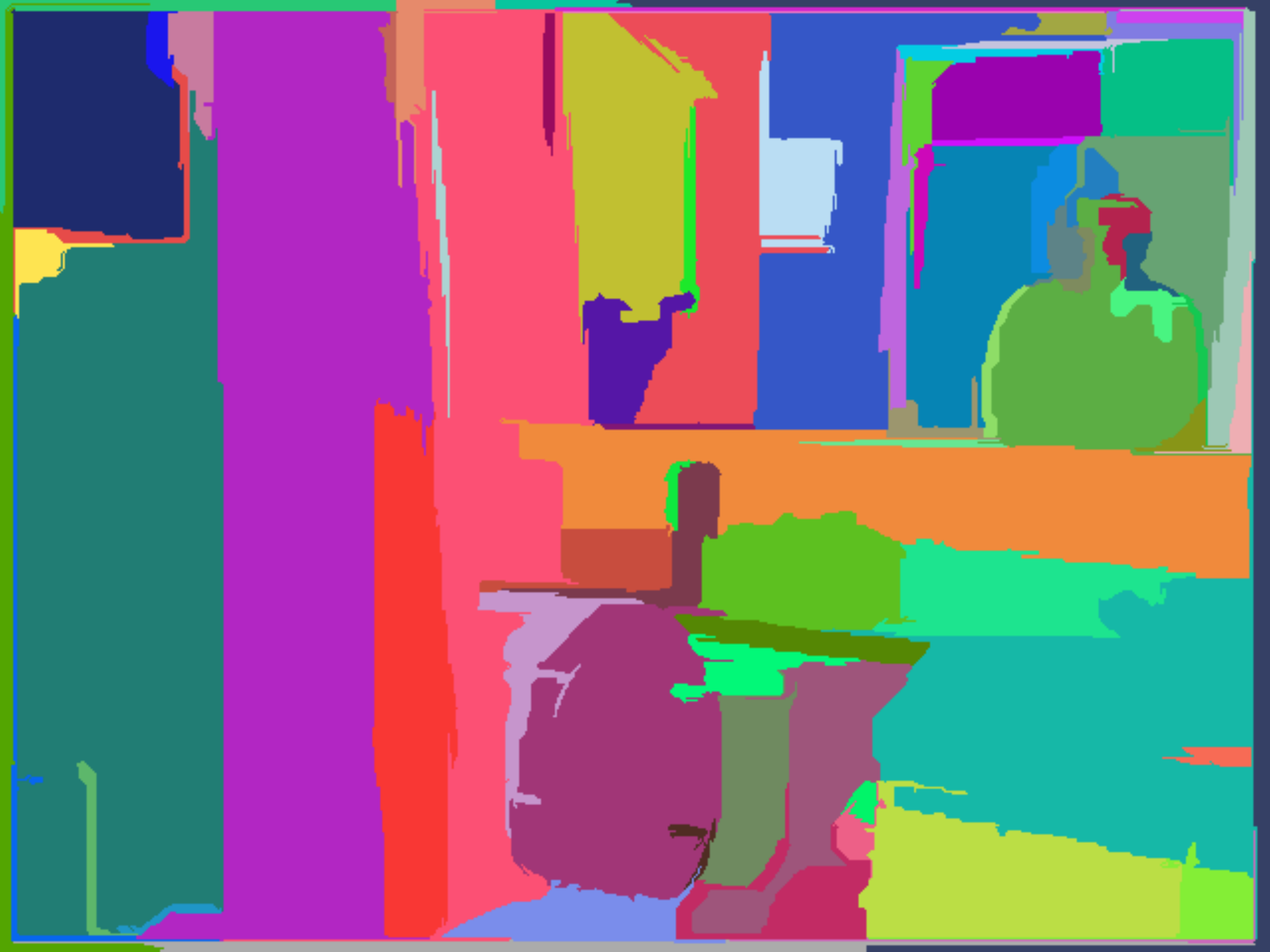} &
			\includegraphics[width=0.15\linewidth]{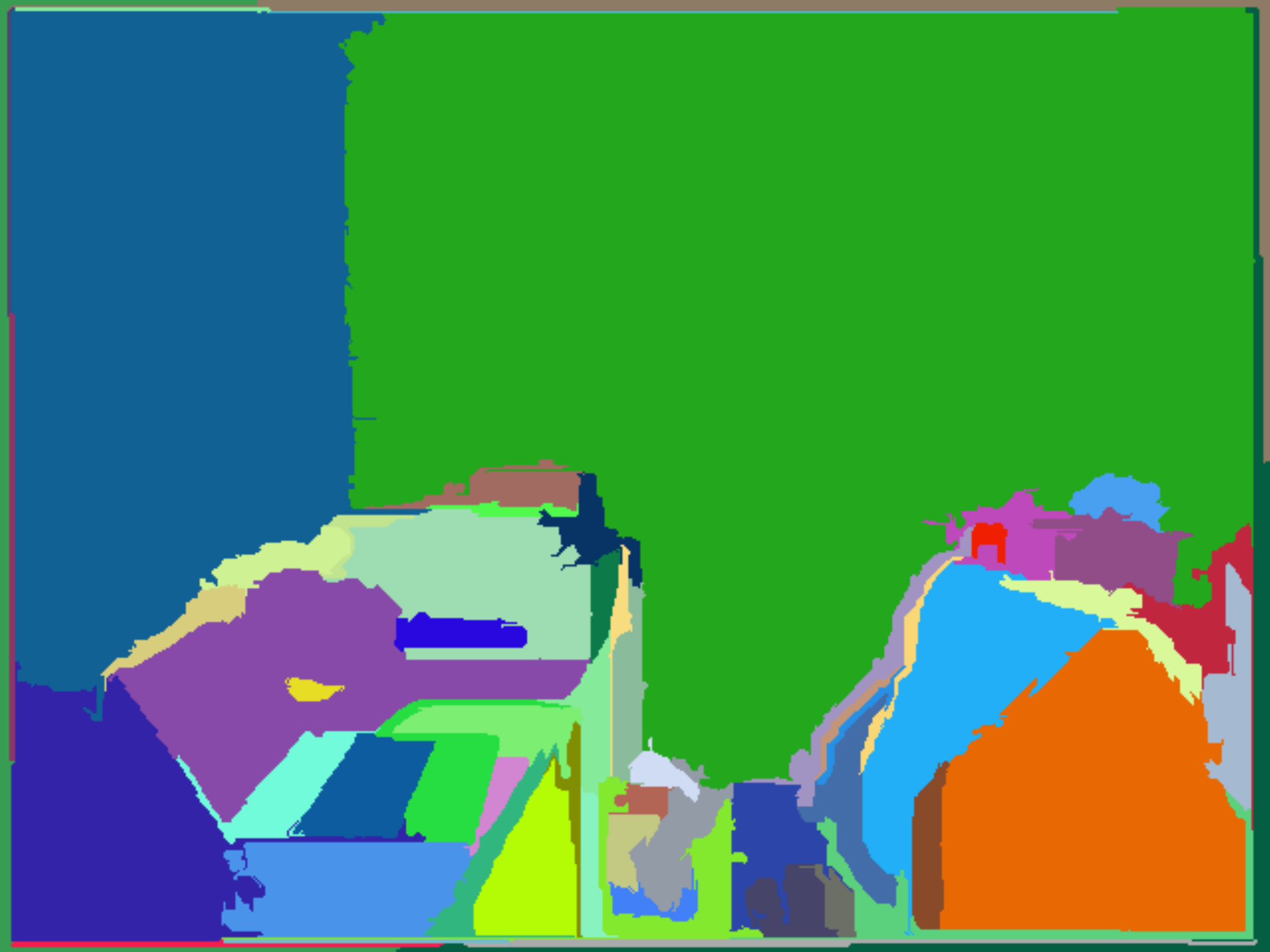} &
			\includegraphics[width=0.15\linewidth]{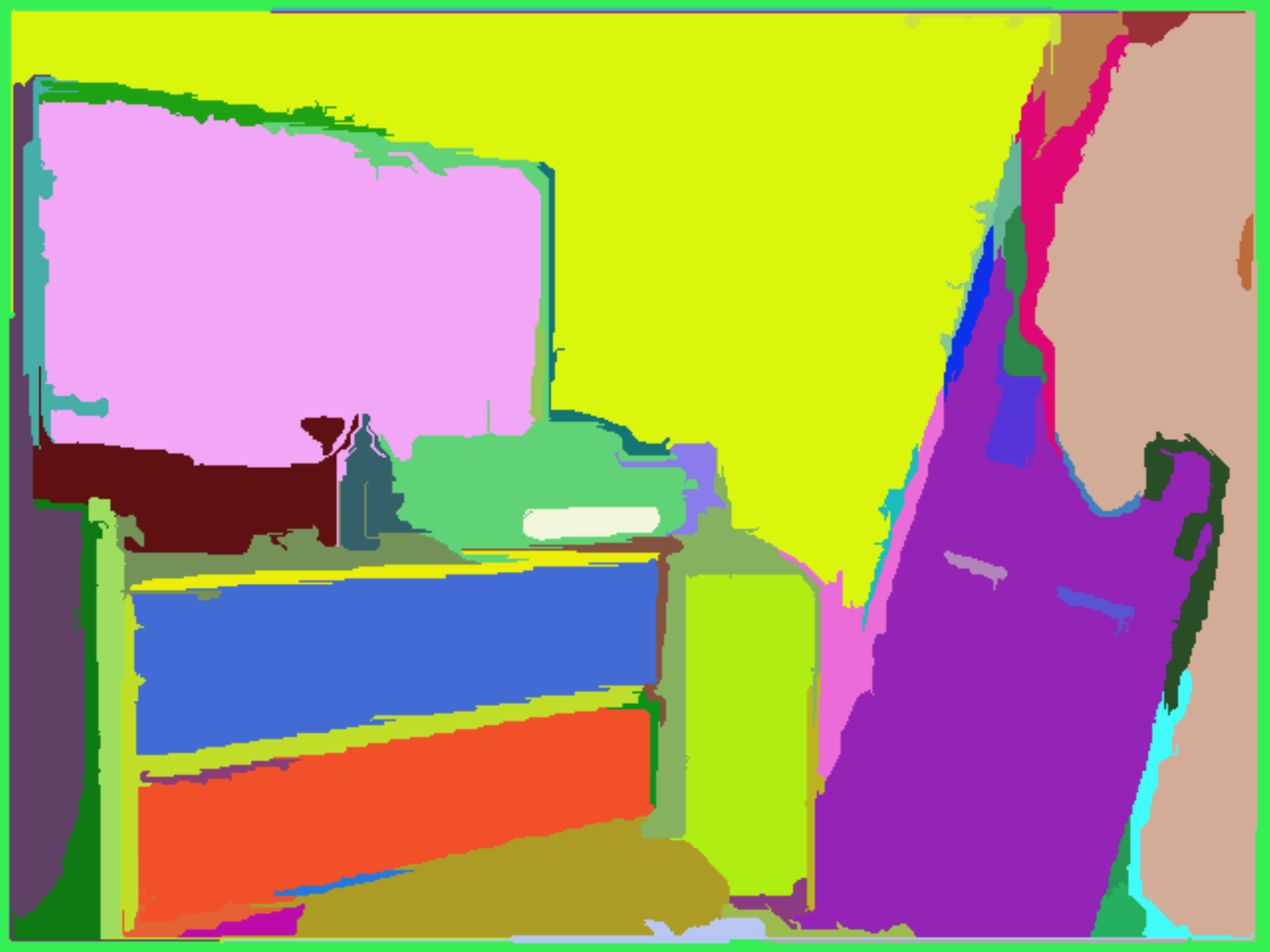}
    \end{tabular}
\caption{Segmentation results from different 3D RGBD scenes from the NYU data set\cite{Silberman2012}. The top row is the images, the 2nd row is the hierarchical result only using color with a hierarchical level of 0.7 described in \cite{Grundmann2010}, and the bottom row is our approach with a hierarchical level of 0.7.}
    \label{fig_fig14}
\end{figure*}

\noindent
Figure~\ref{fig_fig9b} shows the error using the hierarchical method. On the left is a set of plots of all three sequences for both approaches, where we vary the hierarchical tree percentage cut. It can be seen that the hierarchy makes us invariant to the values of $k_{depth}$ and $k_{color}$ just as in \cite{Grundmann2010}. On the right are plots of the error of our hierarchical step applied to the output of both our algorithm (initial over-segmentation step) and the algorithm of~\cite{Felzenszwalb2004}.  These plots answer two questions, namely, whether the hierarchical step improves results, and whether our initial over-segmentation step is needed (or whether the hierarchical step would work just as well if we simply used the algorithm of~\cite{Felzenszwalb2004}).  The answer to both questions is in the affirmative. The rightmost figure shows that our method also has a lower explained variation. Further proof that having depth is better than just color is shown in the comparison of RGBD images from the NYU Dataset \cite{Silberman2012} in Figure~\ref{fig_fig14}.

\noindent
Now that we have quantified our algorithm, we show segmentation results of the entire approach on three different 4D RGBD sequences compared to the state of the art video sequences of \cite{Xuy2012} and \cite{Grundmann2010}, shown in Figures~\ref{fig_fig13a}, \ref{fig_fig13b}, and \ref{fig_fig13c}. Note that as people move around the scene, the camera moves, and brief partial occlusions occur, the segmentation results remain fairly consistent and the identity of regions are mostly maintained. It can be seen that our approach seems to always outperform \cite{Xuy2012} in both segmentation and temporal consistency. We maintain boundaries better and segment better than \cite{Grundmann2010}. \cite{Grundmann2010} also tends to merge segments together improperly such as the people merged into the background in Figures~\ref{fig_fig13b} and \ref{fig_fig13c}. Although \cite{Grundmann2010} maintains temporal consistency better than our algorithm, it processes all frames at once as opposed to being stream-based and cannot run indefinitely, which is why last image in Sequence (C) is always N/A. Although there is a perceived error in boundaries, it is only due to the lack of overlap in the registration of the depth and color images inherent from the RGBD camera. 
Our algorithm runs at approximately 0.8 fps on 8 frame sequences of 640x480 RGB and depth images. \cite{Grundmann2010} runs in 0.2 fps on downscaled 240x180 videos and \cite{Xuy2012} runs in 0.25 fps on downscaled 240x180 videos. See the project web page for videos.

\section{Conclusion, Limitations, and Future Work}

\noindent
In this paper, we have presented an unsupervised hierarchical segmentation algorithm for RGBD videos.  The approach exhibits accurate performance in terms of segments quality, temporal region identity coherence, and computational time. The algorithm uses a novel approach for combining the depth and color information to generate over-segmented regions in a sequence of frames. A hierarchical tree merges the resulting regions to a level defined by the user, and a bipartite matching algorithm is used to ensure temporal continuity. We showed that our method outperforms a linear combination of color and depth. We performed comparison against different graph segmentation combinations showing lower errors in terms of quality of the segmentation and thoroughly analyzing the effects of various parameters. 
There are occasional problems with noise from the depth image and maintaining temporal consistency. Future work will be aimed at improving results even further by incorporating boundary information, as well as incorporating higher-level information.

\section{Acknowledgments}

\noindent
The authors would like to thank Prof. Mubarak Shah and his PhD student Gonzalo Vaca-Castano for their mentorship and guidance to the primary author of this paper, when he participated in the National Science Foundation funded "REU Site: Research Experience for Undergraduates in Computer Vision" (\#1156990) in 2012 at the University of Central Florida's Center for research in Computer Vision. 




{\small
\bibliographystyle{ieee}
\bibliography{cvpr2013}
}
\end{document}